\documentclass{article}

\usepackage[preprint]{neurips_2025}

\usepackage[utf8]{inputenc}
\usepackage[T1]{fontenc}    
\usepackage{hyperref}       
\usepackage{url}            
\usepackage{booktabs}       
\usepackage{amsfonts}       
\usepackage{nicefrac}       
\usepackage{microtype}      
\usepackage{xcolor}         
\usepackage{graphicx}
\usepackage{subcaption}
\usepackage{float}
\usepackage{amsmath}
\DeclareMathOperator{\softmax}{softmax}
\usepackage{tabularx}
\usepackage{array}
\newcolumntype{Y}{>{\raggedright\arraybackslash}X} 
\usepackage{afterpage}

\title{Directional Reasoning Trajectory Change (DRTC): Identifying Critical Trace Segments in Reasoning Models}

\author{
  Waldemar Chang \\
  Johns Hopkins University \\
  \texttt{wchang58@jh.edu} \\
}

\begin{document}

\maketitle

\begin{abstract}
Understanding how language models carry out long-horizon reasoning remains an open challenge. Existing interpretability methods often highlight tokens correlated with an answer, but rarely reveal \emph{where} consequential reasoning turns occur, \emph{which} earlier context triggers them under causal intervention, or whether highlighted text actually \emph{steers} the rollout. We introduce \emph{Directional Reasoning Trajectory Change} (DRTC), a process-causal method that (i) detects pivot decision points via uncertainty and distribution-shift signals and (ii) applies receiver-side interventions that preserve the realized continuation \emph{without resampling} while blocking information flow from selected earlier chunks \emph{only at a pivot}. DRTC measures how each intervention redirects the log-probability trajectory relative to the realized rollout direction, yielding signed per-chunk attributions; we also compute logit-space curvature changes and curvature signatures as a complementary geometric diagnostic. Across four reasoning models, influence is sharply concentrated (Gini $\approx 0.50$--$0.58$, top-$5\%$ mass $\approx 0.23$--$0.28$), and learned pivots induce stronger effects than matched random spans. In a 500-problem MATH scaling study with \texttt{R1-Distill-Qwen-1.5B}, learned spans continue to outperform matched random spans (median $\Delta=0.409$, 355/500 positive; $p=2.3\times10^{-21}$), and curvature-impact co-localizes with DRTC within traces (diagnostic). We benchmark against gradient- and perturbation-based chunk attributions and show graded outcome linkage: under embedding-interpolation edits, top-ranked DRTC chunks reduce teacher-forced gold-answer log-probability more than strict position-matched random chunks on a stability-filtered subset. Overall, DRTC provides a causally grounded view of how specific context elements steer on-policy reasoning trajectories.
\end{abstract}

\afterpage{%
\begin{figure*}[t]
  \centering
  \includegraphics[width=\textwidth]{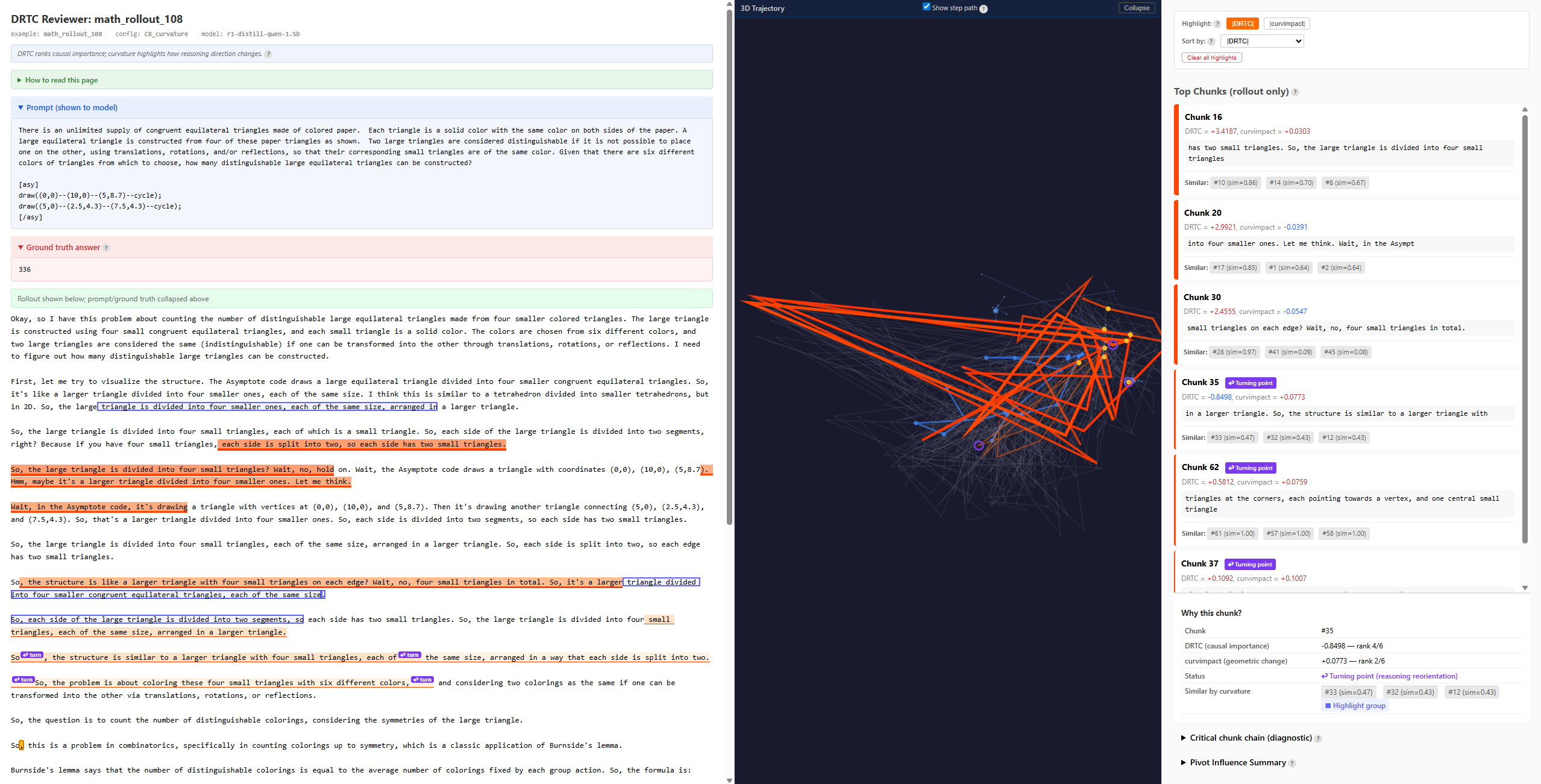}
  \caption{\textbf{DRTC inspection interface.}
  The 3D PCA view is qualitative; curvature is computed on logits.
  The UI is for auditing only and is not used to define pivots or scores (see Appendix~\ref{app:ui}).}
  \label{fig:drtc_ui_demo}
\end{figure*}
}

\section{Introduction}

Reasoning models increasingly solve problems by generating long, winding traces with backtracking, verification, and strategy shifts, rather than simple forward inference \citep{kim2026reasoningmodelsgeneratesocieties}. This creates a central interpretability gap: for many deployments, we do not just want to know \emph{what} a model answered, but \emph{how it got there}---when it changed its mind, what information triggered the change, and which earlier context causally shaped the computation that followed.

\paragraph{Why standard interpretability breaks for reasoning traces.}
Consider a common pattern in math and planning: the model writes dozens of sentences pursuing an incorrect approach, then produces a ``wait''-style hesitation, checks its work, abandons the approach, adopts a new strategy, and only much later commits to the final answer \citep{lee2025evaluatingstepbystepreasoningtraces,huang2025efficientreasoninglargereasoning}.
In such traces, it is not clear which internal computation should be analyzed, since modern reasoning chains interleave exploratory, corrective, and outcome-determining steps, and a single sampled trace need not cleanly reflect the underlying causal structure \citep{macar2025thoughtbranchesinterpretingllm}.
Studying the forward pass that emits the final answer is often unhelpful: long-horizon generation is mediated by intermediate reasoning steps, later tokens depend on earlier commitments, and multi-token answers introduce additional sequential dependencies \citep{pan2026longhorizoninterpretabilityefficientfaithful,ling2023deductiveverificationchainofthoughtreasoning}.
More fundamentally, the reasoning process is \emph{path-dependent}: once the model commits to a line of thought, subsequent generations are constrained by that commitment, making it ill-posed to isolate a single decisive computation.
Causal interventions are likewise difficult to interpret: editing the trace while holding later text fixed is off-policy, whereas resampling after an edit often produces a qualitatively different trajectory that is hard to compare \citep{macar2025thoughtbranchesinterpretingllm}.
These issues reflect a deeper fact: long-horizon reasoning is a sampled, sequential computation with serial depth that exceeds a single forward pass, and standard attribution tools are poorly matched to this regime \citep{emmons2025chainthoughtnecessarylanguage,pan2026longhorizoninterpretabilityefficientfaithful}.

\paragraph{What we need instead.}
These failure modes suggest that analyzing a single forward pass or performing unrestricted trace edits is insufficient. Instability is also not incidental: when the next-token distribution is flat (high entropy / low margin), multiple continuations are plausible, and prior work shows it is beneficial to focus exploration or intervention on these high-uncertainty steps, whereas low-uncertainty steps can be followed greedily \citep{wang2025cautioustokenprediction,li2026entropygatedbranchingefficienttesttime,yang2026moreimprovingllmreasoning}. An adequate method must therefore (i) localize such \emph{critical decision moments} within a realized trace, and (ii) perform \emph{temporally valid} causal tests of whether specific earlier context segments influenced the model \emph{at those moments}, while preserving the realized rollout as the reference trajectory rather than editing the trace or resampling alternative continuations. This should remain meaningful even when ``the outcome'' is ambiguous (open-ended tasks) or not yet produced (truncated rollouts), since many real settings involve unfinished or non-terminal reasoning.

\paragraph{Our approach: Directional Reasoning Trajectory Change (DRTC).}
We introduce \emph{Directional Reasoning Trajectory Change} (DRTC), a process-causal framework for interpreting long-horizon reasoning from a single on-policy rollout.
DRTC makes four methodological contributions:
\begin{itemize}
  \item \textbf{Pivot-localized decision-point discovery:}
  We identify \emph{decision pivots} within a realized chain-of-thought using uncertainty and distribution-shift signals (entropy, top-2 margin, and Jensen--Shannon divergence), targeting moments where the computation is unstable and strategy may change.

  \item \textbf{Temporally valid, on-policy causal interventions:}
  At each pivot, we apply receiver-side attention masking that preserves the realized rollout \emph{without resampling the continuation} while blocking information flow from selected earlier chunks \emph{only at that pivot}, yielding deterministic chunk--pivot causal effects and enabling screening gates that downweight noisy probes.

  \item \textbf{Directional trajectory attribution as the causal target:}
  We measure whether each intervention redirects the \emph{direction} of the model's log-probability trajectory relative to the realized rollout direction, aggregating pivot importance, causal relevance, and directional effect into a signed per-chunk attribution score rather than relying on final-answer flips or likelihood-only criteria.

  \item \textbf{Curvature-signature diagnostics from intervention-response geometry:}
  We compute turning-angle curvature changes in logit space as a complementary \emph{diagnostic} signal and introduce \emph{curvature signatures} that summarize how masking a chunk changes turning angles across pivots; cosine similarity between signatures defines curvature similarity between chunks.
  These signatures induce diagnostic \emph{curvature roles} that group chunks by shared intervention-response geometry; in qualitative inspection, roles often align with semantically coherent trace segments, enabling compact pivot-aligned \emph{role paths}.
\end{itemize}

\paragraph{Diagnostics and artifacts for verification.}
To enable reviewer-auditable inspection, each run exports full provenance
artifacts grounded in the model’s \emph{exact tokenized input}.
These include:
(i) per-example summary tables with pivot statistics and attribution aggregates,
(ii) per-configuration ranked chunk lists with stable chunk identifiers,
(iii) integrity-checked mappings between chunk indices and token spans,
and (iv) a self-contained interactive HTML reviewer interface that links
raw text, pivot locations, masking interventions, and trajectory
visualization (Figure~\ref{fig:drtc_ui_demo}). All reported DRTC and curvature quantities are reproducible directly from the exported JSON bundles without re-running model inference.
This design allows reviewers to verify pivot discovery, intervention
construction, per-pivot effects $(\delta_{k,i}, w_{k,i})$, and final
aggregation mechanics step-by-step on the exact realized rollout. In practice, the signed per-chunk scores are sharply concentrated, yielding a compact, auditable shortlist of context segments to inspect within long traces and a concrete, mechanically grounded starting point for subsequent circuit-level or mechanistic analyses.

\section{Related work}

Recent interpretability work has shifted from single-pass explanations toward causal analysis of multi-step reasoning traces. \citet{zhao2026ahamomentsfakeidentifying} formalize \emph{decorative thinking} and propose True Thinking Score (TTS), estimating step-level causal contribution via stochastic perturbation and resampling; \citet{macar2025thoughtbranchesinterpretingllm} similarly use on-policy resampling to study semantic resilience across trajectories. Complementary lines of work model influence across autoregressive generations with attribution graphs \citep{walker2025explainingreasoninglargelanguage}, analyze reasoning as a geometric trajectory in representation space \citep{zhou2025geometryreasoningflowinglogics,manson2025curvedinferenceconcernsensitivegeometry}, and study logit/layer-wise dynamics for interpretability and decoding \citep{he2025deltadecodingstrategybased,yan2025additionmovementsmappinglayerwise}.

DRTC targets which earlier context chunks \emph{steer} a single realized long-horizon trace under on-policy generation, using pivot-local deterministic receiver-side interventions and a directional target in log-probability space (curvature is diagnostic only). We benchmark against chunk-level attribution baselines at matching granularity on the same fixed rollout: representation erasure/occlusion \citep{li2017understandingneuralnetworksrepresentation}, integrated gradients (grad$\times$input surrogate) \citep{sundararajan2017axiomaticattributiondeepnetworks}, optimized smooth masking \citep{Fong_2017}, and activation patching \citep{zhang2024bestpracticesactivationpatching}. Circuit-discovery methods (e.g., CD-T, path patching, ACDC/EAP) are complementary and operate at internal-mechanism granularity; DRTC can supply candidate chunks/pivots to seed such analyses. Resampling-based step-faithfulness methods address necessity under alternate traces, whereas DRTC uses no-resampling pivot-local counterfactuals; head-to-head resampling comparisons are left to future work. A detailed comparison appears in Appendix~\ref{app:related_work_expanded} (Table~\ref{tab:related_work_comparison}).

\section{Methodology}

\subsection{Overview}
\begin{figure*}[t]
  \centering
  \includegraphics[width=\textwidth]{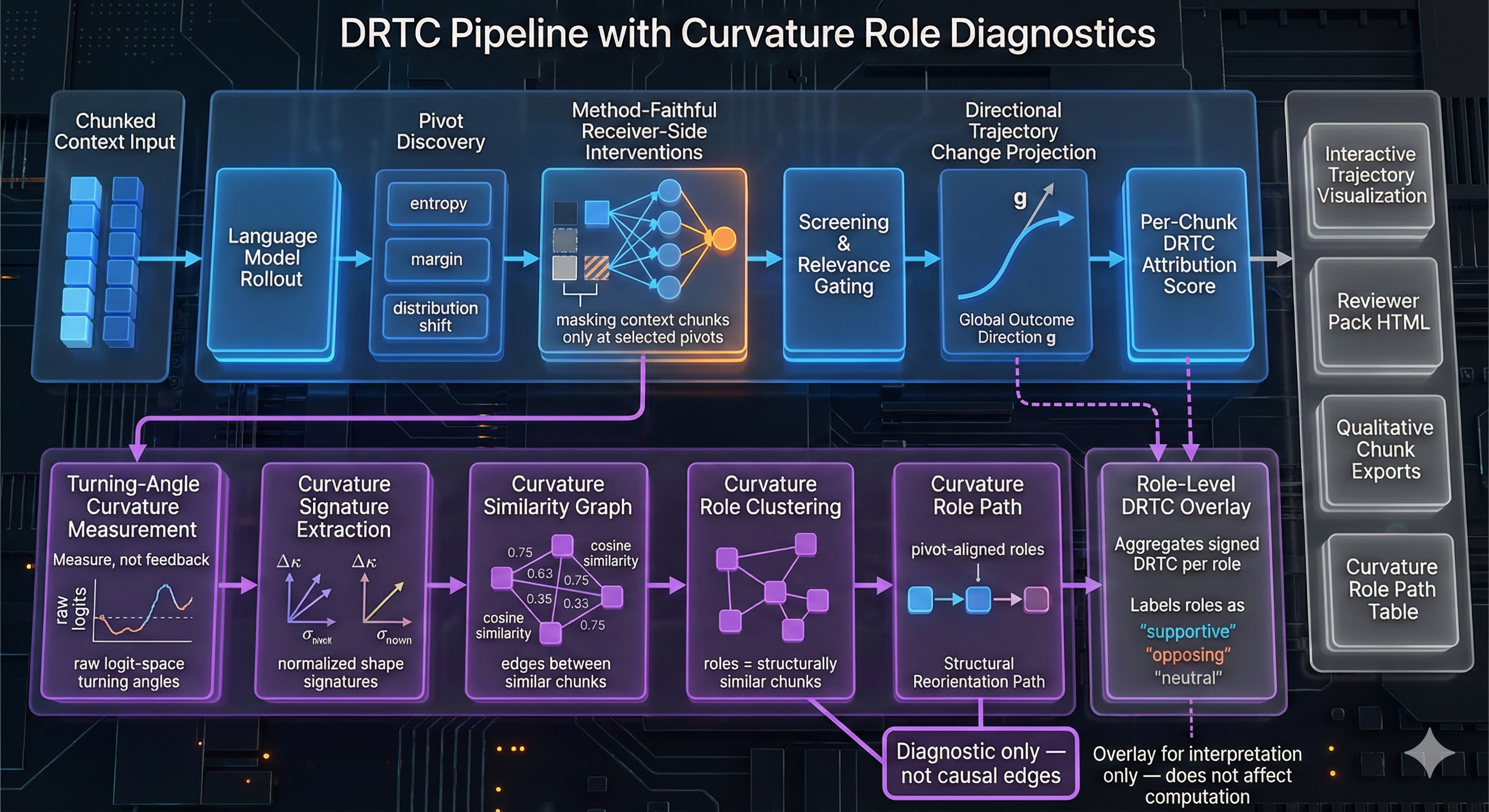}
  \caption{\textbf{DRTC pipeline overview.} Curvature is diagnostic only and is not used to define pivots or scores.}
  \label{fig:drtc_overview}
\end{figure*}

DRTC attributes long-horizon reasoning by combining (i) \emph{pivot discovery} to localize consequential decision points in a single on-policy rollout, (ii) \emph{receiver-side} interventions that block information from specific earlier chunks \emph{only at those pivots} while preserving the realized generated text, and (iii) a \emph{directional} causal target that measures whether an intervention redirects the model's log-probability trajectory toward or away from the realized rollout direction.
We also compute turning-angle curvature in logit space as a complementary \emph{geometric diagnostic}.
Curvature is never used to select pivots or define DRTC scores.

\subsection{Problem setup}

Given a prompt, the model produces a single on-policy autoregressive reasoning rollout with tokens $\{y_t\}_{t=1}^T$.
We segment the rollout text into $N$ contiguous fixed-stride chunks $\{c_1,\dots,c_N\}$ for attribution; prompt text and ground-truth solutions are retained for auditability but excluded from pivot discovery, interventions, and scoring. We use fixed-stride token chunks (16 tokens) rather than sentence segmentation to keep interventions length-controlled and reproducible under tokenization, avoid confounding ablation effect size with variable-length units, and keep long-horizon attribution tractable \citep{kiss-strunk-2006-unsupervised,fong2019understandingdeepnetworksextremal,deyoung2020eraserbenchmarkevaluaterationalized,pan2026longhorizoninterpretabilityefficientfaithful}.

At step $t$, the model defines a distribution $p_t(\cdot)$ over the vocabulary $\mathcal{V}$ with logits $z_t\in\mathbb{R}^{|\mathcal{V}|}$.
We perform directional comparisons in log-probability space,
\begin{equation}
s_t \;=\; \log p_t(\cdot) \;=\; \log\softmax(z_t)\in\mathbb{R}^{|\mathcal{V}|},
\end{equation}
and compute curvature diagnostics separately in raw logit space.
Our objective is to identify which chunks causally influence key decision points and whether that influence steers the reasoning trajectory along the realized rollout.

\subsection{Pivot discovery and weighting}

DRTC identifies a small set of pivot positions where the model is likely to commit, revise, or redirect its reasoning.
For each position $t$ we compute an uncertainty/shift-based pivot score
\begin{equation}
\tilde{u}_t
=
\gamma_H H_t
+
\gamma_M \bigl(1-\mathrm{margin}_t\bigr)
+
\gamma_S S_t ,
\end{equation}
where $H_t$ is entropy, $\mathrm{margin}_t$ is the top-2 probability margin, and $S_t$ is a local distribution-shift signal measured by Jensen--Shannon divergence between token windows before and after $t$.
We use $(\gamma_H,\gamma_M,\gamma_S)=(1.0,1.0,0.5)$.
We compute $S_t$ as the Jensen--Shannon divergence between the \emph{mean next-token distributions} (averaged softmax probabilities over $\mathcal{V}$) in a $W$-token window before $t$ (indices $[t\!-\!W, t)$) and a $W$-token window after $t$ (indices $(t, t\!+\!W]$), excluding position $t$ from both windows (default $W=5$).

We select the top $K$ pivots $\{\tau_k\}_{k=1}^K$ by $\tilde{u}_t$ subject to spacing constraints and assign each pivot a normalized importance weight via a softmax:
\begin{equation}
u_k \;=\; \frac{\exp(\tilde{u}_{\tau_k})}{\sum_{j=1}^{K}\exp(\tilde{u}_{\tau_j})},
\qquad \sum_{k=1}^{K} u_k = 1.
\end{equation}
Uniform pivot weights are used as an ablation.

\noindent\textbf{Defaults.} Unless otherwise stated, we use $K=8$ pivots, enforce a minimum pivot spacing of 4 tokens to avoid near-duplicate pivots, and compute the JSD shift signal using 5-token windows before and after $t$; sensitivity to these choices is reported in Appendix~\ref{app:sensitivity}.

\subsection{Receiver-side interventions and relevance screening}

At each pivot $\tau_k$, we test whether an earlier chunk $c_i$ is causally relevant by applying a \emph{receiver-side} intervention that blocks information flow from $c_i$ \emph{only at $\tau_k$}, while keeping the realized rollout fixed.
Concretely, we evaluate the model under teacher forcing on the realized prefix and compute the pivot logits at position $\tau_k$ without resampling any continuation tokens.
The intervention modifies only the pivot's attention connectivity: it prevents the pivot query at $\tau_k$ from attending to key/value positions inside chunk $c_i$, leaving all token identities, embeddings, and the rest of the computation unchanged. This intervention blocks \emph{direct} query$\rightarrow$key access from $c_i$ to the pivot at $\tau_k$, but it does not erase information already integrated into upstream hidden states, so it estimates a pivot-local counterfactual rather than full causal mediation through the entire prefix.

\paragraph{Attention-edge masking at a pivot.}
Let chunk $c_i$ correspond to token indices $\mathcal{I}_i \subset \{1,\dots,\tau_k-1\}$ in the realized prefix.
For each transformer layer $\ell$ and query head $h$, let $s^{(\ell,h)}_{\tau_k,j}$ denote the pre-softmax attention score from query position $\tau_k$ to key position $j$.
Our receiver-side mask sets
\begin{equation}
s^{(\ell,h)}_{\tau_k,j} \leftarrow -\infty \quad \text{for all } j \in \mathcal{I}_i,
\end{equation}
and leaves all other scores unchanged, so that attention mass from $\tau_k$ to $c_i$ is exactly zero after the softmax.
We apply this mask at every layer and across all query heads.\footnote{
Under grouped/multi-query attention (GQA/MQA), keys/values may be shared across head groups; our masking operates on the query$\rightarrow$key score tensor, so it is applied to every query head regardless of key/value sharing.
}
This edge-level intervention is closely related to causal tracing and activation/patching-style probes in mechanistic interpretability \citep{geva2023dissectingrecallfactualassociations,conmy2023automatedcircuitdiscoverymechanistic,zhang2024bestpracticesactivationpatching}.

\paragraph{Pivot-local screening effect.}
Let $\ell_t(\cdot)$ denote pre-softmax logits at position $t$.
We define a pivot-local screening effect on the baseline top token $y_{\tau_k}^\star$:
\begin{equation}
a_{k,i}
=
\ell_{\tau_k}(y_{\tau_k}^\star)
-
\ell_{\tau_k}^{(-i,k)}(y_{\tau_k}^\star),
\end{equation}
where $( -i,k )$ denotes masking chunk $c_i$ \emph{only at} pivot $\tau_k$ as above.
Intuitively, $a_{k,i}$ measures whether removing access to $c_i$ weakens the pivot decision in the realized rollout context.

\paragraph{Relevance gate.}
We map screening effects to a bounded relevance weight
\begin{equation}
w_{k,i} \;=\; \sigma\!\left(\beta_k \,\hat{a}_{k,i}\right)\in[0,1],
\end{equation}
where $\hat{a}_{k,i}$ is a robustly scaled version of $a_{k,i}$ (e.g., median/MAD robust $z$-scoring with clipping) and $\beta_k$ is a pivot-specific calibration parameter chosen from a robust statistic of nonzero screening magnitudes at pivot $\tau_k$.
The gate downweights chunk--pivot pairs that do not reliably affect the pivot decision while preserving ordering among stronger effects.

\subsection{Directional trajectory change and DRTC aggregation}

We model reasoning as a trajectory through log-probability space and measure whether an intervention redirects that trajectory along the realized rollout direction.
Let $s_t$ and $s_t^{(-i,k)}$ denote the baseline and intervened log-probability vectors.
We define a global rollout direction using the endpoints of the pivot sequence:
\begin{equation}
g \;=\; \frac{s_{\tau_K}-s_{\tau_1}}{\|s_{\tau_K}-s_{\tau_1}\|}.
\end{equation}
For each pivot $\tau_k$ and chunk $c_i$, define the pivot-local trajectory effect
\begin{equation}
e_{k,i} \;=\; s_{\tau_k}-s_{\tau_k}^{(-i,k)},
\end{equation}
and its directional component
\begin{equation}
\delta_{k,i} \;=\; \langle e_{k,i}, g\rangle .
\end{equation}
\noindent\textbf{Pivot-local readout.} Although $g$ is defined globally from the pivot endpoints, the intervention effect is read out \emph{locally at each pivot} via $e_{k,i}=s_{\tau_k}-s_{\tau_k}^{(-i,k)}$ and $\delta_{k,i}$; DRTC does not rely on a final-pivot-only score. Alternative definitions of $g$ and meander robustness are reported in Appendix~\ref{app:g_robustness}.
Positive $\delta_{k,i}$ indicates that chunk $c_i$ supports the realized rollout direction at pivot $k$.
For stability, we robustly normalize $\delta_{k,i}$ within each pivot (median/MAD scaling with clipping; sensitivity and implementation details are in Appendix~\ref{app:sensitivity}).

Finally, we aggregate across pivots using pivot weights and relevance gates:
\begin{equation}
\mathrm{DRTC}(i)
=
\sum_{k=1}^{K} u_k \, w_{k,i} \, \delta_{k,i}.
\end{equation}
We treat $\mathrm{DRTC}(i)$ as the primary attribution score; auxiliary diagnostic decompositions and sanity checks, including pivot-weight ablations and per-component analyses, are reported in the Appendix (Appendix~\ref{app:uk} and Appendix~\ref{app:sensitivity}).

\subsection{Curvature diagnostics and role paths}
\label{subsec:curvature_roles}

In addition to directional alignment, we compute a geometric \emph{diagnostic} of how interventions change local turning behavior in raw logit space.
For consecutive logits $(z_{t-1},z_t,z_{t+1})$, define
\begin{equation}
v_{\mathrm{prev}} = z_t - z_{t-1},\qquad
v_{\mathrm{next}} = z_{t+1} - z_t,\qquad
\kappa(z_{t-1},z_t,z_{t+1})
=
\arccos\!\left(
\frac{\langle v_{\mathrm{prev}}, v_{\mathrm{next}}\rangle}
{\|v_{\mathrm{prev}}\|\,\|v_{\mathrm{next}}\|}
\right),
\end{equation}
with standard clipping and small-$\varepsilon$ stabilization in implementation.
For each pivot $\tau_k$ and chunk $c_i$, we measure the intervention-induced curvature change
\begin{equation}
\Delta\kappa_{k,i}
=
\kappa(z_{\tau_k-1}, z_{\tau_k}^{(-i,k)}, z_{\tau_k+1})
-
\kappa(z_{\tau_k-1}, z_{\tau_k}, z_{\tau_k+1}),
\qquad
\mathrm{CurvImpact}(i) \;=\; \sum_{k=1}^{K} u_k \,\Delta\kappa_{k,i}.
\end{equation}
$\mathrm{CurvImpact}$ is diagnostic only: it summarizes intervention-response geometry and is not used as a causal attribution score.

\paragraph{Curvature signatures and similarity.}
We summarize curvature response patterns with a normalized signature
\[
s_i \;=\; \mathrm{normalize}\!\left([\Delta\kappa_{1,i},\dots,\Delta\kappa_{K,i}]\right),
\]
and define curvature similarity between chunks $i$ and $j$ as $\langle s_i,s_j\rangle$.

\paragraph{Roles and curvature role paths.}
To compactly summarize long rollouts, we form an undirected graph over chunks where an edge connects $(i,j)$ if $\langle s_i,s_j\rangle$ exceeds a fixed threshold; connected components define curvature \emph{roles} $r$.
Roles are purely structural groupings defined by curvature signatures and do not depend on DRTC or screening.
For each pivot $\tau_k$, we score the reorientation attributed to role $r$ by
\[
\mathrm{RoleImpact}(r,k) \;=\; u_k \sum_{c_i\in r} |\Delta\kappa_{k,i}|,
\]
select the maximizing role $r_k^\star=\arg\max_r \mathrm{RoleImpact}(r,k)$, and choose a representative chunk
$i_k^\star=\arg\max_{c_i\in r_k^\star} |\Delta\kappa_{k,i}|$.
Ordering the dominant roles $(r_k^\star)_{k=1}^K$ by pivot index yields a pivot-aligned \emph{curvature role path}. Optionally, we annotate each role with aggregated directional attribution
\[
\mathrm{DRTC}(r) \;=\; \sum_{c_i\in r} \mathrm{DRTC}(i),
\]
as a descriptive overlay (it does not affect role construction or pivot assignment).
Overall, curvature role paths provide a compact synopsis of intervention-induced reorientation types that complements chunk-level DRTC attribution.

\section{Results}
\label{sec:results}

\paragraph{Setup.}
We evaluate DRTC on a fixed 24-example slice from \texttt{math-rollouts} \citep{bogdan2025thoughtanchorsllmreasoning}.
For each example, the model generates a greedy on-policy reasoning rollout (fixed seeds), segmented into fixed-stride chunks for attribution; prompt text and ground truth are retained for auditability but excluded from pivot discovery, interventions, and scoring.
Unless otherwise stated, we use $K=8$ pivots per example.
We report results for four open-weight reasoning models---\textbf{LFM2.5-1.2B-Thinking}, \textbf{Ministral-3B-Reasoning}, \textbf{Phi-4-Mini-Reasoning}, and \textbf{R1-Distill-Qwen-1.5B} \citep{liquidai2025lfm2,liu2026ministral,peng2025phi4mini,deepseek2025r1}---evaluated on the same examples with identical decoding settings. We compare \textbf{C0} (DRTC only), \textbf{C8} (DRTC + curvature diagnostics), and \textbf{C9} (matched random-span control with identical masking and scoring).
On \texttt{R1-Distill-Qwen-1.5B} with $K=8$ on the 24-example slice, median wall time is 12.28s per example with 4864MB peak GPU memory (Appendix~\ref{app:cost}).

\subsection{Core findings: invariance, concentration, and falsification}
\label{subsec:core_results}

\begin{figure}[H]
  \centering
  \includegraphics[width=0.4\linewidth]{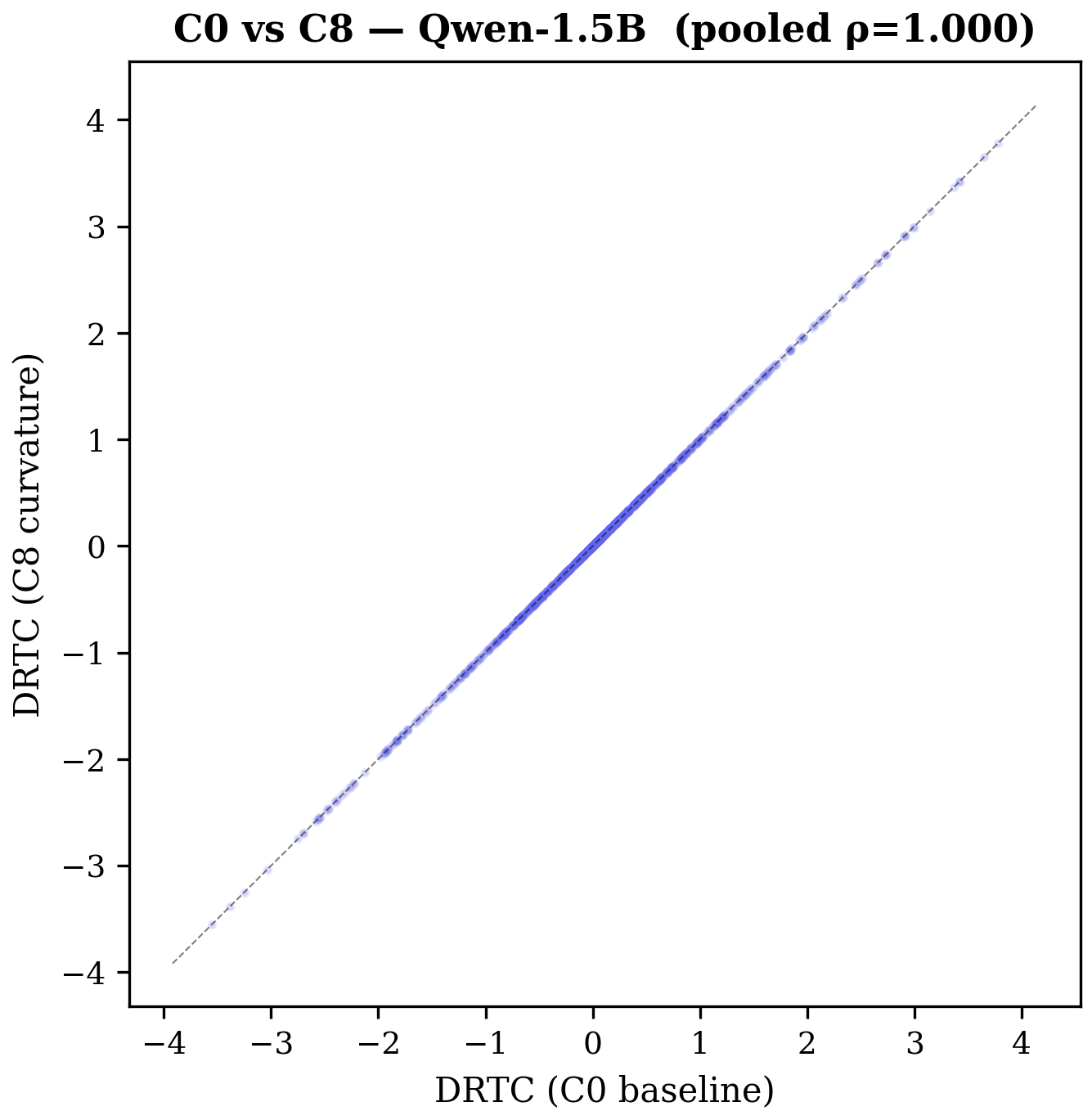}
  \caption{
  Curvature invariance under diagnostic logging (representative model: \texttt{R1-Distill-Qwen-1.5B}).
  Per-chunk DRTC scores from C0 and C8 lie on the identity line ($\rho=1.000$), confirming curvature computation is purely diagnostic.
  }
  \label{fig:c0_c8_identity_r1}
\end{figure}

\paragraph{Curvature logging is strictly diagnostic.}
C0 and C8 differ only in whether curvature diagnostics are computed and logged; curvature is never used in pivot discovery, intervention construction, gating, or DRTC aggregation.
Across all four models (24 examples each), per-example Spearman correlations between C0 and C8 chunk rankings are identically $\rho=1.000$ (median $1.000$, 95\% CI $[1.000,1.000]$).
Figure~\ref{fig:c0_c8_identity_r1} shows a representative identity scatter for \texttt{R1-Distill-Qwen-1.5B}; full invariance diagnostics appear in Appendix~\ref{app:c0_c8_invariance}.

\paragraph{DRTC yields a compact shortlist of influential chunks.}
Across models, directional influence is concentrated rather than diffuse:
using per-example normalized $|\mathrm{DRTC}|$ mass shares, median Gini coefficients range from $0.50$--$0.58$ and the top $5\%$ of chunks account for roughly $0.23$--$0.28$ of total influence.
Cross-model histograms, concentration curves, and bootstrap confidence intervals are in Appendix~\ref{app:sparsity}.

\paragraph{Random-span falsification: learned pivots induce stronger interventions.}
As a falsification signal, we compare learned pivots (C8) to matched random spans (C9) under identical masking and scoring.
Across all four models, C8 exhibits strictly higher median per-example mean pivot-local intervention magnitude $\mathbb{E}_k[|\delta_{k,i}|]$ than C9, with median gaps (C8$-$C9) ranging from $+0.039$ to $+0.178$ (95\% bootstrap CIs in Figure~\ref{fig:c8_vs_c9_magnitude}); per-example diagnostics are in Appendix~\ref{app:falsification}.

\begin{figure}[H]
  \centering
  \includegraphics[width=0.6\linewidth]{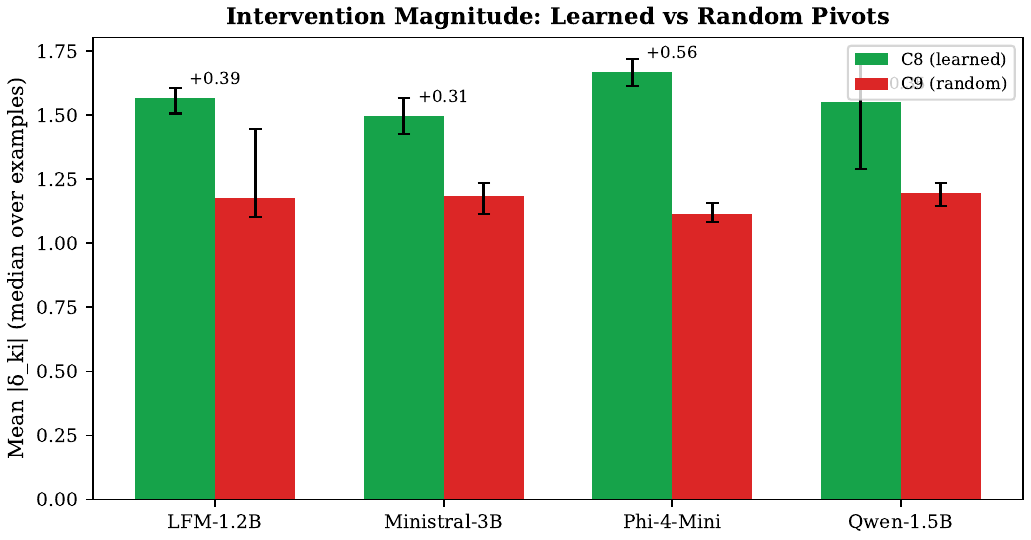}
  \caption{
  Cross-model comparison of median per-example mean pivot-local intervention magnitude.
  Learned pivots (C8) induce stronger interventions than matched random spans (C9) across all four models.
  Error bars denote 95\% bootstrap confidence intervals.
  }
  \label{fig:c8_vs_c9_magnitude}
\end{figure}

\paragraph{Pivot weighting is non-redundant.}
Recomputing DRTC \emph{post hoc} with uniform pivot weights ($u_k=1/K$) preserves the dominant structure but induces non-trivial rank reshuffling: weighted-vs-uniform Spearman correlations remain high ($\rho \approx 0.97$--$0.99$) but strictly below $1.000$ across all models.
Full ablations and scatter plots are in Appendix~\ref{app:uk}.

\subsection{Robustness, baselines, curvature calibration, and outcome linkage}
\label{subsec:robustness_bundle}

We summarize additional quantitative evidence here and provide full tables/figures in the Appendix.

\paragraph{Robustness to direction estimation and heuristic knobs.}
Chunk rankings are stable to alternate direction-estimation strategies for $g$ (Appendix~\ref{app:g_robustness}).
A targeted sensitivity sweep shows near-identity agreement under $\beta$-gate rescaling (median Spearman vs.\ baseline $=0.9901$ at $\beta$-scale $0.5$; $0.9892$ at $\beta$-scale $2.0$), while varying the pivot budget is more consequential than $\beta$ rescaling (median Spearman vs.\ baseline $=0.3376$ at $K=4$ and $0.7298$ at $K=12$; baseline $K=8$). Additional knob sweeps and concentration metrics are reported in Appendix~\ref{app:sensitivity}.

\paragraph{Intervention specification and masking ablations.}
Receiver-side attention-edge masking is applied at the pivot across all layers and query heads; ablations over head subsets (and related implementation details, including GQA/MQA handling) are reported in Appendix~\ref{app:masking_ablation}.

\paragraph{Comparisons to attribution baselines.}
We compare DRTC to practical chunk-level attribution baselines evaluated on the same realized rollouts (Appendix~\ref{app:baselines}). Agreement is moderate for gradient- and optimized-perturbation methods (median Spearman $\rho=0.53$ for grad$\times$input; $\rho=0.45$ for smooth masking), weaker for occlusion ($\rho=0.30$), and near-zero for our activation patching instantiation ($\rho=0.04$). Top-$k$ overlaps and per-example agreement distributions are reported in Appendix~\ref{app:baselines}.

\paragraph{Curvature calibration.}
Curvature is used only diagnostically: it characterizes the \emph{geometry} of intervention response rather than serving as an attribution or outcome-causal score.
Within traces, curvature-impact and DRTC can co-localize, while cross-example aggregates can be small.
A calibration and reconciliation of within-trace vs.\ cross-example statistics is provided in Appendix~\ref{app:curvature_calibration}, with additional cross-model diagnostic visualizations in Appendix~\ref{app:curvature_diag}.

\paragraph{Outcome linkage (U1b).}
Using graded embedding-interpolation interventions and an outcome-facing metric (teacher-forced $\Delta\log p(\text{gold})$), top-ranked DRTC chunks degrade gold-answer log-probability more than strict quartile-matched random controls on a stability-filtered subset. In the stable subset ($|\mathrm{orig\_margin}|>1.0$), the best regime (cs$=4$, $\alpha=0.2$) achieves median paired difference $0.3398$ with 95\% bootstrap CI $[0.0269, 1.1851]$ and sign fraction $74\%$ (Appendix~\ref{app:outcome_u1b}). We report outcome linkage under greedy decoding with teacher-forced $\Delta\log p(\text{gold})$; robustness across sampling temperatures/decoding regimes and whether end-to-end answer flips track DRTC ranks are left for future work.

\subsection{Scaling study on MATH (R1-Distill-Qwen-1.5B)}
\label{subsec:math_scale_r1}

To test robustness beyond the 24-example slice, we run DRTC on $N=500$ MATH problems \citep{hendrycks2021measuringmathematicalproblemsolving} with \texttt{R1-Distill-Qwen-1.5B} (greedy decoding, fixed seeds, $K=8$; median trace length 2694 tokens, IQR [2131, 3068]).
Compute scales with the number of pivots and evaluated chunks per example; practical cost and memory profiles are summarized in Appendix~\ref{app:cost}.
The primary falsification gap persists at scale: learned spans (C8) produce stronger pivot-local interventions than matched random spans (C9), with median $\Delta=\mathbb{E}_k[|\delta|]_{\textsc{C8}}-\mathbb{E}_k[|\delta|]_{\textsc{C9}}=0.409$ (95\% CI [0.354, 0.453]) and $\Delta>0$ for 355/500 examples (sign test $p=2.3\times 10^{-21}$).
Attribution remains moderately concentrated (median Gini 0.586; median top-1 mass 0.089; median top-5\% mass 0.209).
Additional scaling statistics and curvature calibration details are in Appendix~\ref{app:falsification} and Appendix~\ref{app:curvature_calibration}, with further diagnostic visualizations in Appendix~\ref{app:curvature_diag}.

\subsection{Qualitative interpretation}
\label{subsec:qualitative_interpretation}

Across models, top-magnitude DRTC chunks typically correspond to strategy-setting constraints or key structural commitments (e.g., introducing an invariant or enforcing a structural condition), while opposing-sign chunks often align with early orientation text or exploratory detours that are later revised.
Curvature magnitude frequently coincides with visible phase transitions (e.g., restatement $\rightarrow$ derivation), providing a complementary view of reorientation intensity.
Detailed case studies and exported reviewer-auditable artifacts appear in Appendix~\ref{app:qual_case_studies}.

\subsection{Summary: DRTC as a process-level steering map}
\label{subsec:results_summary}

Taken together, the results support DRTC as a \emph{process-level steering map} for a single realized rollout:
curvature logging is strictly diagnostic (Section~\ref{subsec:core_results}), influence is concentrated into an auditable shortlist (Appendix~\ref{app:sparsity}), learned pivots outperform matched random spans under identical logic (Figure~\ref{fig:c8_vs_c9_magnitude}; Appendix~\ref{app:falsification}), and pivot weighting is non-redundant (Appendix~\ref{app:uk}).
Robustness diagnostics, baseline comparisons, curvature calibration/diagnostics, and graded outcome linkage are provided in Appendix~\ref{app:g_robustness}--\ref{app:outcome_u1b} (see also Appendix~\ref{app:curvature_diag}).

\section{Conclusion, limitations, and future work}
\label{sec:conclusion}

\paragraph{Limitations.}
DRTC primarily measures directional redirection at pivot points; while graded interventions show outcome-facing sensitivity using teacher-forced $\Delta\log p(\text{gold})$ under position-matched controls on a stability-filtered subset, such effects do not guarantee end-to-end correctness changes, and flip-based metrics can saturate under strong edits. Attribution currently uses fixed-stride chunks and a fixed pivot budget ($K=8$), which stabilizes comparisons but may miss the appropriate granularity of reasoning; adaptive pivots and variable-resolution or hierarchical chunking are natural extensions. Curvature is used strictly as a logit-space geometric descriptor and does not identify the circuits implementing reorientation, motivating integration with mechanistic tracing. Finally, broader evaluation across domains and decoding strategies, together with stronger outcome-level benchmarks, is needed to clarify when local redirections accumulate into correctness changes.

\paragraph{Conclusion.}
We introduced Directional Reasoning Trajectory Change (DRTC), a process-causal interpretability framework for long-horizon reasoning under on-policy autoregressive generation. DRTC localizes pivot decision points and applies receiver-side interventions that block information flow from selected earlier chunks only at those pivots, yielding signed per-chunk steering scores defined by directional changes in log-probability space. Across four reasoning models, attribution is consistently sparse and structured, learned pivots induce stronger intervention magnitudes than matched random spans, and curvature provides complementary geometric diagnostics without altering DRTC scores. Overall, DRTC provides a trajectory-level account of how specific context elements steer reasoning, and graded outcome linkage indicates that top-ranked chunks can have measurable outcome-facing effects (teacher-forced $\Delta\log p(\text{gold})$) under position-matched controls, though broader domains, decoding regimes, and stronger end-task evaluations remain future work.

\section*{Ethics statement}
This work does not introduce new model architectures or collect new datasets. We propose an analysis framework for interpreting existing autoregressive reasoning models via pivot-local, receiver-side input attribution. Consistent with the goals of explainable AI, the method is intended to increase transparency and support auditing of model behavior. As such, it does not introduce material ethical risks beyond those inherent to the deployment and use of the underlying pretrained models and their generated outputs. We evaluate only on publicly available problem statements and do not use personally identifiable information.

\section*{Reproducibility statement}
We describe the full methodology in the paper, including pivot discovery signals, intervention definition, scoring, and aggregation, along with model and decoding settings. Our experiments use fixed seeds and export complete per-example artifacts (pivot locations, chunk boundaries/identifiers, intervention effects, attribution scores, and outcome-linkage summaries) to enable independent auditing. The code and artifact bundle used to generate the tables and figures in this submission are provided as supplementary material; we will publicly release the full implementation and evaluation code, including scripts to reproduce all figures and tables, as well as the self-contained interactive reviewer UI used for qualitative inspection.

\clearpage
\bibliographystyle{plainnat}
\bibliography{references}

\clearpage
\appendix
\section*{Appendix}
\addcontentsline{toc}{section}{Appendix}

\section{Additional quantitative evidence}
\label{app:quant_evidence}

\subsection{C0 vs.\ C8 invariance diagnostics}
\label{app:c0_c8_invariance}

C0 (baseline) and C8 (curvature-enabled) differ only in whether curvature
diagnostics are computed and logged.
Curvature is not used in pivot discovery, receiver-side masking,
relevance gating, or DRTC aggregation.
We therefore test directly whether enabling curvature logging changes
any attribution results.

\paragraph{Per-example rank invariance.}
For each model and each example (24 per model), we compute the Spearman
rank correlation between per-chunk DRTC scores produced under C0 and C8.
Across all four models, every example yields
\[
\rho = 1.000.
\]
The median Spearman correlation is $1.000$ with 95\% bootstrap confidence
interval $[1.000, 1.000]$ for each model.

\paragraph{Score-level identity.}
Beyond rank invariance, we verify score-level equality.
For every example across all models, the maximum absolute difference
between corresponding C0 and C8 DRTC scores is zero within floating-point
tolerance.
Thus, enabling curvature logging does not alter attribution magnitudes.

\begin{figure*}[h]
  \centering
  \includegraphics[width=0.28\textwidth]{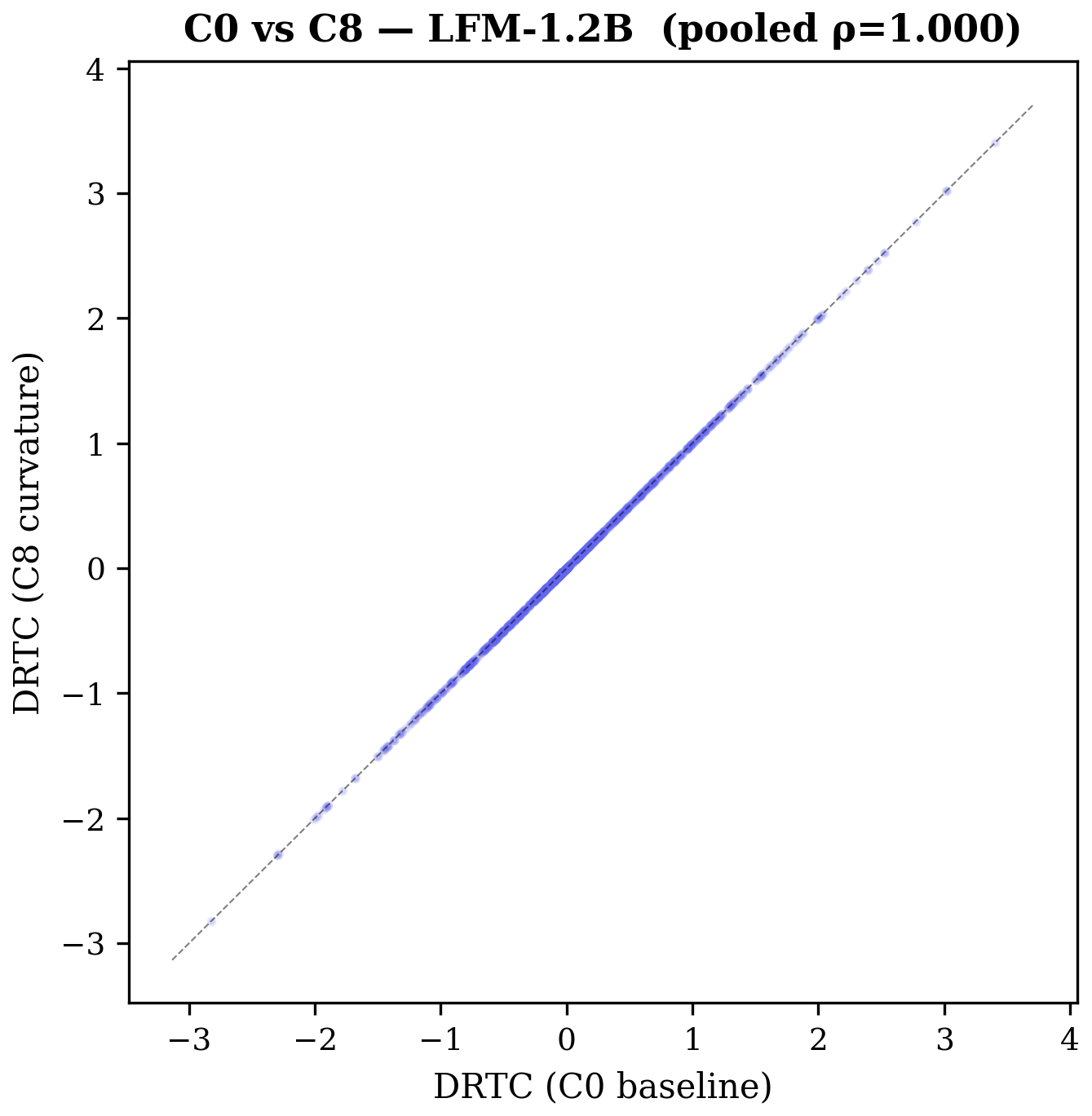}
  \includegraphics[width=0.28\textwidth]{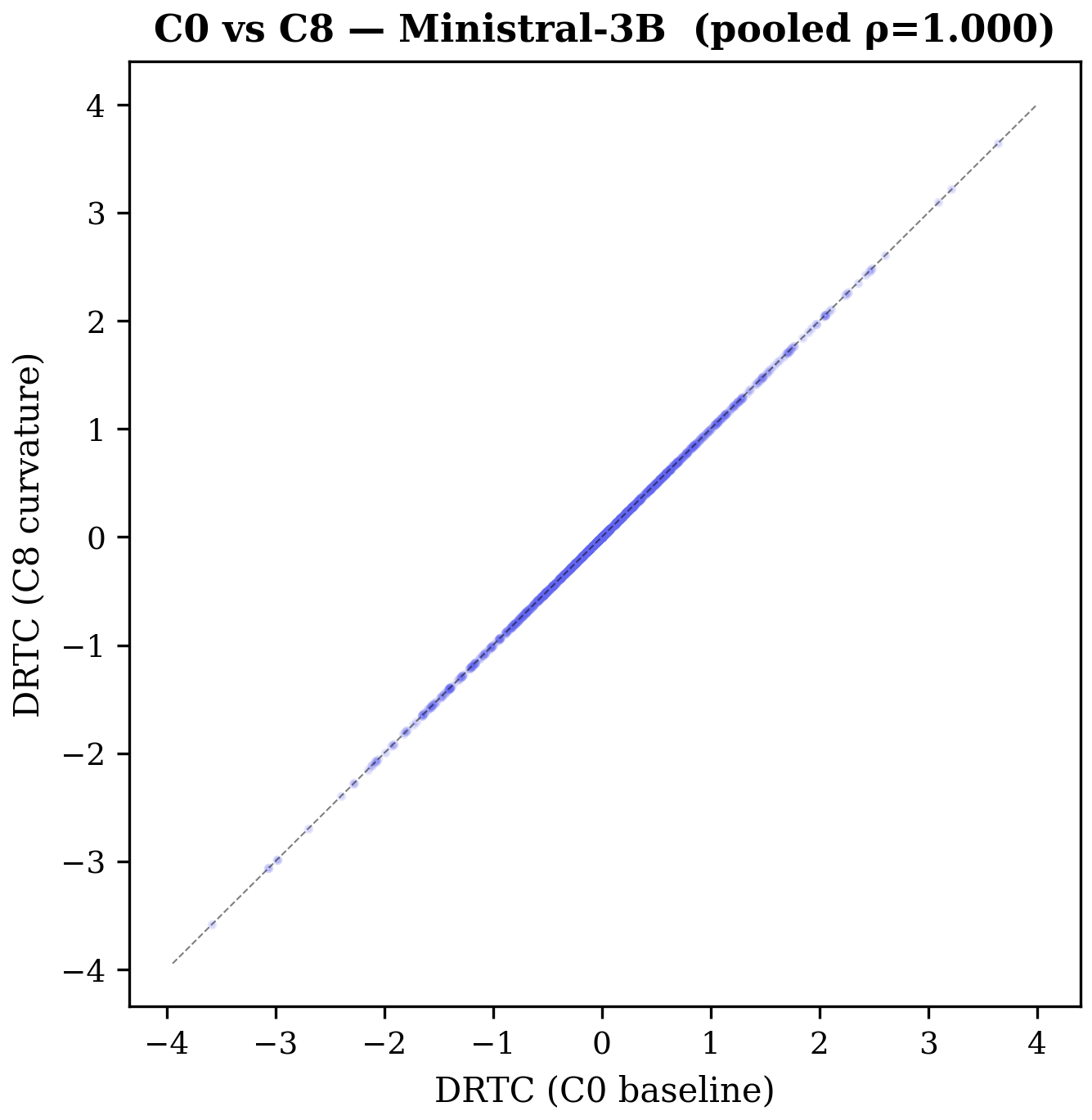}

  \vspace{6pt}

  \includegraphics[width=0.28\textwidth]{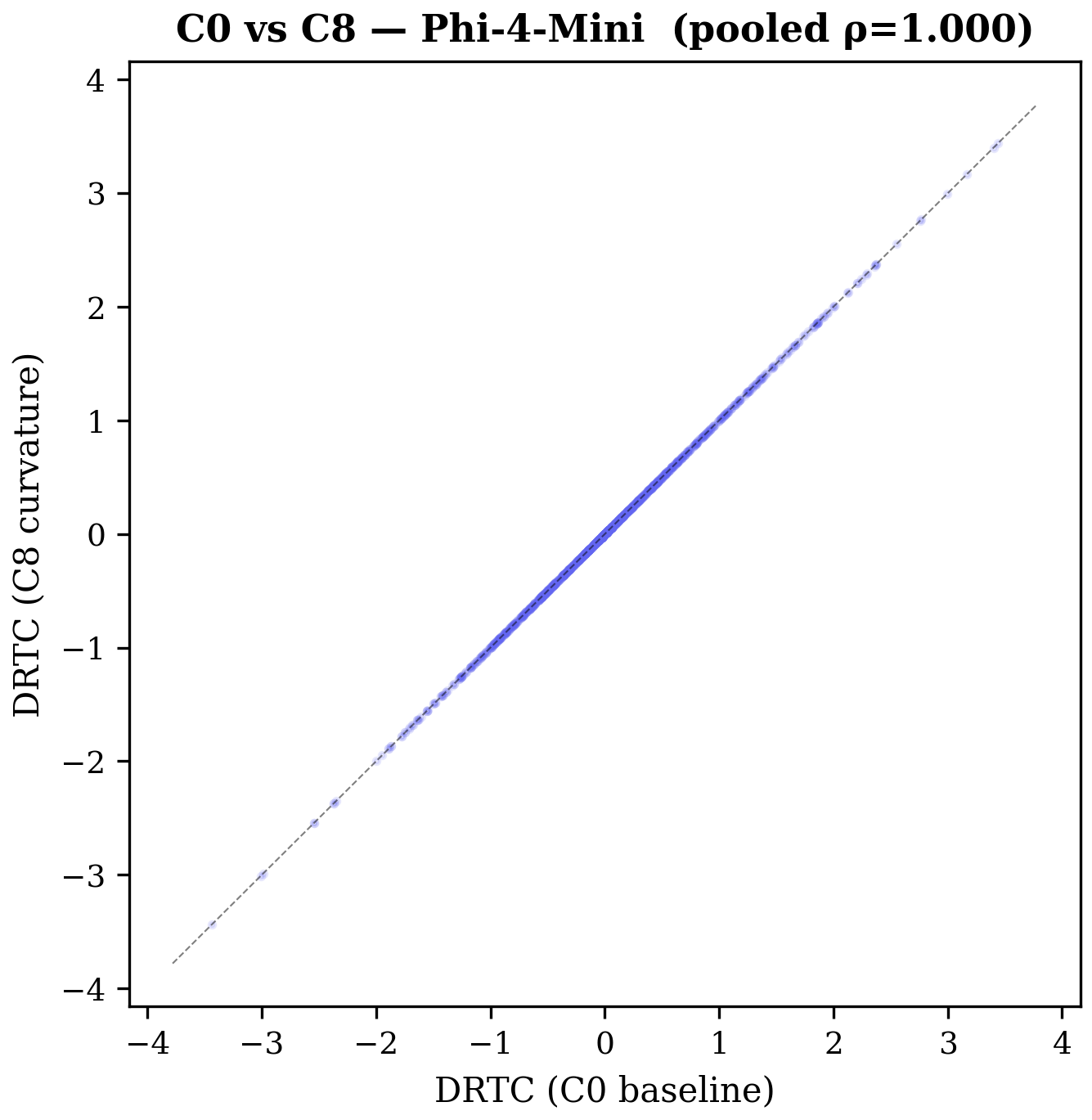}
  \includegraphics[width=0.28\textwidth]{_posthoc_analysis/analysis_figures/math-problems_r1-distill-qwen-1.5b/c0_vs_c8_scatter.png}

  \caption{
  C0 vs.\ C8 invariance across models.
  Each panel shows per-chunk DRTC scores under C0 (baseline)
  versus C8 (curvature-enabled).
  Points lie exactly on the identity line, confirming that curvature
  computation is strictly diagnostic and does not alter attribution.
  }
  \label{fig:c0_c8_invariance_all}
\end{figure*}

\paragraph{Visual confirmation.}
Figure~\ref{fig:c0_c8_invariance_all} shows representative per-model
scatter plots of C0 versus C8 DRTC scores.
All points lie exactly on the identity line.
Corresponding per-example histograms of Spearman correlations
are degenerate at $\rho=1$ for every model.

\paragraph{Conclusion.}
These results establish exact attribution invariance between C0 and C8.
Curvature computation is strictly diagnostic and does not influence
pivot selection, intervention construction, gating, aggregation,
or final DRTC rankings.

\subsection{Robustness to direction estimation and trajectory meander}
\label{app:g_robustness}

We evaluate robustness to the global direction estimate $g$ by recomputing chunk rankings under multiple
direction-estimation strategies (endpoint difference, PCA, robust PCA, piecewise, piecewise-signed) and
meander diagnostics. We summarize per-example Spearman correlations between rankings and report the range
of meander ratios observed.

\begin{figure}[H]
  \centering
  \includegraphics[width=0.65\linewidth]{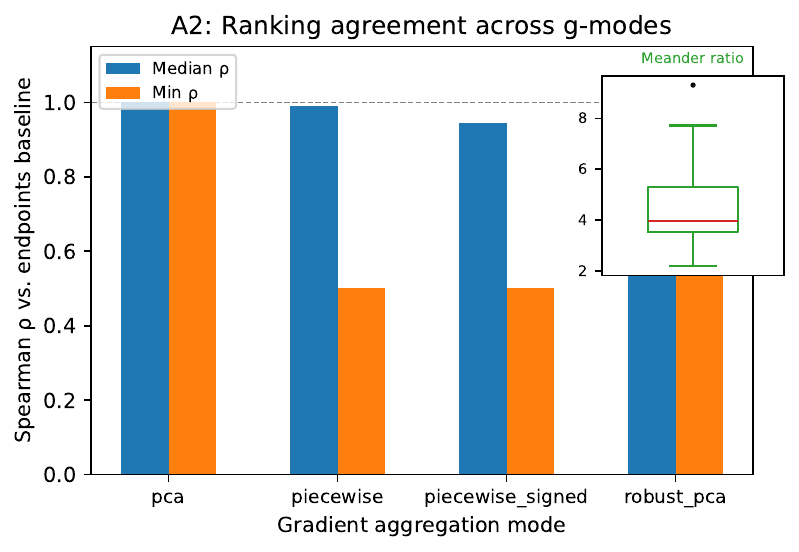}
  \caption{Direction-estimation robustness and meander diagnostics (A2).}
  \label{fig:a2_gmode_meander}
\end{figure}

\begin{table}[H]
\centering
\small
\begin{tabular}{lccc}
\toprule
Comparison & Median $\rho$ & Min $\rho$ & Notes \\
\midrule
PCA vs endpoints & 1.00 & 1.00 & 24/24 identical \\
Robust PCA vs endpoints & 1.00 & 1.00 & 24/24 identical \\
Piecewise vs endpoints & $\approx$1.00 & 0.50 & short-trace edge case \\
Piecewise-signed vs endpoints & $\approx$1.00 & 0.50 & same edge case \\
\bottomrule
\end{tabular}
\caption{Ranking invariance across $g$ estimators on the 24-example slice (model: R1-Distill-Qwen-1.5B).}
\label{tab:a2_gmode_meander}
\end{table}

\clearpage
\subsection{DRTC concentration across models}
\label{app:sparsity}

\paragraph{Concentration metrics.}
Across models, DRTC influence is \emph{moderately concentrated} rather than diffuse: using per-example normalized
$|\mathrm{DRTC}|$ mass shares (computed within each example and summarized as the median across 24 examples),
baseline (C0) Gini ranges from $0.50$--$0.58$, with the top $5\%$ of chunks carrying roughly $0.23$--$0.28$ of total
$|\mathrm{DRTC}|$ mass.
These are not extreme-sparsity values (e.g., Gini $\gg 0.8$), but they are consistent across architectures and sufficient to reject
``everything matters equally.''
(C8 matches C0 exactly by construction and is therefore omitted from the tables below.)

\paragraph{Architecture note.}
LFM2.5 occasionally exhibits higher top-1 mass outliers than the transformer backbones
(Table~\ref{tab:topk_mass}).
A plausible explanation is architectural inductive bias:
LFM2.5 combines convolutional components with grouped-query attention,
which may alter how influence concentrates across local versus long-range context
\citep{liquidai2025lfm2,wu2019payattentionlightweightdynamic,ainslie2023gqatraininggeneralizedmultiquery}.
We leave targeted tests of this effect (e.g., locality vs.\ long-range dependence of top-attributed chunks) to future work.

\begin{table}[h]
  \caption{Per-example concentration of $|\mathrm{DRTC}(i)|$ summarized by median Gini coefficient (C0)}
  \label{tab:gini_all}
  \centering
  \small
  \setlength{\tabcolsep}{6pt}
  \renewcommand{\arraystretch}{1.1}
  \begin{tabular}{lc}
    \toprule
    Model & Gini (C0) [95\% CI] \\
    \midrule
    LFM2.5-1.2B            & 0.499 [0.479, 0.557] \\
    Ministral-3-3B         & 0.574 [0.519, 0.625] \\
    Phi-4-Mini             & 0.539 [0.491, 0.549] \\
    R1-Distill-Qwen-1.5B   & 0.584 [0.556, 0.670] \\
    \bottomrule
  \end{tabular}

  \vspace{0.5ex}
  \parbox{0.95\linewidth}{\footnotesize
  Median across 24 examples with 95\% bootstrap confidence intervals (2000 resamples).
  Concentration metrics are computed on per-example normalized $|\mathrm{DRTC}|$ shares.
  }
\end{table}

\begin{table}[h]
  \caption{Top-$k$ mass fractions for per-example normalized $|\mathrm{DRTC}|$}
  \label{tab:topk_mass}
  \centering
  \small
  \setlength{\tabcolsep}{6pt}
  \renewcommand{\arraystretch}{1.1}
  \begin{tabular}{lccc}
    \toprule
    Model & Top-1 mass [95\% CI] & Top-5\% mass [95\% CI] & Mean $|\mathrm{DRTC}|$ \\
    \midrule
    LFM2.5-1.2B            & 0.076 [0.066, 0.092] & 0.229 [0.217, 0.258] & 0.492 \\
    Ministral-3-3B         & 0.086 [0.076, 0.097] & 0.262 [0.244, 0.286] & 0.455 \\
    Phi-4-Mini             & 0.071 [0.062, 0.084] & 0.241 [0.217, 0.254] & 0.493 \\
    R1-Distill-Qwen-1.5B   & 0.096 [0.080, 0.118] & 0.284 [0.264, 0.336] & 0.494 \\
    \bottomrule
  \end{tabular}

  \vspace{0.5ex}
  \parbox{0.95\linewidth}{\footnotesize
  Within each example, compute $|\mathrm{DRTC}_i|/\sum_j |\mathrm{DRTC}_j|$ across chunks, then report the median across 24 examples with bootstrapped 95\% confidence intervals. Mean $|\mathrm{DRTC}|$ denotes the per-example mean absolute DRTC value, reported as the median across examples.
  }
\end{table}

\begin{figure*}[h]
  \centering
  \begin{subfigure}[t]{0.95\textwidth}
    \centering
    \includegraphics[width=\linewidth]{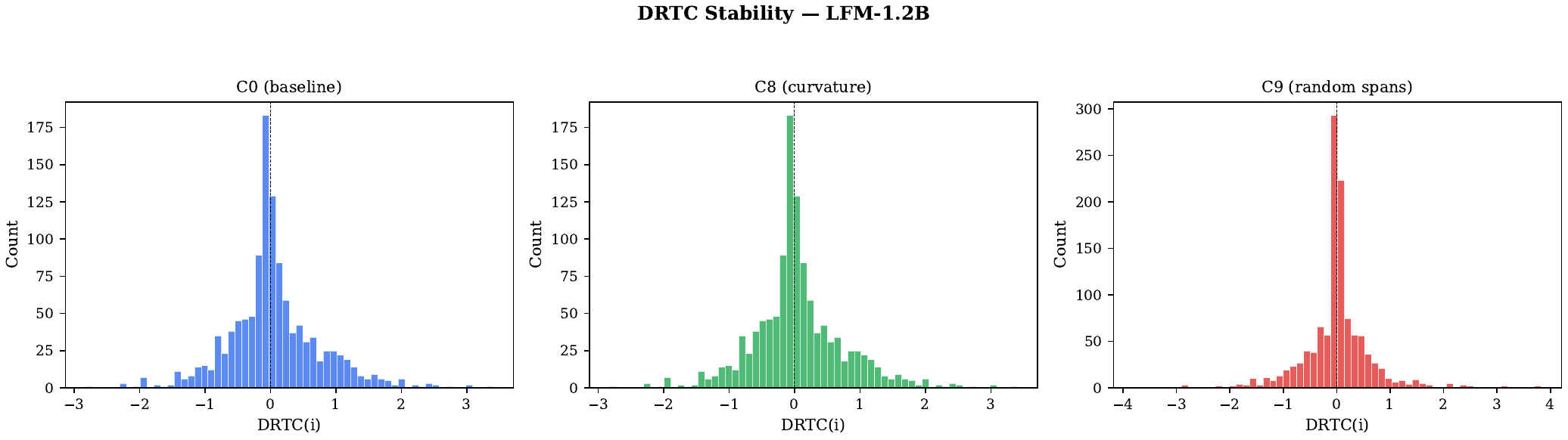}
    \caption{\scriptsize LFM2.5-1.2B: per-chunk $|\mathrm{DRTC}|$ histogram}
  \end{subfigure}

  \vspace{6pt}

  \begin{subfigure}[t]{0.95\textwidth}
    \centering
    \includegraphics[width=\linewidth]{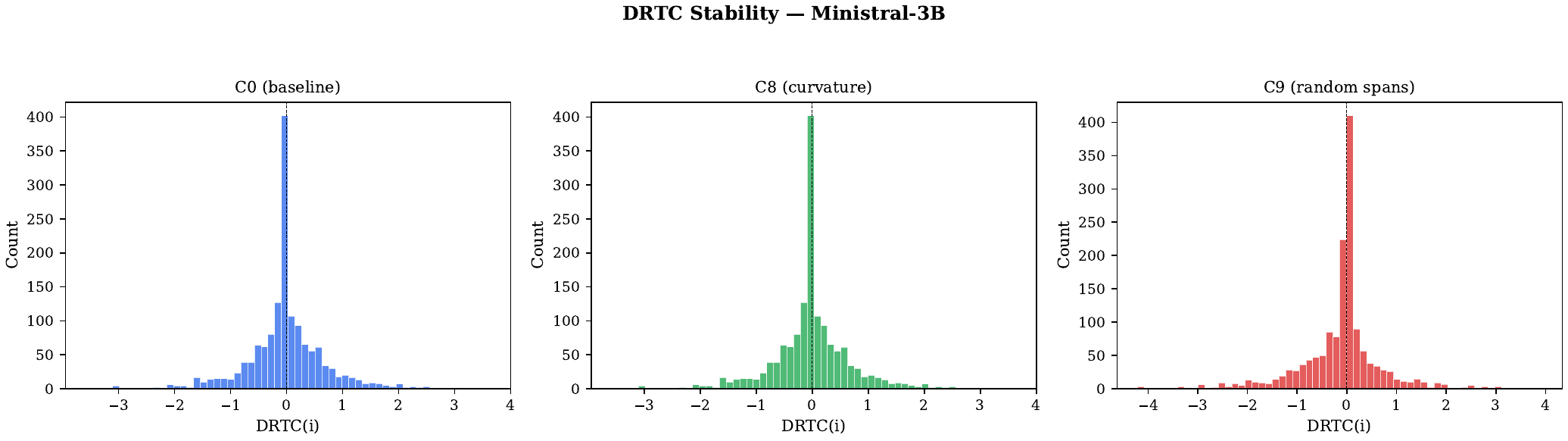}
    \caption{\scriptsize Ministral-3-3B: per-chunk $|\mathrm{DRTC}|$ histogram}
  \end{subfigure}

  \vspace{6pt}

  \begin{subfigure}[t]{0.95\textwidth}
    \centering
    \includegraphics[width=\linewidth]{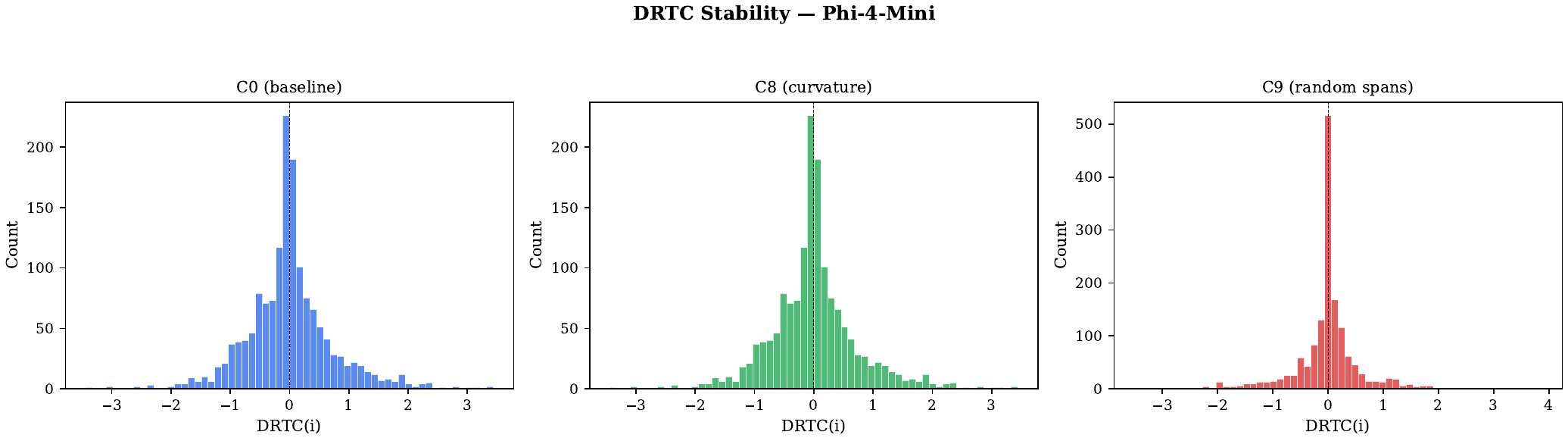}
    \caption{\scriptsize Phi-4-Mini: per-chunk $|\mathrm{DRTC}|$ histogram}
  \end{subfigure}

  \vspace{6pt}

  \begin{subfigure}[t]{0.95\textwidth}
    \centering
    \includegraphics[width=\linewidth]{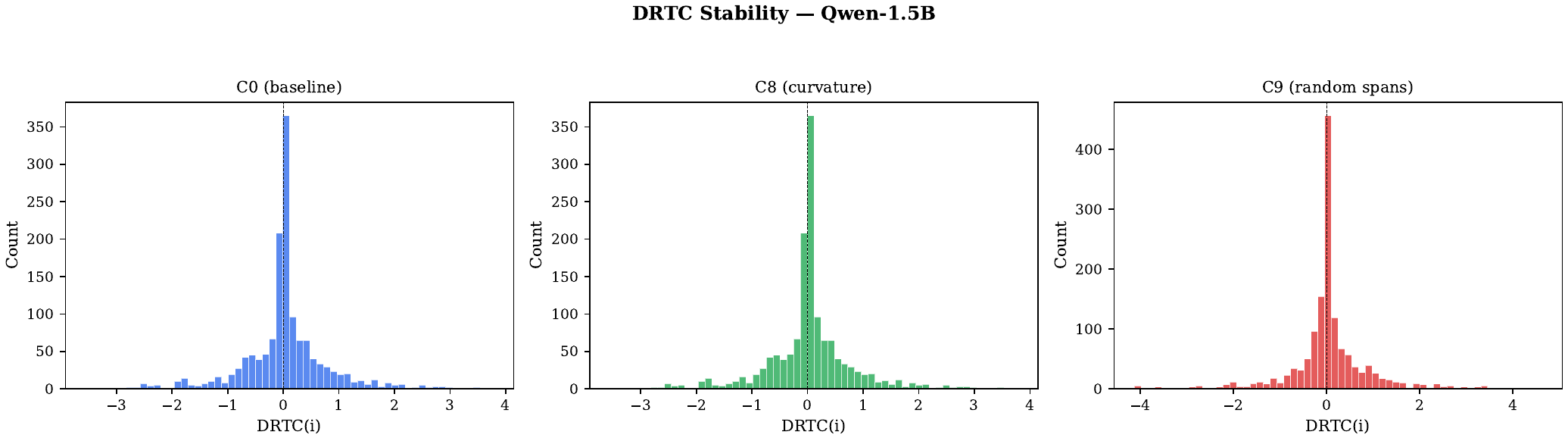}
    \caption{\scriptsize R1-Distill-Qwen-1.5B: per-chunk $|\mathrm{DRTC}|$ histogram}
  \end{subfigure}

  \vspace{6pt}
\end{figure*}

\begin{figure*}[t]
  \centering

  \begin{subfigure}[t]{0.48\textwidth}
    \centering
    \includegraphics[width=\linewidth]{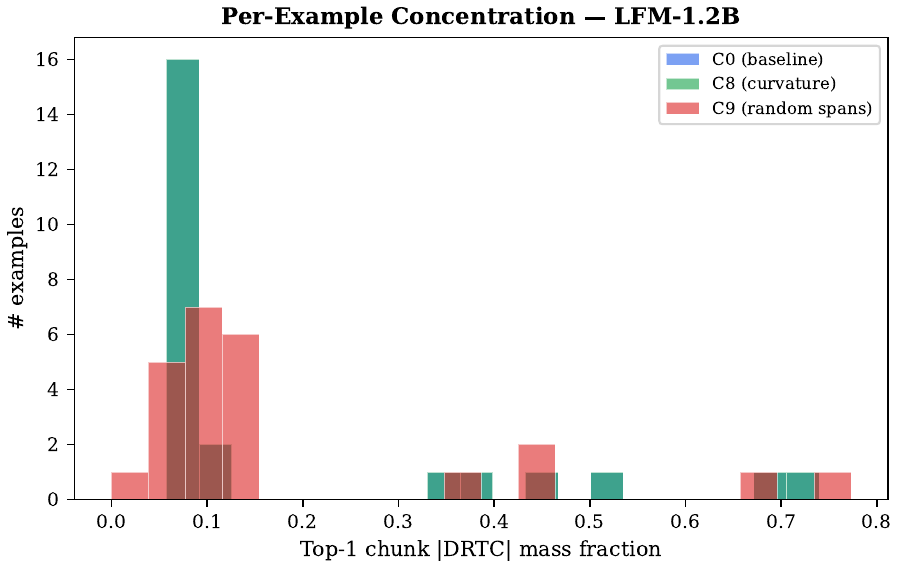}
    \caption{\scriptsize LFM2.5-1.2B}
  \end{subfigure}
  \hfill
  \begin{subfigure}[t]{0.48\textwidth}
    \centering
    \includegraphics[width=\linewidth]{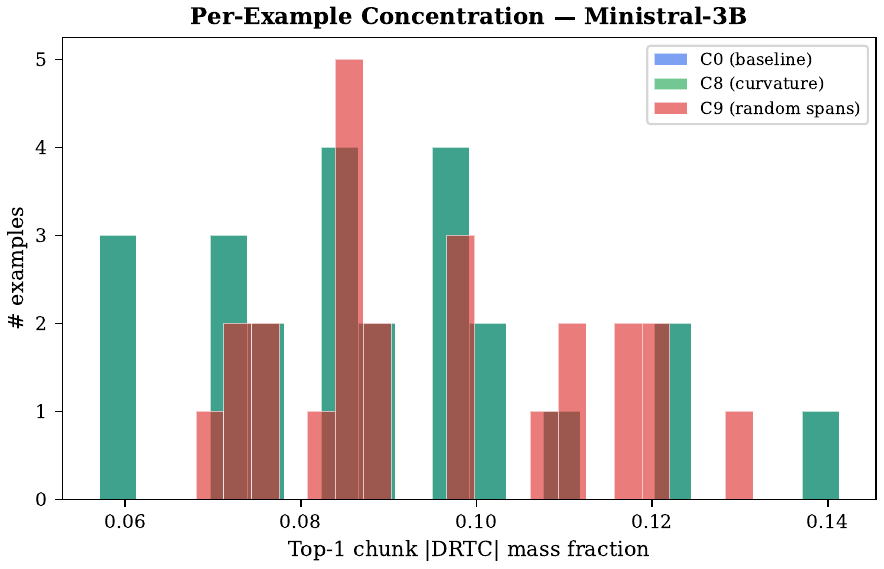}
    \caption{\scriptsize Ministral-3-3B}
  \end{subfigure}

  \vspace{6pt}

  \begin{subfigure}[t]{0.48\textwidth}
    \centering
    \includegraphics[width=\linewidth]{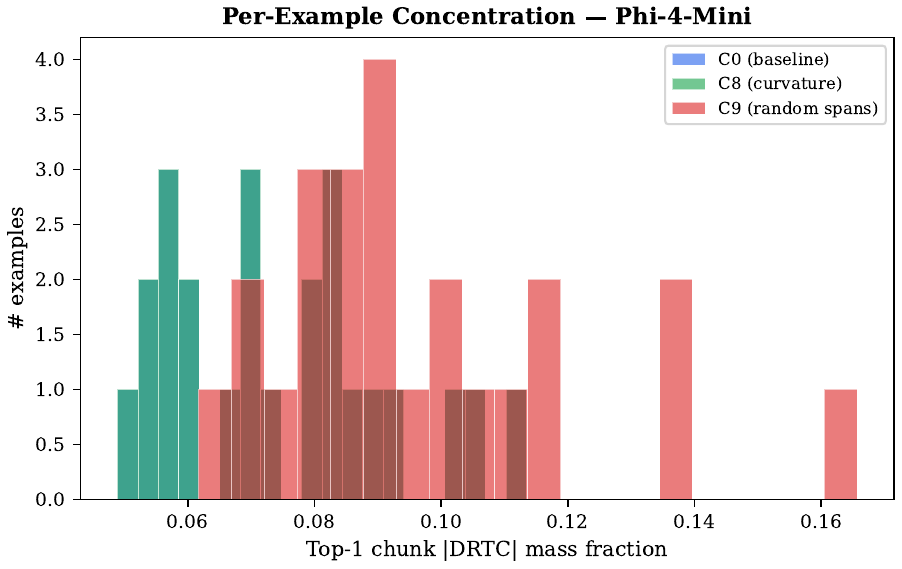}
    \caption{\scriptsize Phi-4-Mini}
  \end{subfigure}
  \hfill
  \begin{subfigure}[t]{0.48\textwidth}
    \centering
    \includegraphics[width=\linewidth]{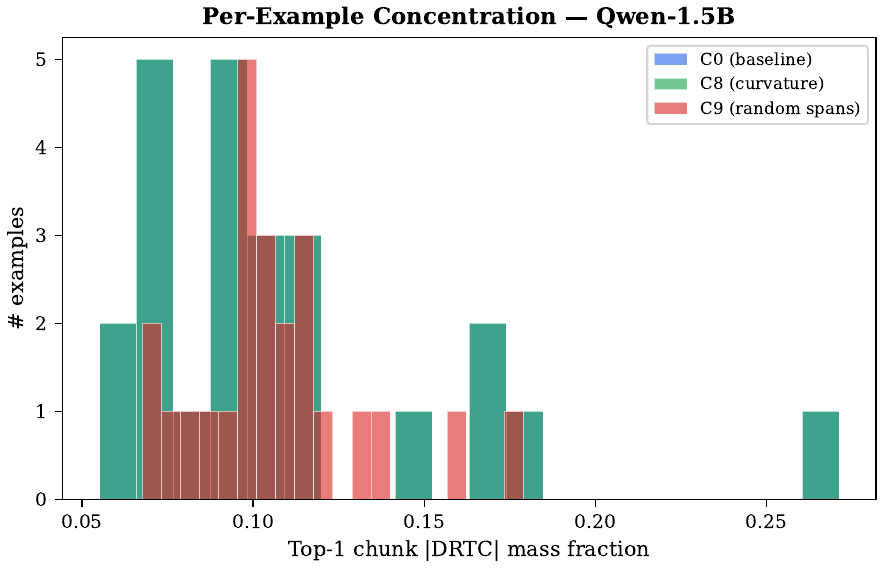}
    \caption{\scriptsize R1-Distill-Qwen-1.5B}
  \end{subfigure}

  \caption{
  Per-example DRTC attribution concentration across models.
  Each panel shows the fraction of total per-example $|\mathrm{DRTC}|$ mass captured by the top-ranked chunk(s) within an example.
  Across architectures, a small number of chunks accounts for a disproportionate share of directional influence,
  consistent with moderate but robust sparsity (see Tabs.~\ref{tab:gini_all}--\ref{tab:topk_mass}).
  }
  \label{fig:drtc_concentration_all}
\end{figure*}

\clearpage
\subsection{Random-span falsification: magnitude primary, structure secondary}
\label{app:falsification}

\paragraph{Primary metric (magnitude).}
Pivot-associated spans (C8) produce stronger interventions than matched random spans (C9) across all four models.
We quantify this using the per-example mean absolute pivot-local directional effect,
\[
\mathbb{E}_{k}\bigl[|\delta_{k,i}|\bigr],
\]
aggregated as the median across examples.
We treat magnitude as the primary falsification signal because it directly measures effect strength rather than rank preservation.

\paragraph{Random-span falsification control (C8 vs.\ C9).}
Both C8 and C9 use identical pivot discovery: we score token positions by uncertainty and distribution shift
(entropy, top-2 margin, and local Jensen--Shannon divergence) and select the top-$K$ pivots under spacing constraints.
The configurations differ only in \emph{which span of earlier tokens} is treated as the candidate cause whose
information flow is blocked at a pivot.

\textbf{Pivot spans (C8).}
For each pivot, we define an associated span window $[s_k, e_k)$ in the realized rollout (excluding prompt tokens),
placed by the pivot/span selection procedure. Receiver-side masking then blocks attention from the pivot position
$\tau_k$ to tokens in this pivot-associated span.

\textbf{Random spans (C9).}
After normal pivot discovery, we replace each pivot-associated span with a uniformly random span of identical length.
Concretely, for a pivot span of length $L_k = e_k - s_k$, we sample a start index $\tilde{s}_k$ uniformly over rollout
tokens (excluding prompt tokens), set $\tilde{e}_k = \tilde{s}_k + L_k$, and reject samples that overlap protected
regions (including the original pivot spans) up to a fixed retry budget.
The causal intervention mechanism is otherwise unchanged: the model is probed by blocking attention from $\tau_k$
to tokens in the (now random) control span.

\textbf{What stays fixed vs.\ what changes.}
Pivot positions $\{\tau_k\}$, the masking operator, and span lengths $\{L_k\}$ are identical across C8 and C9;
only span placement differs (pivot-associated vs.\ uniform random).
Curvature logging, when enabled, remains strictly diagnostic in both conditions.

\textbf{Interpretation.}
If DRTC is capturing meaningful causal influence, then masking pivot-associated spans (C8) should induce larger
pivot-local trajectory effects than masking matched-length random spans (C9).
We observe this consistently across models in the magnitude results reported in this section (see Fig.~\ref{fig:c8_vs_c9_magnitude} in the main paper for the cross-model summary).

\begin{table}[h]
  \caption{Primary falsification metric: per-example mean absolute pivot-local intervention magnitude ($\mathrm{std}_d$)}
  \label{tab:falsification_magnitude}
  \centering
  \small
  \setlength{\tabcolsep}{6pt}
  \renewcommand{\arraystretch}{1.1}
  \begin{tabular}{lccc}
    \toprule
    Model & C8 median $\mathrm{std}_d$ [95\% CI] & C9 median $\mathrm{std}_d$ [95\% CI] & Gap \\
    \midrule
    LFM2.5-1.2B            & 1.086 [1.007, 1.290] & 0.907 [0.852, 0.976] & +0.178 \\
    Ministral-3-3B         & 1.065 [0.982, 1.119] & 1.026 [0.957, 1.179] & +0.039 \\
    Phi-4-Mini             & 1.089 [0.971, 1.162] & 0.926 [0.862, 1.012] & +0.163 \\
    R1-Distill-Qwen-1.5B   & 1.290 [1.193, 1.367] & 1.208 [1.131, 1.254] & +0.081 \\
    \bottomrule
  \end{tabular}

  \vspace{0.5ex}
  \parbox{0.95\linewidth}{\footnotesize
  Median across 24 examples. Gap denotes C8 minus C9.
  }
\end{table}

\begin{figure*}[h]
  \centering
  \includegraphics[width=0.75\textwidth]{_posthoc_analysis/analysis_figures/cross_model/c8_vs_c9_magnitude.pdf}

  \vspace{6pt}

  \includegraphics[width=0.75\textwidth]{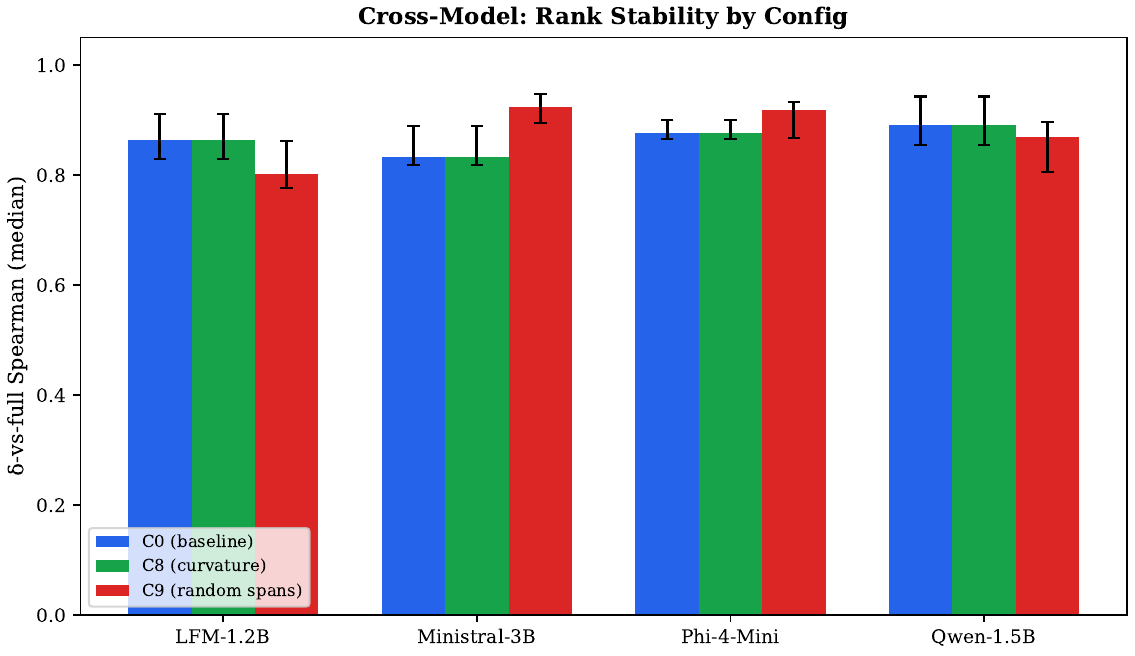}

  \caption{
  Cross-model falsification summary.
  \emph{Top}: primary falsification (median per-example intervention magnitude; std\_d), showing C8 $>$ C9 for all four models.
  \emph{Bottom}: $\delta$-vs-full Spearman rank-stability diagnostic; two models show C9 $>$ C8, which we interpret as rank-preservation effects rather than stronger causal interventions.
  Error bars denote 95\% bootstrap confidence intervals.
  }
  \label{fig:falsification_cross_model}
\end{figure*}

\paragraph{Metric clarification.}
Table~\ref{tab:falsification_magnitude} reports $\mathrm{std}_d$, our primary magnitude diagnostic
(per-example mean absolute pivot-local intervention magnitude, summarized by the median across examples).
In contrast, Figure~\ref{fig:falsification_cross_model} (top panel) reports the median per-example
$\mathrm{mean}\,|\delta_{k,i}|$, where $\delta_{k,i}=\langle e_{k,i}, g\rangle$ is the signed directional
projection used in DRTC aggregation.
These quantities are related but not identical: $\mathrm{std}_d$ summarizes overall intervention magnitude
at pivots, while $\mathrm{mean}\,|\delta_{k,i}|$ measures magnitude specifically along the realized
rollout direction.
Accordingly, absolute gap values differ across the table and figure, though both consistently show
stronger effects for pivot-associated spans (C8) than matched random spans (C9).

\paragraph{Secondary metric (rank stability).}
We also report $\delta$-vs-full Spearman as a rank-stability diagnostic.
In Ministral-3-3B and Phi-4-Mini, C9 exceeds C8 on Spearman; we report this transparently and interpret it as a property of correlation-based rank preservation rather than stronger intervention magnitude.

\clearpage
\begin{table}[h]
  \caption{Secondary falsification diagnostic: $\delta$-vs-full Spearman rank agreement}
  \label{tab:falsification_spearman}
  \centering
  \small
  \setlength{\tabcolsep}{6pt}
  \renewcommand{\arraystretch}{1.1}
  \begin{tabular}{lccc}
    \toprule
    Model & C8 median $\rho$ [95\% CI] & C9 median $\rho$ [95\% CI] & $\Delta$ \\
    \midrule
    LFM2.5-1.2B            & 0.863 [0.829, 0.910] & 0.802 [0.777, 0.842] & +0.061 \\
    Ministral-3-3B         & 0.832 [0.818, 0.889] & 0.924 [0.896, 0.949] & -0.092 \\
    Phi-4-Mini             & 0.877 [0.866, 0.900] & 0.918 [0.867, 0.933] & -0.041 \\
    R1-Distill-Qwen-1.5B   & 0.892 [0.854, 0.943] & 0.869 [0.805, 0.896] & +0.023 \\
    \bottomrule
  \end{tabular}

  \vspace{0.5ex}
  \parbox{0.95\linewidth}{\footnotesize
  Spearman rank agreement between pivot-local directional contributions ($\delta$) and full DRTC scores.
  }
\end{table}

\subsection{Computational cost}
\label{app:cost}

We report end-to-end computational cost for the DRTC pipeline on the 24-example slice (model: \texttt{R1-Distill-Qwen-1.5B}, $K=8$, greedy decoding, fixed seeds). Costs include pivot discovery, receiver-side interventions, and scoring.

\begin{table}[H]
\centering
\small
\begin{tabular}{lcccc}
\toprule
Metric & Mean & Median & Min & Max \\
\midrule
Wall time per example (s) & 11.8 & 12.28 & 3.72 & 17.03 \\
Peak GPU memory (MB) & 4835 & 4864 & 4164 & 5333 \\
\bottomrule
\end{tabular}
\caption{Compute cost summary from \texttt{main\_text\_table\_cost.csv}.}
\label{tab:cost}
\end{table}

\subsection{Masking ablation and implementation details}
\label{app:masking_ablation}

We ablate receiver-side attention-edge masking by masking subsets of heads (and optionally subsets of layers),
and compare resulting rankings against the default all-head, all-layer intervention.

\begin{table}[H]
\centering
\small
\begin{tabular}{lcc}
\toprule
Masking variant & $\rho$ vs all-heads (range) & Notes \\
\midrule
Head subsets (fixed layers) & 0.22 -- 1.00 & agreement can degrade sharply \\
\bottomrule
\end{tabular}
\caption{Head-subset masking can substantially alter rankings, motivating the default all-head intervention.}
\label{tab:b2_masking_heads}
\end{table}

\begin{figure}[H]
  \centering
  \includegraphics[width=0.75\linewidth]{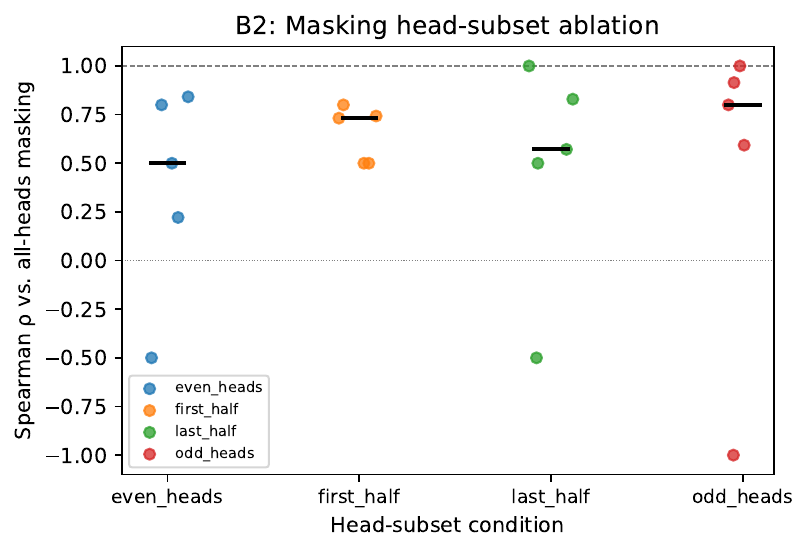}
  \caption{Representative ranking agreement for head-subset masking variants.}
  \label{fig:b2_masking_heads}
\end{figure}

\subsection{Hyperparameter sensitivity analysis}
\label{app:sensitivity}

We evaluate sensitivity of chunk rankings to key heuristic knobs (window size, pivot spacing, gating calibration,
and scaling/clipping parameters) by sweeping each knob over 2--3 values and measuring Spearman agreement
with the default setting.

\begin{table}[H]
\centering
\small
\begin{tabular}{lcc}
\toprule
Knob & Values & Median $\rho$ range \\
\midrule
window\_size & 3, 5, 8 & 0.93 -- 1.00 \\
min\_pivot\_spacing & 2, 4, 8 & 0.90 -- 1.00 \\
\textit{(others)} & \textit{(see CSV)} & $\ge 0.96$ typical \\
\bottomrule
\end{tabular}
\caption{Sensitivity grid summary (full table in the exported CSV).}
\label{tab:c3_sensitivity_summary}
\end{table}

\begin{figure}[H]
  \centering
  \includegraphics[width=0.75\linewidth]{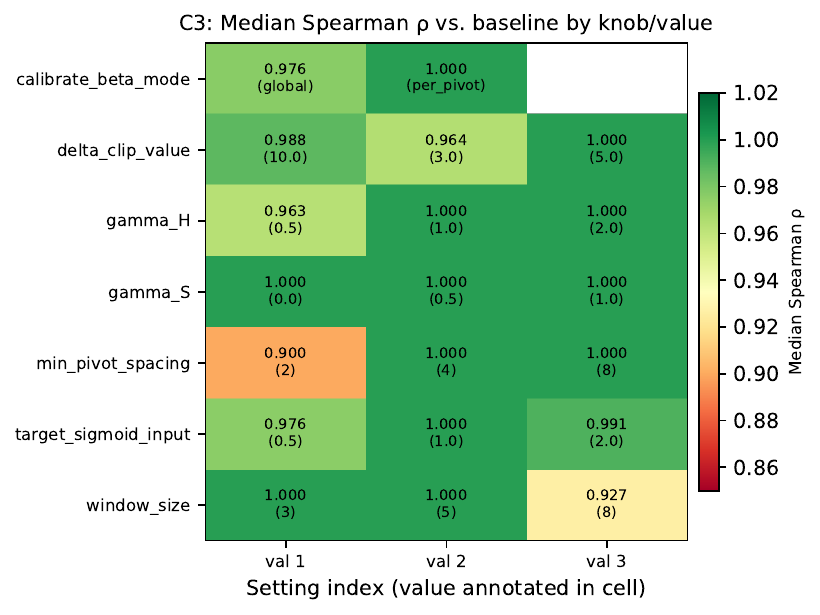}
  \caption{Sensitivity heatmap over knob sweeps.}
  \label{fig:c3_sensitivity_heatmap}
\end{figure}

\subsection{Baseline attribution comparisons}
\label{app:baselines}

We compare DRTC rankings to chunk-level baseline attributions computed on the same chunks and the same
directional scalar where applicable (occlusion, grad$\times$input surrogate, smooth masking, activation patching),
and report agreement and top-$k$ overlap.

\begin{table}[H]
\centering
\small
\begin{tabular}{lccc}
\toprule
Method & Median $\rho$ vs DRTC & Top-5 overlap & Top-10 overlap \\
\midrule
Grad$\times$Input & 0.53 & 0.67 & 0.82 \\
Smooth masking & 0.45 & 0.43 & 0.82 \\
Chunk occlusion & 0.30 & 0.43 & 0.67 \\
Activation patching & 0.04 & 0.43 & 0.67 \\
\bottomrule
\end{tabular}
\caption{Baseline agreement summary (R1-Distill-Qwen-1.5B, 24-example slice).}
\label{tab:baseline_summary}
\end{table}

\begin{figure}[H]
  \centering
  \includegraphics[width=0.75\linewidth]{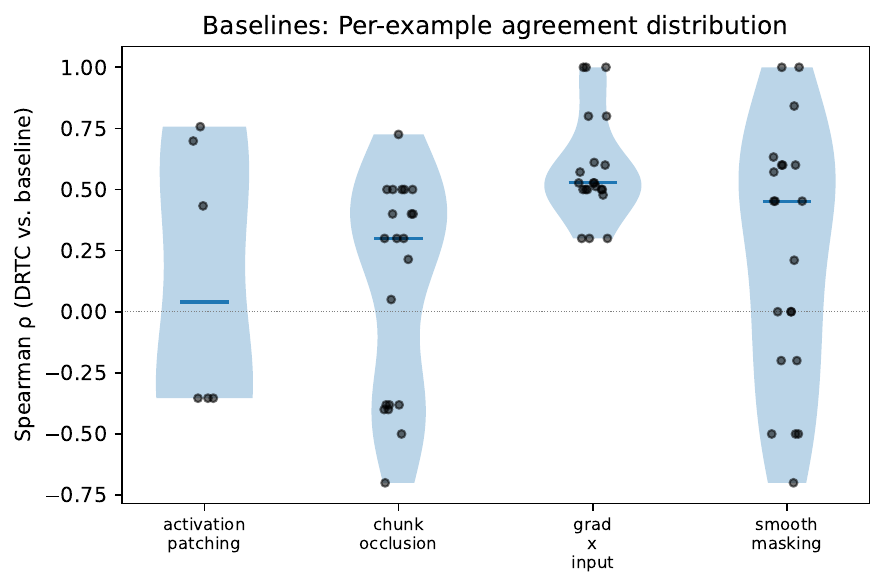}
  \caption{Distribution of per-example Spearman agreement between DRTC and baseline methods.}
  \label{fig:baselines_spearman_dist}
\end{figure}

\subsection{Curvature calibration and reconciliation}
\label{app:curvature_calibration}

We distinguish (i) within-trace co-localization between chunk-level curvature-impact and DRTC, from
(ii) cross-example correlations between per-example summary statistics. These measure different quantities
and need not agree.

\begin{table}[H]
\centering
\small
\begin{tabular}{lc}
\toprule
Statistic (within-trace, N=423) & Value \\
\midrule
Median $\rho(\mathrm{CurvImpact}, \mathrm{DRTC})$ & 0.77 \\
Bootstrap 95\% CI & [0.70, 0.80] \\
Fraction $p<0.05$ & 48\% \\
\bottomrule
\end{tabular}
\caption{Within-trace curvature-impact co-localization at scale (competition-math scaling run).}
\label{tab:e3_curvature_within_trace}
\end{table}

\begin{figure}[H]
  \centering
  \includegraphics[width=0.75\linewidth]{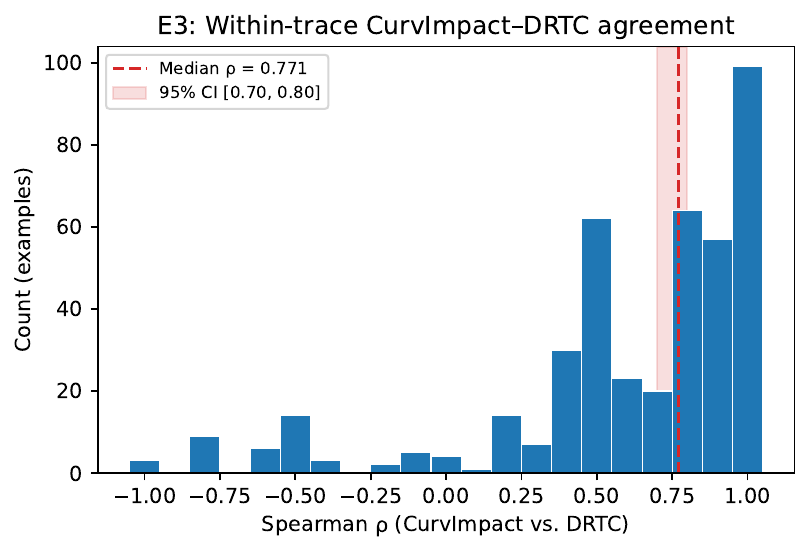}
  \caption{Distribution of within-trace Spearman correlations for curvature-impact vs DRTC.}
  \label{fig:e3_curvature_dist}
\end{figure}

\subsection{Curvature diagnostic: geometry stratifies intervention-response types}
\label{app:curvature_diag}

Curvature (turning-angle / CurvImpact) is a diagnostic layer: it describes the \emph{geometry} of intervention response rather than asserting causal influence on correctness.
Absolute CurvImpact magnitudes vary substantially by architecture, but relative structure is stable within each model.

\begin{figure*}[t]
  \centering

  \begin{subfigure}[t]{0.8\textwidth}
    \centering
    \includegraphics[width=\linewidth]{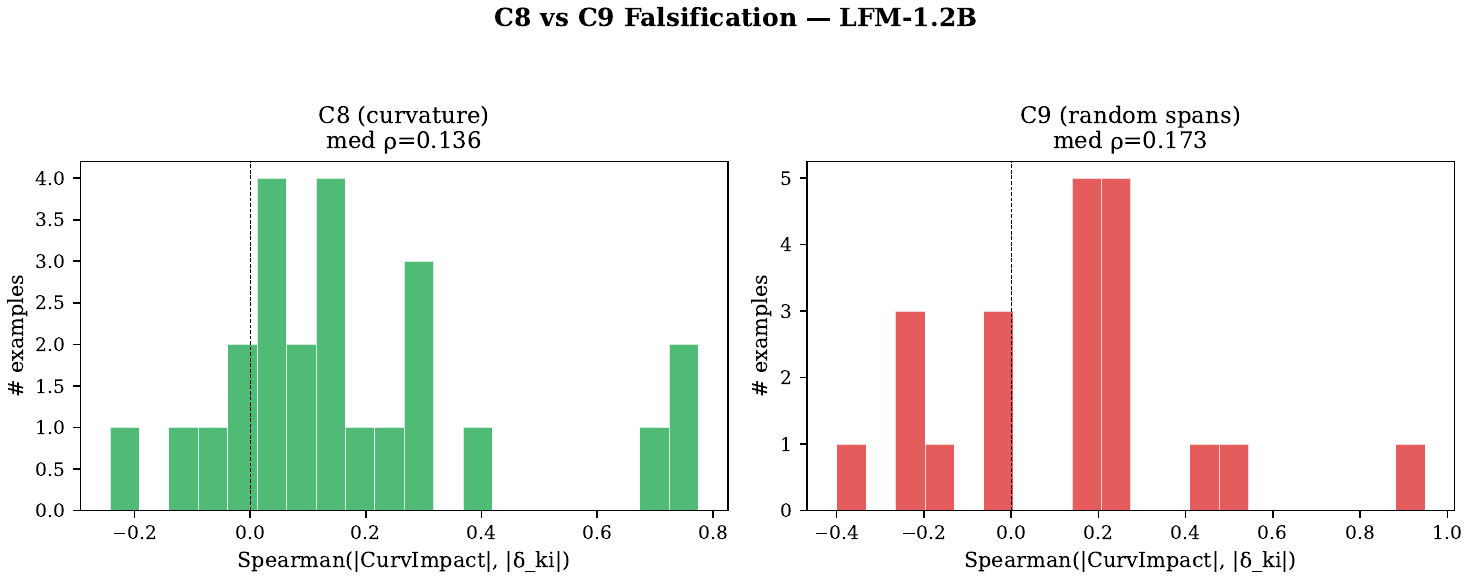}
    \caption{\scriptsize LFM2.5-1.2B}
  \end{subfigure}
  \hfill
  \begin{subfigure}[t]{0.8\textwidth}
    \centering
    \includegraphics[width=\linewidth]{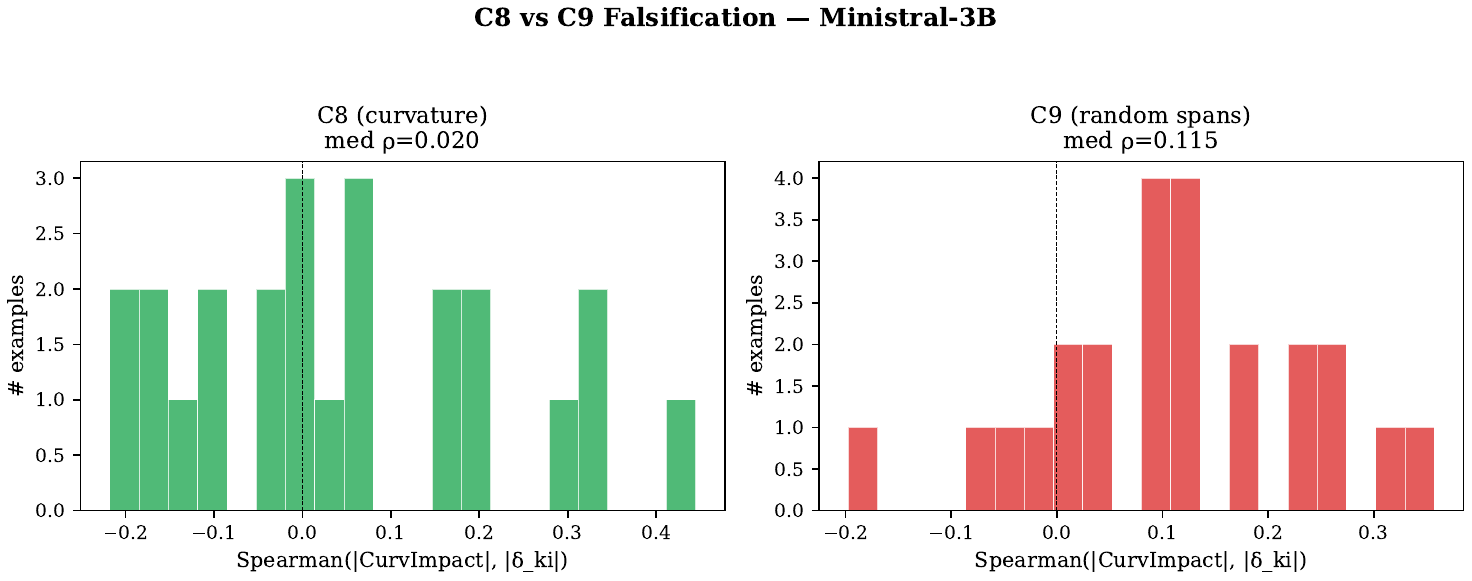}
    \caption{\scriptsize Ministral-3-3B}
  \end{subfigure}

  \vspace{6pt}

  \begin{subfigure}[t]{0.8\textwidth}
    \centering
    \includegraphics[width=\linewidth]{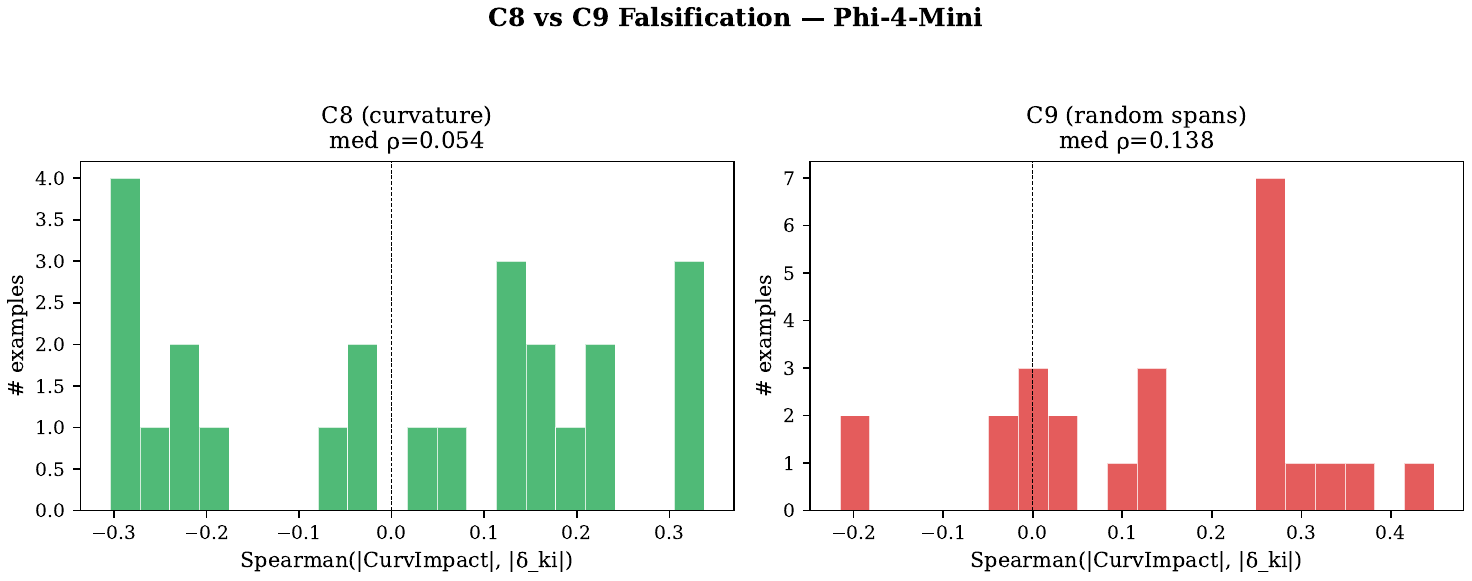}
    \caption{\scriptsize Phi-4-Mini}
  \end{subfigure}
  \hfill
  \begin{subfigure}[t]{0.8\textwidth}
    \centering
    \includegraphics[width=\linewidth]{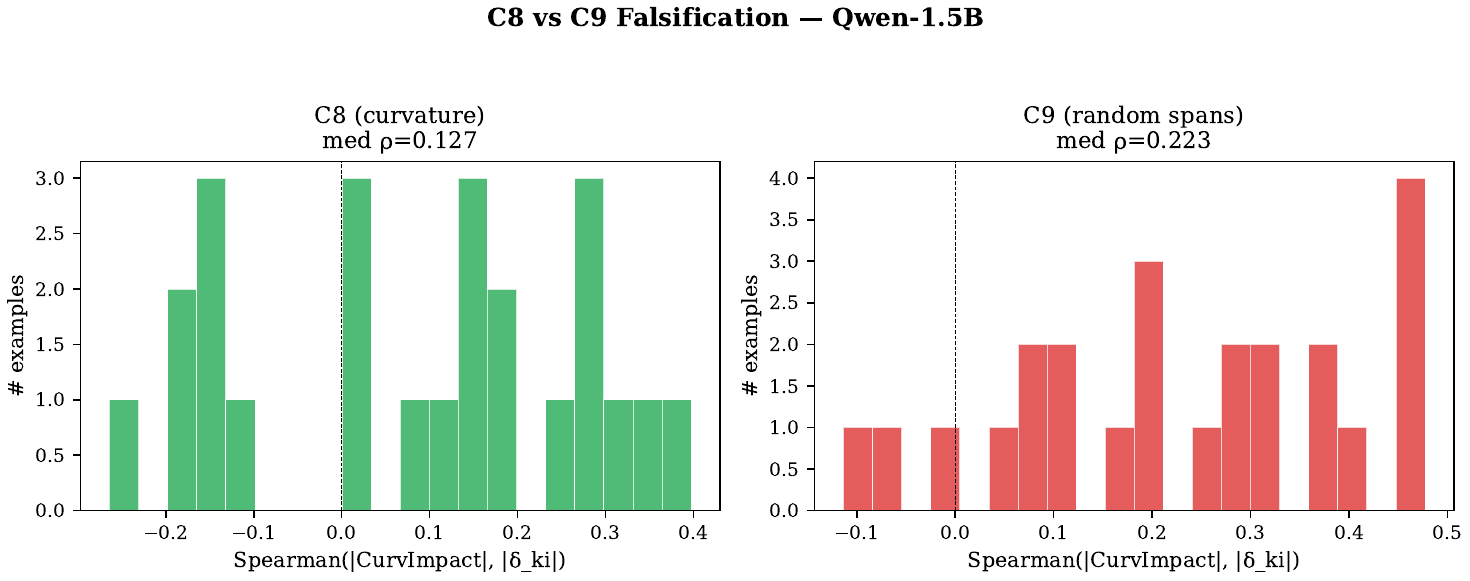}
    \caption{\scriptsize R1-Distill-Qwen-1.5B}
  \end{subfigure}

  \caption{
  Curvature diagnostic across models: relationship between curvature magnitude (CurvImpact) and attribution magnitude (DRTC), shown for both pivot probes (C8) and random-span control (C9).
  Curvature is used diagnostically to stratify how chunks bend the trajectory under intervention; it is not interpreted as outcome-causal evidence.
  }
  \label{fig:curvimpact_corr_all}
\end{figure*}

\paragraph{Representative curvature diagnostic visualization.}
\label{app:curvature_diag_vis}

\begin{figure}[t]
  \centering
  \includegraphics[width=\linewidth]{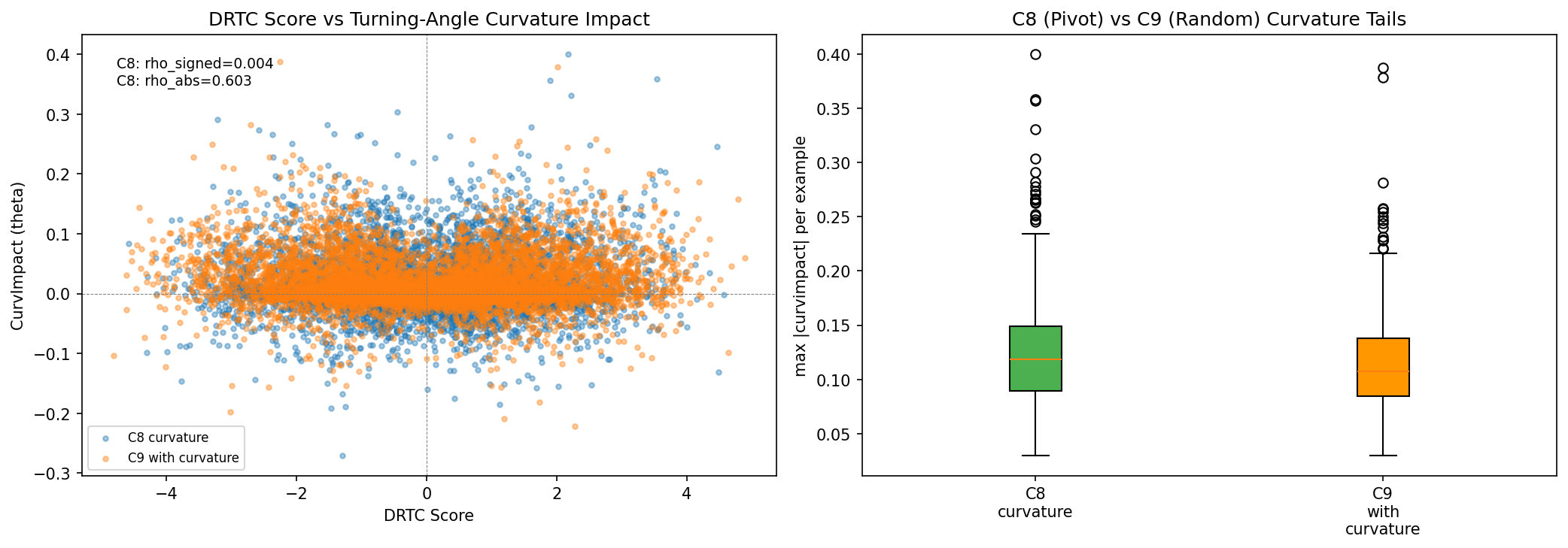}
  \caption{
  Representative curvature diagnostic visualization.
  \textbf{Left:} Scatter of $\mathrm{DRTC}(i)$ vs.\ $\mathrm{CurvImpact}(i)$ for a single example
  under \textsc{C8} (pivot spans) and \textsc{C9} (matched random spans).
  Signed association is near zero, while magnitude association can be positive,
  consistent with curvature tracking \emph{reorientation intensity} rather than
  directional steering.
  \textbf{Right:} Per-example maxima of $|\mathrm{CurvImpact}|$ under \textsc{C8}
  and \textsc{C9} (same masking logic, random-span replacement in \textsc{C9}),
  shown as an auxiliary diagnostic of geometric response magnitude.
  }
  \label{fig:curvature_diagnostics_example}
\end{figure}

Figure~\ref{fig:curvature_diagnostics_example} provides an intuition-building
curvature diagnostic visualization comparing the pivot condition (\textsc{C8})
to the matched random-span control (\textsc{C9}). The left panel plots per-chunk
$\mathrm{DRTC}(i)$ against turning-angle curvature impact
$\mathrm{CurvImpact}(i)$ (computed in raw logit space) for a single example.
Consistent with the aggregate results, signed $\mathrm{CurvImpact}$ exhibits
near-zero association with signed $\mathrm{DRTC}$, while larger reorientation
intensity (higher $|\mathrm{CurvImpact}|$) can coincide with larger directional
attribution magnitude $|\mathrm{DRTC}|$.
The right panel provides a per-example summary of maximum $|\mathrm{CurvImpact}|$
under \textsc{C8} and \textsc{C9} as an auxiliary geometric diagnostic.

\clearpage
\subsection{Adjacency effects and shuffle control}
\label{app:adjacency}

Adjacent chunks (near the pivot-containing chunk) have 2--3$\times$ higher mean $|\mathrm{DRTC}|$ contribution than distant chunks; this is expected because nearby text is more likely to be locally relevant.
However, distant chunks retain non-trivial mean influence (Table~\ref{tab:adjacency}), so DRTC is not fully explained by proximity.
Moreover, proximity alone would predict that distance-matched random spans behave similarly to learned pivots; empirically, learned pivots induce larger intervention magnitudes than matched random spans (C8 vs.\ C9; Appendix~\ref{app:falsification}).

\begin{table}[H]
  \caption{Adjacency analysis based on chunk-index distance}
  \label{tab:adjacency}
  \centering
  \small
  \setlength{\tabcolsep}{6pt}
  \renewcommand{\arraystretch}{1.1}
  \begin{tabular}{lcccc}
    \toprule
    Model & Adjacent mean $|\mathrm{DRTC}|$ & Distant mean $|\mathrm{DRTC}|$ & Ratio & p-value \\
    \midrule
    LFM2.5-1.2B            & 0.2718 & 0.0971 & 2.80$\times$ & $1.65\times10^{-69}$ \\
    Ministral-3-3B         & 0.2822 & 0.0917 & 3.08$\times$ & $1.13\times10^{-95}$ \\
    Phi-4-Mini             & 0.2234 & 0.1047 & 2.13$\times$ & $2.78\times10^{-40}$ \\
    R1-Distill-Qwen-1.5B   & 0.2749 & 0.0873 & 3.15$\times$ & $4.68\times10^{-92}$ \\
    \bottomrule
  \end{tabular}

  \vspace{0.5ex}
  \parbox{0.95\linewidth}{\footnotesize
  Adjacent: $|\Delta i|\le 1$ from pivot-containing chunk; distant: $|\Delta i|\ge 5$.
  }
\end{table}

\begin{figure*}[h]
  \centering

  \begin{subfigure}[t]{0.48\textwidth}
    \centering
    \includegraphics[width=\linewidth]{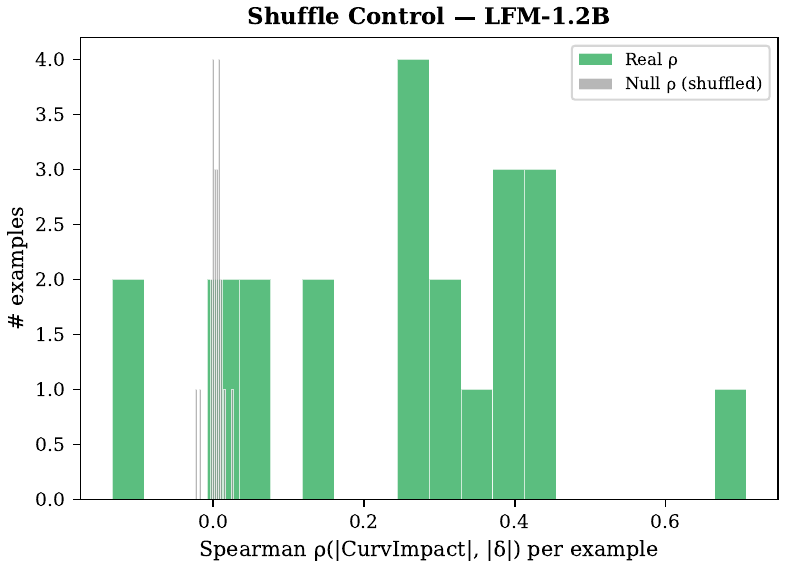}
    \caption{\scriptsize LFM2.5-1.2B}
  \end{subfigure}
  \hfill
  \begin{subfigure}[t]{0.48\textwidth}
    \centering
    \includegraphics[width=\linewidth]{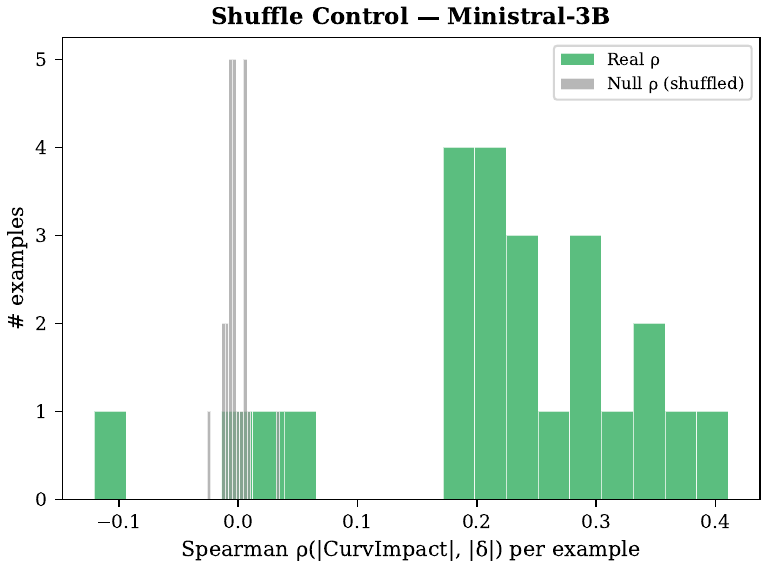}
    \caption{\scriptsize Ministral-3-3B}
  \end{subfigure}

  \vspace{6pt}

  \begin{subfigure}[t]{0.48\textwidth}
    \centering
    \includegraphics[width=\linewidth]{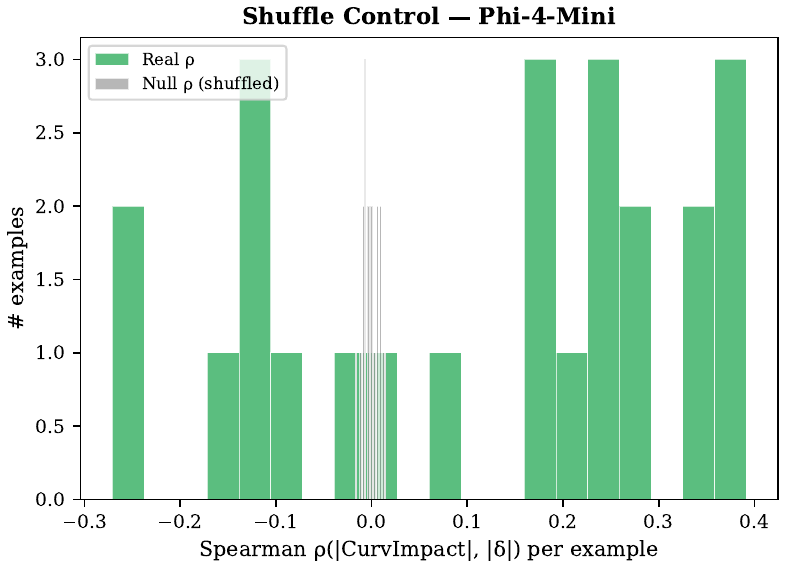}
    \caption{\scriptsize Phi-4-Mini}
  \end{subfigure}
  \hfill
  \begin{subfigure}[t]{0.48\textwidth}
    \centering
    \includegraphics[width=\linewidth]{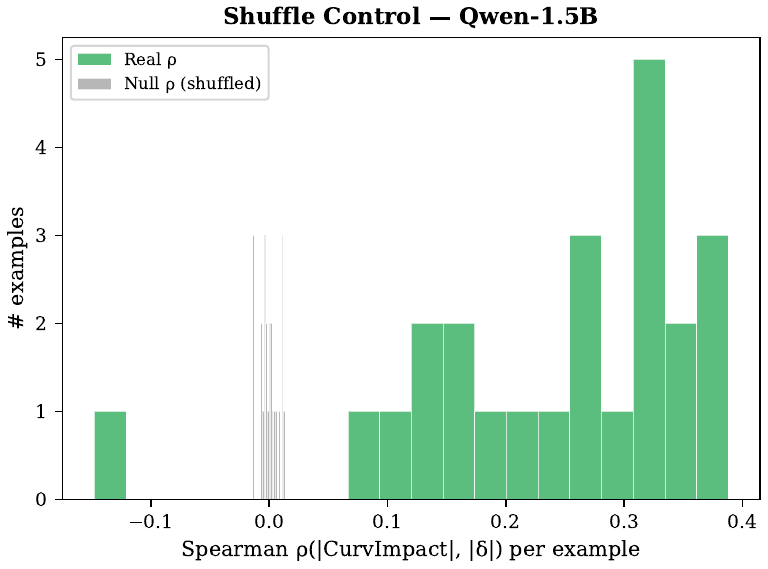}
    \caption{\scriptsize R1-Distill-Qwen-1.5B}
  \end{subfigure}

  \caption{
  Shuffle control across models. Randomly permuting curvature signatures destroys role structure, supporting the claim that curvature similarity is not a trivial artifact of the clustering mechanism.
  }
  \label{fig:shuffle_all}
\end{figure*}

\begin{table}[H]
  \caption{Shuffle control for curvature–DRTC correlation.
  Per-example Spearman correlation $\rho(\mathrm{CurvImpact}, |\delta|)$
  compared against a null obtained by randomly permuting curvature signatures.
  }
  \label{tab:shuffle_control}
  \centering
  \small
  \setlength{\tabcolsep}{6pt}
  \renewcommand{\arraystretch}{1.1}
  \begin{tabular}{lccc}
    \toprule
    Model & Real median $\rho$ & Null median $\rho$ & $p$-value \\
    \midrule
    LFM2.5-1.2B            & 0.243 & 0.003 & $1.4\times10^{-5}$ \\
    Ministral-3-3B         & 0.218 & -0.002 & $9.7\times10^{-9}$ \\
    Phi-4-Mini             & 0.112 & -0.001 & $1.4\times10^{-2}$ \\
    R1-Distill-Qwen-1.5B   & 0.243 & 0.000 & $1.5\times10^{-9}$ \\
    \bottomrule
  \end{tabular}

  \vspace{0.5ex}
  \parbox{0.95\linewidth}{\footnotesize
  Correlations computed per example and summarized by median across 24 examples.
  Null distribution obtained by permuting curvature signatures within each example.
  $p$-values from two-sided test comparing real vs.\ null distributions.
  }
\end{table}

\subsection{Pivot weighting ($u_k$) concentration}
\label{app:uk}

Pivot weighting is concentrated: the maximum pivot weight per example is typically $\approx 0.15$--$0.17$, with similar entropy across models.

\begin{table}[H]
  \caption{Pivot-weight distribution summary across models}
  \label{tab:uk_summary}
  \centering
  \small
  \setlength{\tabcolsep}{6pt}
  \renewcommand{\arraystretch}{1.1}
  \begin{tabular}{lcc}
    \toprule
    Model & Median max($u_k$) & Median entropy($u_k$) \\
    \midrule
    LFM2.5-1.2B            & 0.162 & 2.067 \\
    Ministral-3-3B         & 0.172 & 2.053 \\
    Phi-4-Mini             & 0.169 & 2.080 \\
    R1-Distill-Qwen-1.5B   & 0.148 & 2.072 \\
    \bottomrule
  \end{tabular}
\end{table}

\begin{table}[H]
    \caption{Pivot-weighting ablation: Spearman rank correlation between
    learned-weight DRTC and post-hoc uniform aggregation}
    \label{tab:uk_spearman}
    \centering
    \small
    \setlength{\tabcolsep}{6pt}
    \renewcommand{\arraystretch}{1.1}
    \begin{tabular}{lc}
      \toprule
      Model & Median Spearman $\rho$ [95\% CI] \\
      \midrule
      LFM2.5-1.2B            & 0.990 [0.966, 0.994] \\
      Ministral-3B            & 0.975 [0.962, 0.986] \\
      Phi-4-Mini              & 0.988 [0.981, 0.991] \\
      R1-Distill-Qwen-1.5B   & 0.988 [0.979, 0.995] \\
      \bottomrule
    \end{tabular}

    \vspace{0.5ex}
    \parbox{0.95\linewidth}{\footnotesize
    Per-example Spearman rank correlation between DRTC(i) computed with learned pivot weights $u_k$
    and post-hoc uniform aggregation ($u_k = 1/K$), summarized as the median across 24 examples with
    bootstrapped 95\% confidence intervals.
    }
  \end{table}

\begin{figure*}[t]
  \centering

  \begin{subfigure}[t]{0.99\textwidth}
    \centering
    \includegraphics[width=\linewidth]{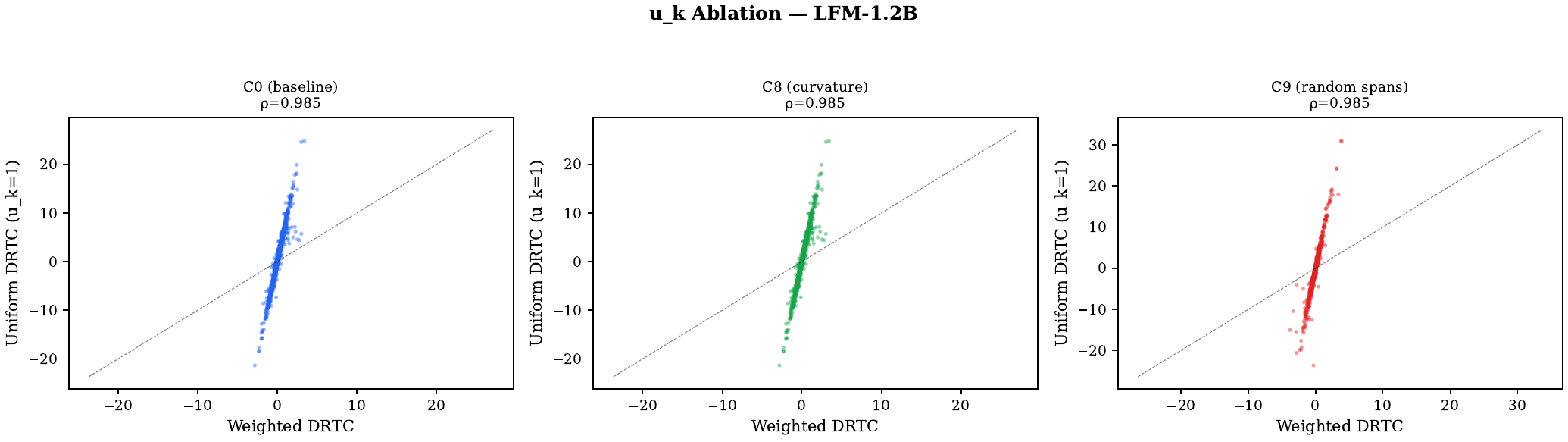}
    \caption{\scriptsize LFM2.5-1.2B}
  \end{subfigure}
  \hfill
  \begin{subfigure}[t]{0.99\textwidth}
    \centering
    \includegraphics[width=\linewidth]{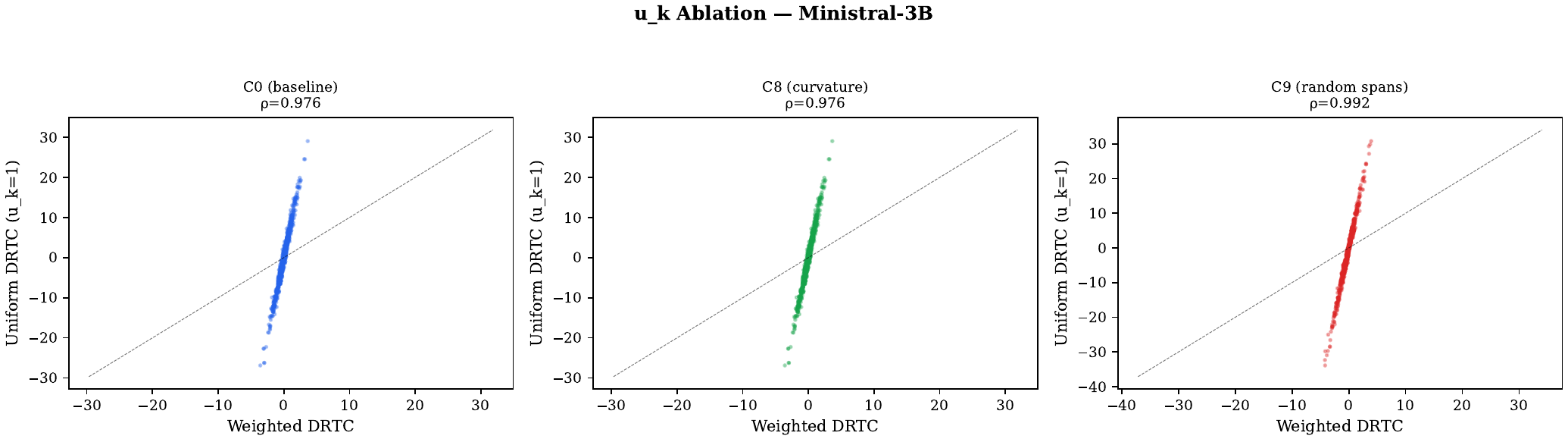}
    \caption{\scriptsize Ministral-3-3B}
  \end{subfigure}

  \vspace{6pt}

  \begin{subfigure}[t]{0.99\textwidth}
    \centering
    \includegraphics[width=\linewidth]{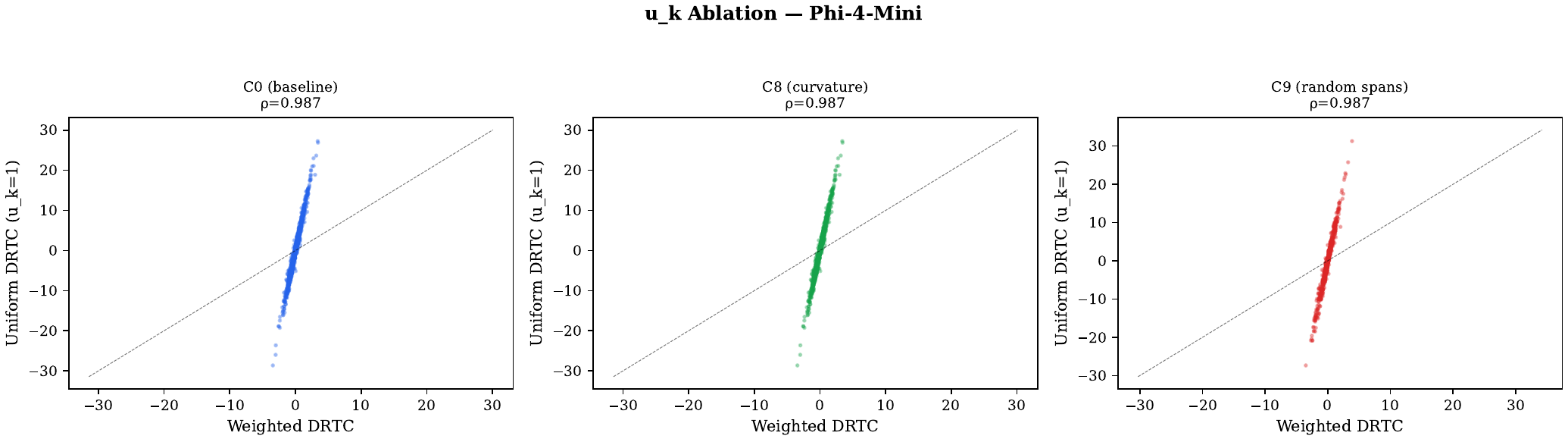}
    \caption{\scriptsize Phi-4-Mini}
  \end{subfigure}
  \hfill
  \begin{subfigure}[t]{0.99\textwidth}
    \centering
    \includegraphics[width=\linewidth]{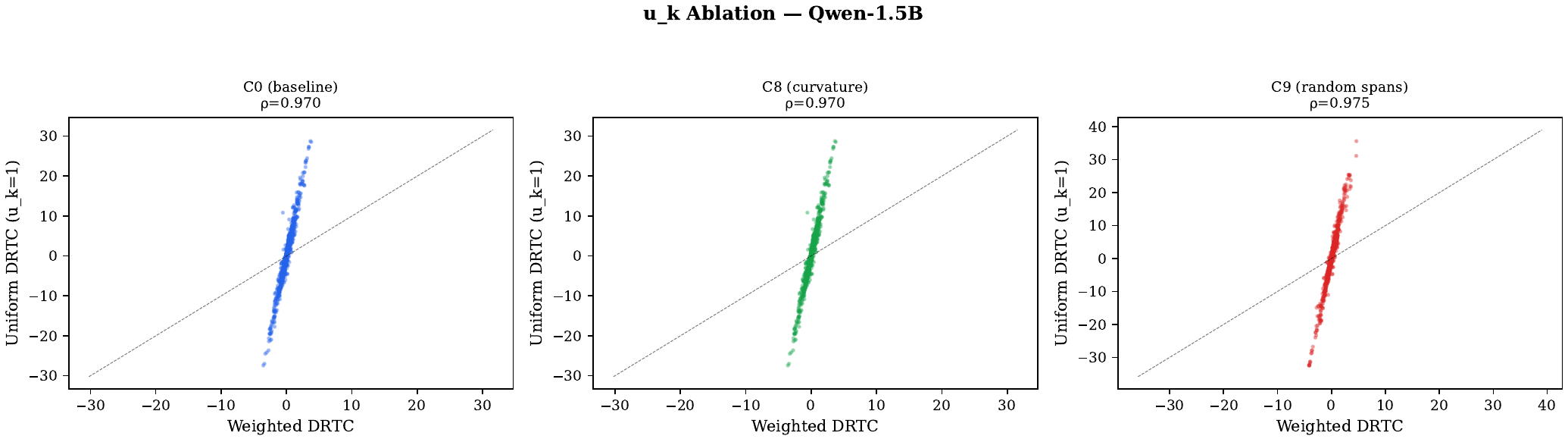}
    \caption{\scriptsize R1-Distill-Qwen-1.5B}
  \end{subfigure}

  \caption{
  Pivot-weighting ablation diagnostic across models.
  Points compare DRTC(i) computed with learned pivot weights $u_k$
  to post-hoc uniform aggregation ($u_k = 1/K$) using identical
  per-pivot effects $(\delta_{k,i}, w_{k,i})$.
  Because $u_k$ enters only at final aggregation and does not influence
  $\delta_{k,i}$ or $w_{k,i}$, this is mathematically equivalent to rerunning
  the method with uniform pivot weights.
  Spearman correlations are high but strictly less than 1, indicating that
  pivot importance modulates attribution rankings without collapsing them
  to a single dominant pivot.
  }
  \label{fig:uk_scatter_all}
\end{figure*}

\clearpage
\subsection{Outcome linkage via graded interventions}
\label{app:outcome_u1b}

We evaluate whether top-ranked DRTC chunks have stronger outcome-level effects than position-matched random
chunks using graded perturbations and an outcome-facing metric: teacher-forced $\Delta \log p(\text{gold})$.
Random controls are strictly quartile-matched to the top chunk’s position bin to remove positional confounds.

\begin{table}[H]
\centering
\small
\begin{tabular}{lccccc}
\toprule
Cohort & Regime $(cs,\alpha)$ & $N$ & Median paired diff & 95\% CI & Sign frac \\
\midrule
Stable subset ($|\mathrm{orig\_margin}|>1$) & (4, 0.2) & 19 & 0.34 & [0.03, 1.19] & 0.74 \\
All usable examples & (8, 0.7) & 22 & 0.44 & [0.04, 1.53] & 0.73 \\
\bottomrule
\end{tabular}
\caption{
U1b outcome linkage using teacher-forced $\Delta \log p(\text{gold})$.
Paired diff is defined as $\mathbb{E}[\Delta\log p(\text{gold})]_{\text{random}} - \Delta\log p(\text{gold})_{\text{top}}$,
so positive values indicate the top chunk is more damaging than matched random.
}
\label{tab:u1b_main}
\end{table}

\begin{figure}[H]
  \centering
  \includegraphics[width=\linewidth]{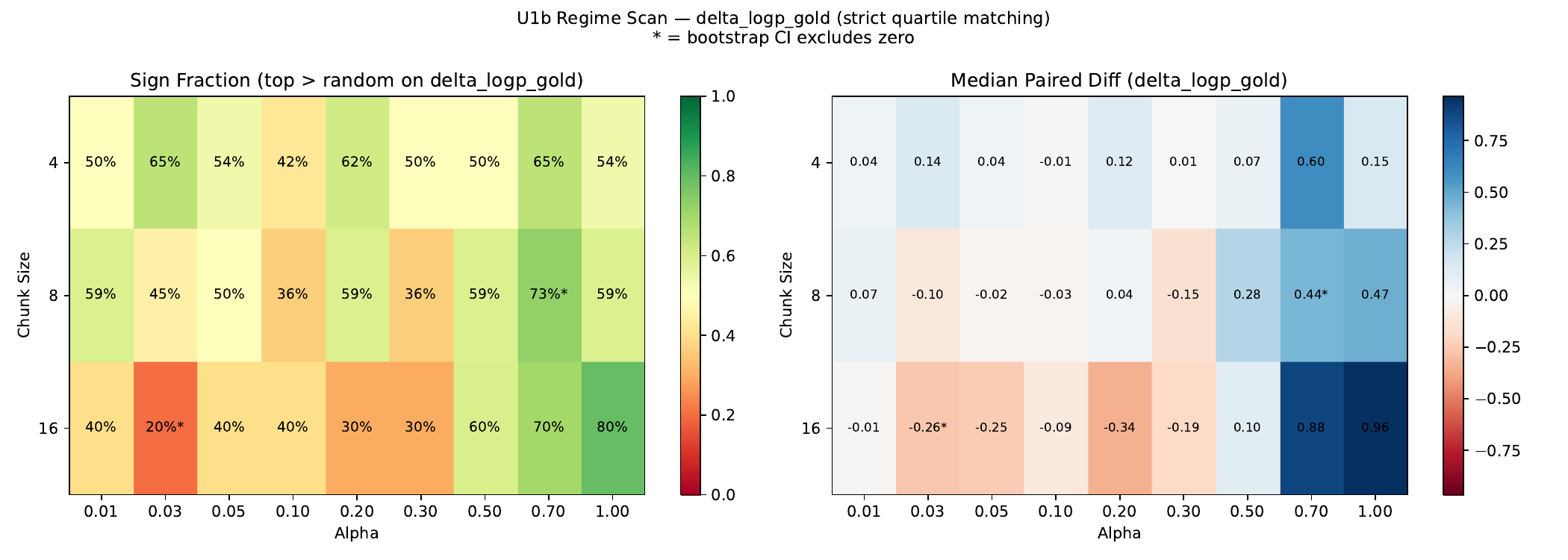}
  \caption{U1b regime scan over $(cs,\alpha)$ showing discrimination on $\Delta\log p(\text{gold})$.}
  \label{fig:u1b_heatmap}
\end{figure}

\begin{figure}[H]
  \centering
  \includegraphics[width=0.75\linewidth]{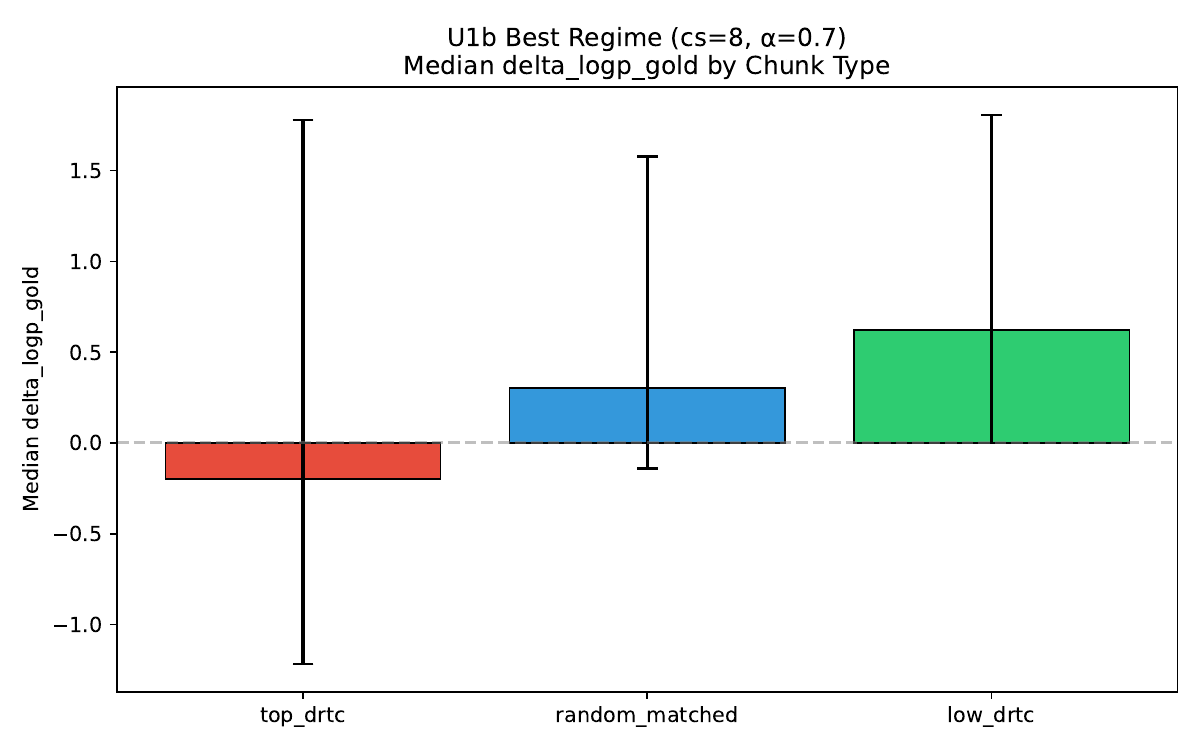}
  \caption{Best-regime effect sizes by chunk type (top/low/random) with bootstrap CIs.}
  \label{fig:u1b_best_regime}
\end{figure}

\begin{figure}[H]
  \centering
  \includegraphics[width=\linewidth]{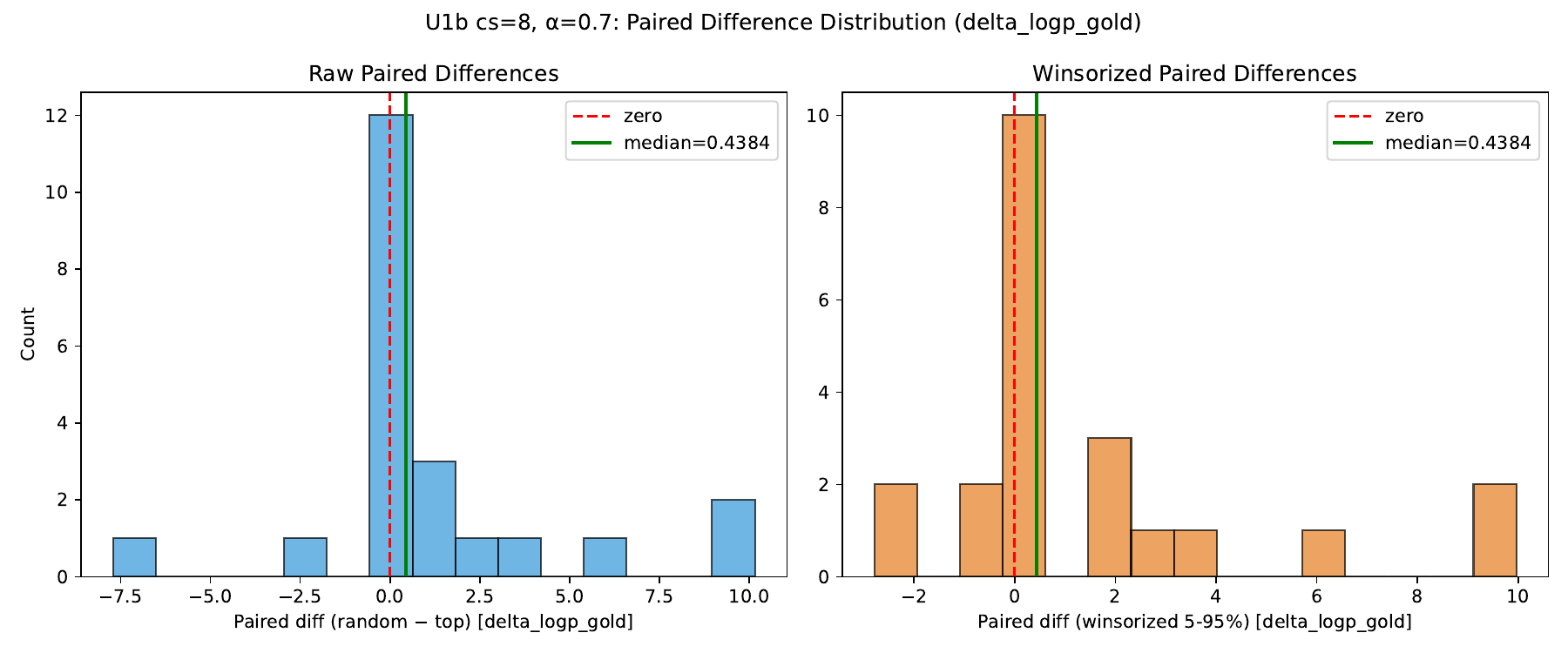}
  \caption{Distribution of paired differences for $\Delta\log p(\text{gold})$ under strict quartile matching.}
  \label{fig:u1b_paired_diff}
\end{figure}

\section{Qualitative case studies}
\label{app:qual_case_studies}

\paragraph{Reading guide.}
Each case study reports: (i) problem and gold answer, (ii) top DRTC-contributing chunks (with excerpts), (iii) opposing-sign chunks, and (iv) large-magnitude curvature diagnostic chunks (CurvImpact).
DRTC is \emph{process-causal}: a chunk’s signed score indicates whether its pivot-local masking intervention pushes the pivot distribution \emph{toward} (positive) or \emph{away from} (negative) the realized rollout direction.
CurvImpact is a \emph{diagnostic} turning-angle signal computed in raw logit space and is not used to define pivots or compute DRTC.
All interpretations are conservative: we do not make outcome-causal claims about correctness.

\subsection{LFM2.5-1.2B}
\label{app:lfm_cases}

\subsubsection{\texttt{math\_rollout\_2312} (LFM2.5-1.2B)}
\paragraph{Problem.}
A nonzero polynomial with rational coefficients has all of the numbers
$1+\sqrt{2}, 2+\sqrt{3}, 3+\sqrt{4}, \dots, 1000+\sqrt{1001}$ as roots.
\paragraph{Gold answer.} 1970.

\begin{figure}[h]
  \centering
  \begin{subfigure}[t]{0.48\textwidth}
    \centering
    \includegraphics[width=\linewidth]{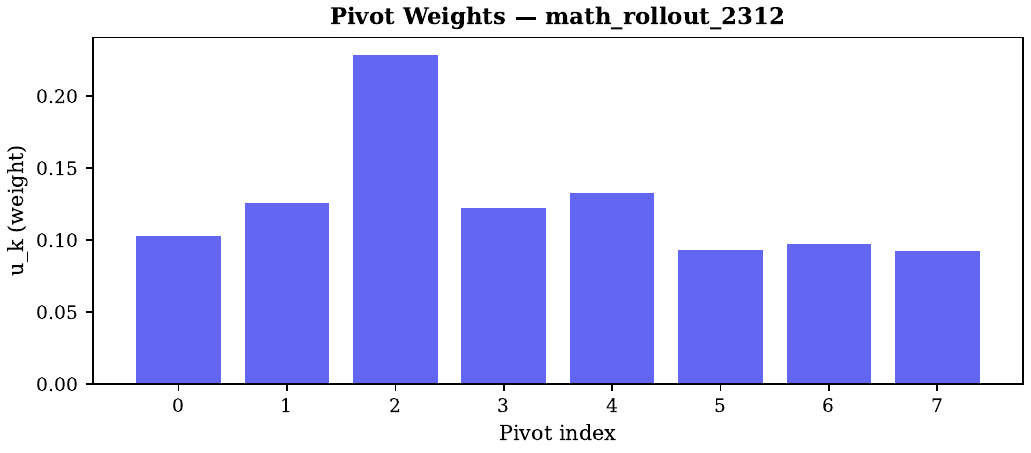}
    \caption{\scriptsize Pivot-weight distribution ($u_k$) for \texttt{math\_rollout\_2312} (LFM2.5-1.2B)}
  \end{subfigure}
  \hfill
  \begin{subfigure}[t]{0.48\textwidth}
    \centering
    \includegraphics[width=\linewidth]{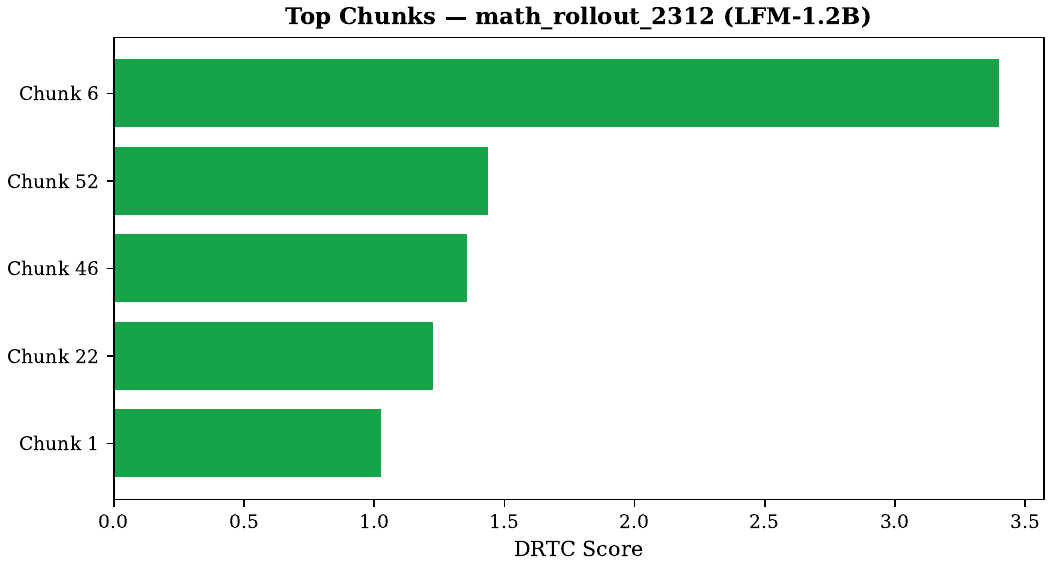}
    \caption{\scriptsize Top chunk attributions (DRTC) for \texttt{math\_rollout\_2312} (LFM2.5-1.2B)}
  \end{subfigure}
  \caption{Case-study visual summary for \texttt{math\_rollout\_2312} (LFM2.5-1.2B).}
  \label{fig:lfm_2312_visuals}
\end{figure}

\begin{table*}[h]
  \caption{
  LFM2.5-1.2B, \texttt{math\_rollout\_2312}: top contributing chunks by DRTC.
  Positive values align with the realized rollout direction; negative values oppose it.
  CurvImpact is a geometric diagnostic.
  }
  \label{tab:lfm_2312_top}
  \centering
  \scriptsize
  \setlength{\tabcolsep}{4pt}
  \renewcommand{\arraystretch}{1.15}
  \begin{tabularx}{\textwidth}{r c r r r Y}
    \toprule
    Rank & Chunk idx & DRTC & $|\mathrm{DRTC}|$ & CurvImpact & Excerpt \\
    \midrule
    1 & 6  & +3.4020 & 3.4020 & 0.0080 & polynomial has rational coefficients, then any irrational root must come with \dots \\
    2 & 0  & -1.9895 & 1.9895 & 0.0189 & \texttt{<think>} Okay, let me try to tackle this problem. So the question is \\
    3 & 52 & +1.4424 & 1.4424 & 0.0154 & 1000 of these quadratic factors. But wait, maybe some of these \\
    \bottomrule
  \end{tabularx}
\end{table*}

\paragraph{Opposing-sign chunks (directional tension).}
Chunk 6 and chunk 52 have positive DRTC, meaning that (after pivot weighting and relevance gating) their information flow
tends to redirect pivot log-probability vectors in the same direction as the realized rollout.
Chunk 0 has negative DRTC, indicating that masking this early preamble makes the pivot-local change
\emph{more} aligned with the realized direction $g$; this is consistent with generic orientation text contributing weak,
possibly noisy steering relative to later problem-specific constraints.

\paragraph{Curvature diagnostic (geometry, not attribution).}
CurvImpact is computed from changes in turning angles in raw logit space under the same pivot-local masking interventions.
Large $|\mathrm{CurvImpact}|$ indicates interventions that materially change local trajectory geometry (reorientation intensity),
but this signal is diagnostic and does not define causal importance or the sign of steering.

\paragraph{Qualitative interpretation.}
Chunk 6 (the conjugate-root constraint) has the largest aggregated DRTC, i.e., the strongest pivot-weighted, gated directional
effect among chunks in this rollout.
This matches the core mathematical requirement for rational-coefficient polynomials: irrational roots must appear with their
conjugates, except when $\sqrt{n+1}$ is integer (yielding a rational root).
Given that the realized rollout is correct here, high positive DRTC chunks provide a compact shortlist of trace segments that
most strongly support the realized solution path.

\subsubsection{\texttt{math\_rollout\_2757} (LFM2.5-1.2B)}
\paragraph{Problem.}
\begin{quote}\small
Let $c$ be a complex number. Suppose there exist distinct complex numbers $r$, $s$, and $t$ such that for every complex
number $z$, we have
\[
  (z - r)(z - s)(z - t) = (z - cr)(z - cs)(z - ct).
\]
Compute the number of distinct possible values of $c$.
\end{quote}

\paragraph{Gold answer.}
\begin{quote}\small
$4$
\end{quote}

\begin{figure}[h]
  \centering
  \begin{subfigure}[t]{0.48\textwidth}
    \centering
    \includegraphics[width=\linewidth]{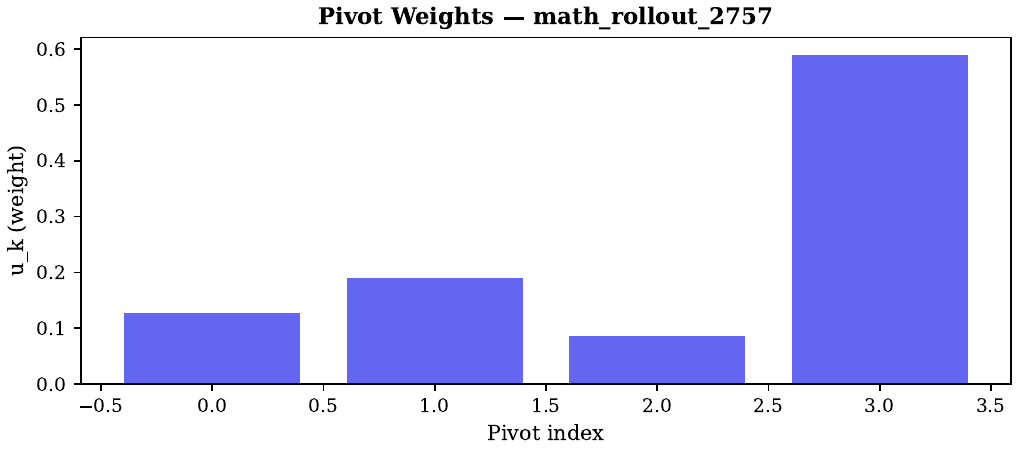}
    \caption{\scriptsize Pivot-weight distribution ($u_k$) for \texttt{math\_rollout\_2757} (LFM2.5-1.2B)}
  \end{subfigure}
  \hfill
  \begin{subfigure}[t]{0.48\textwidth}
    \centering
    \includegraphics[width=\linewidth]{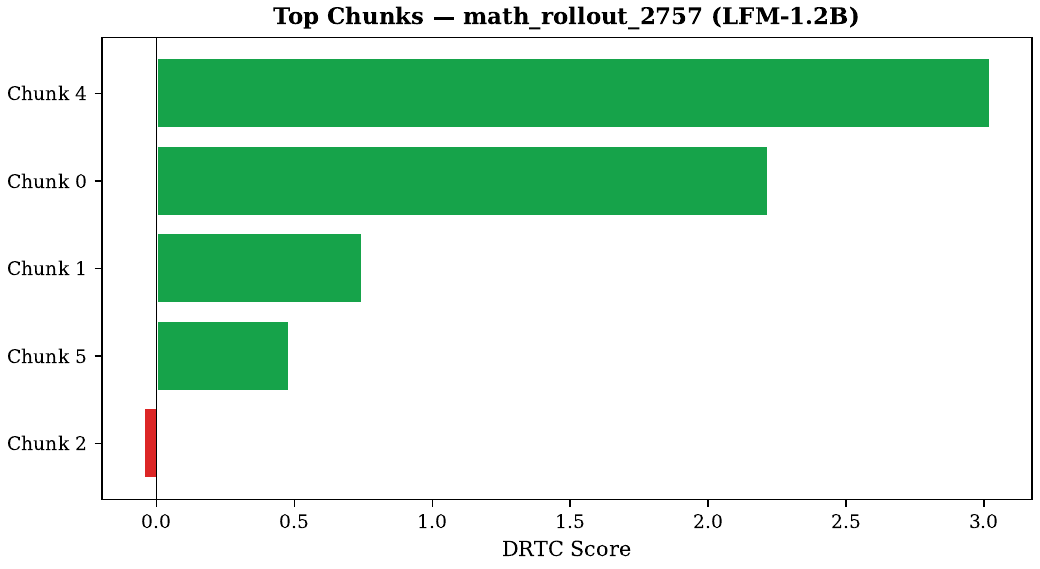}
    \caption{\scriptsize Top chunk attributions (DRTC) for \texttt{math\_rollout\_2757} (LFM2.5-1.2B)}
  \end{subfigure}
  \caption{Case-study visual summary for \texttt{math\_rollout\_2757} (LFM2.5-1.2B).}
  \label{fig:lfm_2757_visuals}
\end{figure}

\begin{table*}[h]
  \caption{
  LFM2.5-1.2B, \texttt{math\_rollout\_2757}: top contributing chunks by DRTC.
  Positive values align with the realized rollout direction; negative values oppose it.
  CurvImpact is a geometric diagnostic.
  }
  \label{tab:lfm_2757_top}
  \centering
  \scriptsize
  \setlength{\tabcolsep}{4pt}
  \renewcommand{\arraystretch}{1.15}
  \begin{tabularx}{\textwidth}{r c r r r Y}
    \toprule
    Rank & Chunk idx & DRTC & $|\mathrm{DRTC}|$ & CurvImpact & Excerpt \\
    \midrule
    1 & 4 & +3.0206 & 3.0206 & 0.0813 & )(z - cs)(z - ct). We need to find the number \\
    2 & 0 & +2.2185 & 2.2185 & 0.0106 & \texttt{<think>} Okay, let me try to tackle this problem. So the question says \\
    3 & 1 & +0.7457 & 0.7457 & 0.0141 & Let c be a complex number. Suppose there exist distinct complex numbers r, \\
    4 & 5 & +0.4803 & 0.4803 & 0.1017 & of distinct possible values of c, and then give the final answer starting with ' \\
    5 & 3 & -0.4079 & 0.4079 & -0.0106 & product (z - r)(z - s)(z - t) equals ( \\
    6 & 2 & -0.0450 & 0.0450 & -0.0060 & numbers r, s, and t such that for every complex number z, the \\
    \bottomrule
  \end{tabularx}
\end{table*}

\paragraph{Opposing-sign chunks (directional tension).}
Chunks 4, 0, and 1 have positive DRTC, indicating that—after pivot weighting and relevance gating—their information flow
tends to steer pivot-local distributions in the same direction as the realized rollout.
Chunk 3 has negative DRTC, meaning that masking it makes the pivot-local trajectory more aligned with the realized direction $g$.
This suggests that portions of the raw polynomial equality restatement exert mild counter-steering relative to the algebraic constraint structure emphasized later in the trace.

\paragraph{Curvature diagnostic (geometry, not attribution).}
CurvImpact measures how pivot-local masking alters turning angles in raw logit space.
Chunk 4 and chunk 5 exhibit comparatively large $|\mathrm{CurvImpact}|$, indicating substantial local geometric reorientation under intervention.
As in other case studies, curvature reflects reorientation intensity rather than directional support, and does not determine causal sign.

\paragraph{Qualitative interpretation.}
The highest-ranked positive chunk (chunk 4) occurs at the transition where the model reframes the equality
\[
(z - r)(z - s)(z - t) = (z - cr)(z - cs)(z - ct)
\]
as a structural constraint on roots.
For two monic cubic polynomials to be identical for all $z$, their root multisets must coincide.
Thus $\{r,s,t\}$ must equal $\{cr,cs,ct\}$ as a set.
High positive DRTC on this structural equivalence step is consistent with it functioning as the key reasoning move in the rollout.

\subsection{Ministral-3-3B}
\label{app:ministral_cases}

\subsubsection{\texttt{math\_rollout\_3295} (Ministral-3-3B)}
\paragraph{Problem.}
\begin{quote}\small
If $f(x) = \frac{1 + x}{1 - 3x},\ f_1(x) = f(f(x)),\ f_2(x) = f(f_1(x)),$ and in general
$f_n(x) = f(f_{n-1}(x)),$ then $f_{1993}(3)=$
\end{quote}

\paragraph{Gold answer.}
\begin{quote}\small
$\frac{1}{5}$
\end{quote}

\begin{figure}[h]
  \centering
  \begin{subfigure}[t]{0.48\textwidth}
    \centering
    \includegraphics[width=\linewidth]{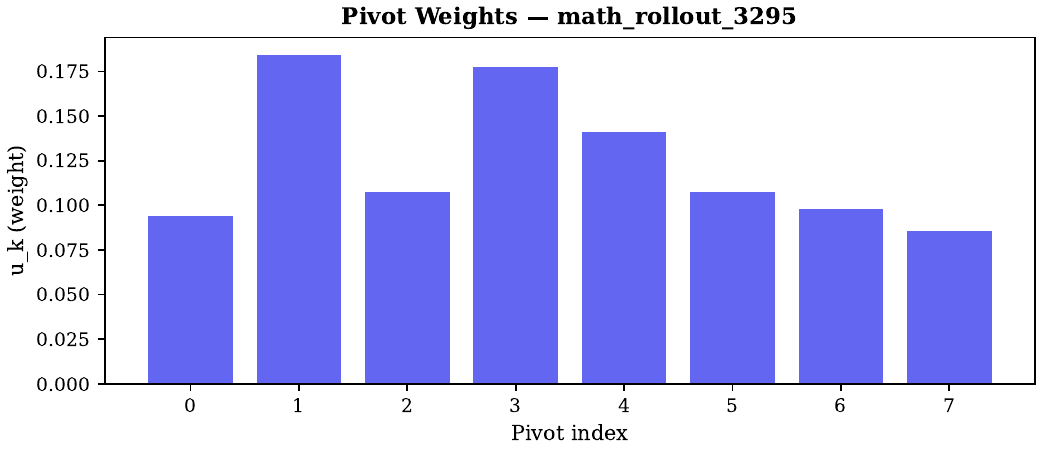}
    \caption{\scriptsize Pivot-weight distribution ($u_k$) for \texttt{math\_rollout\_3295} (Ministral-3-3B)}
  \end{subfigure}
  \hfill
  \begin{subfigure}[t]{0.48\textwidth}
    \centering
    \includegraphics[width=\linewidth]{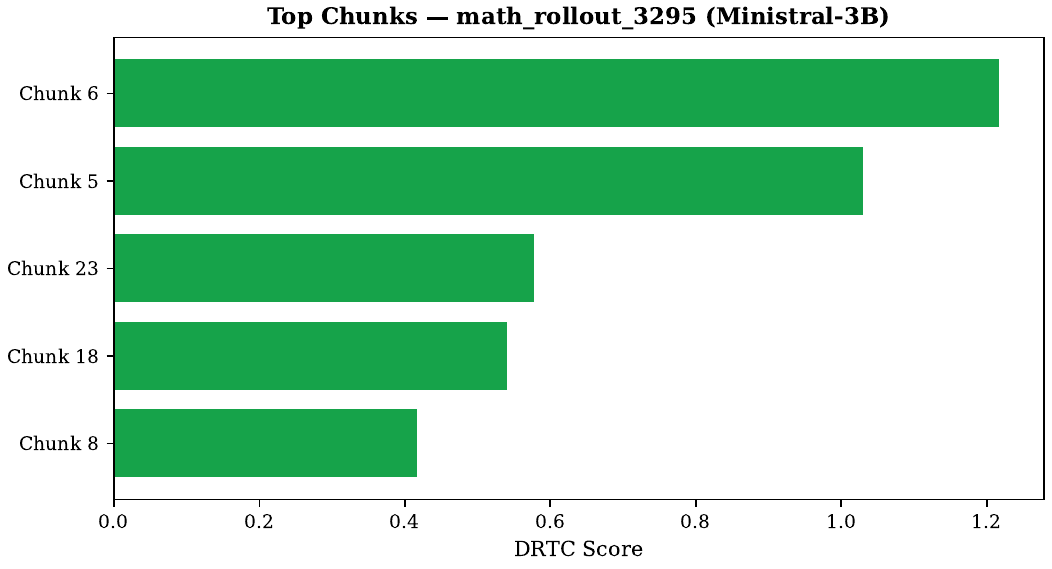}
    \caption{\scriptsize Top chunk attributions (DRTC) for \texttt{math\_rollout\_3295} (Ministral-3-3B)}
  \end{subfigure}

  \caption{
  Case-study visual summary for \texttt{math\_rollout\_3295} (Ministral-3-3B).
  Left: pivot importance weights $u_k$ across the $K$ detected pivots.
  Right: top-ranked chunks by signed DRTC (process-causal directional steering), shown for qualitative auditing.
  }
  \label{fig:min_3295_visuals}
\end{figure}

\begin{table*}[h]
\caption{
Ministral-3-3B, \texttt{math\_rollout\_3295}: top contributing chunks by DRTC.
Positive values align with the realized rollout direction; negative values oppose it.
CurvImpact is a geometric diagnostic.
}
\label{tab:min_3295_top}
\centering
\scriptsize
\setlength{\tabcolsep}{4pt}
\renewcommand{\arraystretch}{1.15}
\begin{tabularx}{\textwidth}{r c r r r Y}
\toprule
Rank & Chunk idx & DRTC & $|\mathrm{DRTC}|$ & CurvImpact & Excerpt \\
\midrule
1  & 9  & -3.5838 & 3.5838 & -0.0238 & at some point, maybe x=3, but maybe it's better \\
2  & 0  & -2.6963 & 2.6963 & 0.0026  & Okay, let's try to solve this problem. We have a function f \\
3  & 10 & -2.1506 & 2.1506 & 0.0609  & to compute the general form of f\_n(x) first. Let \\
4  & 2  & -1.8071 & 1.8071 & 0.0018  & find f\_1993(3), where f \\
5  & 37 & -1.3859 & 1.3859 & -0.0013 & \{1 + x - 1/1 + 3x\}\{ \\
6  & 6  & +1.2177 & 1.2177 & -0.0156 & maybe I should compute the first few iterates to see if there's a \\
7  & 7  & -1.0832 & 1.0832 & 0.0281  & pattern. Let's start with f\_1(x) = f(f(x \\
8  & 5  & +1.0313 & 1.0313 & 0.0010  & (x) = f(f\_1(x)), and so on. First \\
9  & 38 & -1.0309 & 1.0309 & -0.0015 & 1 - 3 * \{x - 1\}\{1 + \\
10 & 3  & -0.6867 & 0.6867 & -0.0011 & f\_n(x) is the nth iterate of the function, meaning \\
\bottomrule
\end{tabularx}
\end{table*}

\paragraph{Opposing-sign chunks (directional tension).}
This rollout is dominated by \emph{negative} DRTC among the top-ranked chunks (e.g., chunks 9, 0, 10, 2, 37),
meaning that masking these segments makes the pivot-local change \emph{more} aligned with the realized rollout direction $g$
(after pivot weighting and relevance gating).
The few positive chunks (notably chunks 6 and 5) indicate segments whose information flow steers pivot distributions
\emph{along} the realized direction, and thus act as the main ``supporting'' influences under DRTC.

\paragraph{Curvature diagnostic (geometry, not attribution).}
CurvImpact reflects intervention-induced turning-angle changes in raw logit space.
Here, several chunks with large $|\mathrm{CurvImpact}|$ (e.g., chunk 10 and chunk 7) coincide with moments where the trace
shifts from informal planning toward explicit functional iteration/algebra, consistent with curvature capturing local reorientation
intensity rather than directional sign.
As elsewhere, CurvImpact is diagnostic and does not determine causal importance or steering direction.

\paragraph{Qualitative interpretation.}
The most negative chunk (chunk 9: ``maybe $x=3$, but maybe it's better \dots'') appears to reflect approach uncertainty
and replanning; under DRTC, such segments counter-steer relative to the final realized solution direction.
In contrast, the positive chunks (chunks 6 and 5) correspond to concretely computing early iterates and looking for a pattern,
which matches the structure of the correct solution for this problem (a short periodic cycle under iteration).
Overall, the sign pattern is consistent with a trace that contains nontrivial planning/hesitation, with the eventual correct branch
supported by a smaller number of pattern-extraction steps.

\subsubsection{\texttt{math\_rollout\_3374} (Ministral-3-3B)}
\paragraph{Problem.}
\begin{quote}\small
When the base-16 number $66666_{16}$ is written in base 2, how many base-2 digits (bits) does it have?
\end{quote}

\paragraph{Gold answer.}
\begin{quote}\small
$19$
\end{quote}

\begin{figure}[h]
  \centering
  \begin{subfigure}[t]{0.48\textwidth}
    \centering
    \includegraphics[width=\linewidth]{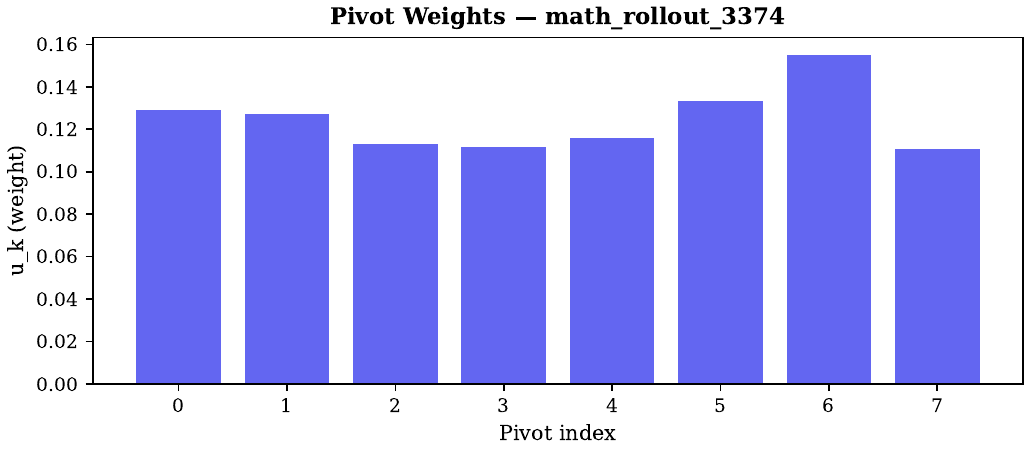}
    \caption{\scriptsize Pivot-weight distribution ($u_k$) for \texttt{math\_rollout\_3374} (Ministral-3-3B)}
  \end{subfigure}
  \hfill
  \begin{subfigure}[t]{0.48\textwidth}
    \centering
    \includegraphics[width=\linewidth]{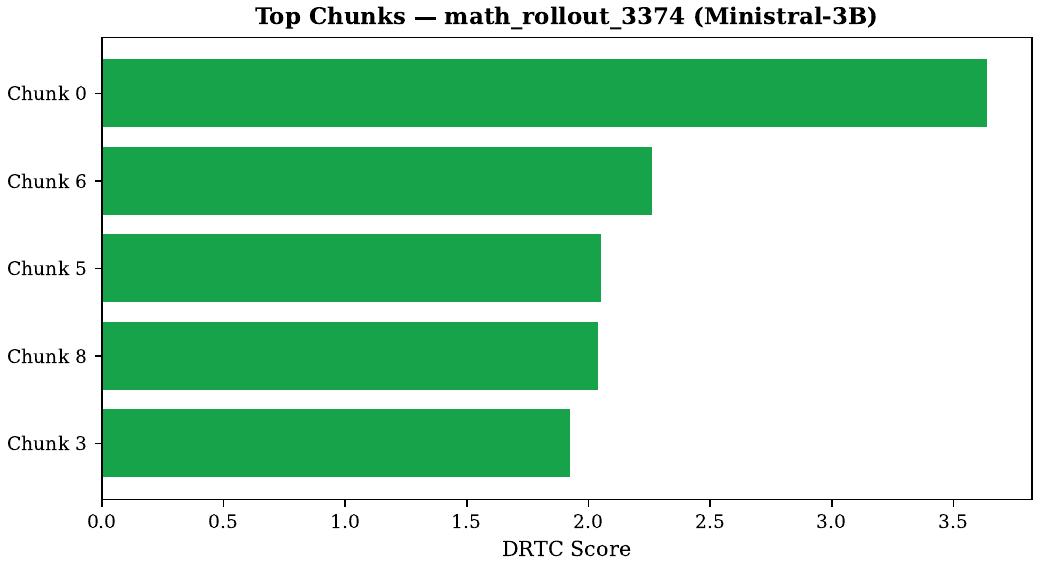}
    \caption{\scriptsize Top chunk attributions (DRTC) for \texttt{math\_rollout\_3374} (Ministral-3-3B)}
  \end{subfigure}

  \caption{
  Case-study visual summary for \texttt{math\_rollout\_3374} (Ministral-3-3B).
  Left: pivot importance weights $u_k$ across the $K$ detected pivots.
  Right: top-ranked chunks by signed DRTC (process-causal directional steering), shown for qualitative auditing.
  }
  \label{fig:min_3374_visuals}
\end{figure}

\begin{table*}[h]
  \caption{
  Ministral-3-3B, \texttt{math\_rollout\_3374}: top contributing chunks by DRTC.
  Positive values align with the realized rollout direction; negative values oppose it.
  CurvImpact is a geometric diagnostic.
  }
  \label{tab:min_3374_top}
  \centering
  \scriptsize
  \setlength{\tabcolsep}{4pt}
  \renewcommand{\arraystretch}{1.15}
  \begin{tabularx}{\textwidth}{r c r r r Y}
    \toprule
    Rank & Chunk idx & DRTC & $|\mathrm{DRTC}|$ & CurvImpact & Excerpt \\
    \midrule
    1  & 0  & +3.6400 & 3.6400 & -0.0593 & Okay, let's see. The problem is asking how many base-2 digits \\
    2  & 6  & +2.2631 & 2.2631 & -0.0151 & bits. First, let's confirm that. Yes, in base-16 \\
    3  & 5  & +2.0559 & 2.0559 & 0.0039  & $16 = 2^4$. So, each hex digit is 4 \\
    4  & 8  & +2.0423 & 2.0423 & -0.0045 & to convert $66666_{16}$ to base-2 \\
    5  & 3  & +1.9297 & 1.9297 & -0.0538 & Hmm, first, I need to remember how to convert from base-1 \\
    6  & 13 & +1.6693 & 1.6693 & 0.0007  & 1010 to 1111. So, 6 \\
    7  & 24 & +1.4739 & 1.4739 & 0.0012  & 's 5 digits. Each hex digit is 4 bits, so 5 \\
    8  & 4  & -1.3760 & 1.3760 & 0.0035  & 6 to base-2. I recall that each hexadecimal digit corresponds to \\
    9  & 30 & +1.2920 & 1.2920 & 0.0012  & are 0110, so the total number of bits should be \\
    10 & 25 & +1.2566 & 1.2566 & -0.0011 & digits $\times$ 4 bits per digit = 20 bits. But wait, \\
    \bottomrule
  \end{tabularx}
\end{table*}

\paragraph{Opposing-sign chunks (directional tension).}
Most top-ranked chunks have positive DRTC (e.g., chunks 0, 6, 5, 8, 3, 13, 24, 30, 25), indicating that—after pivot weighting
and relevance gating—their information flow tends to redirect pivot-local distributions in the same direction as the realized rollout.
The highest-magnitude negative chunk (chunk 4) suggests a local segment whose contribution, when present, mildly steers
opposite to the realized direction $g$; here this is plausibly generic conversion boilerplate rather than a problem-specific constraint.

\paragraph{Curvature diagnostic (geometry, not attribution).}
CurvImpact reflects turning-angle changes in raw logit space under the same pivot-local masking interventions.
Several early chunks (e.g., chunk 0 and chunk 3) show relatively larger $|\mathrm{CurvImpact}|$, consistent with curvature
capturing local reorientation as the trace shifts from problem restatement to the base-conversion strategy.
As elsewhere, curvature is diagnostic and does not determine causal sign.

\paragraph{Qualitative interpretation.}
The dominant positive chunks center on the key structural fact $16=2^4$ (chunk 5), which yields the baseline rule
``each hex digit corresponds to 4 bits'' and drives the initial estimate of $5\times 4=20$ bits (chunks 24--25).
The decisive correction is that the leading hex digit is $6$ (binary $0110$), which does not contribute a leading $1$ in the
most significant position; thus the binary representation does not require the full 4 bits for the first digit.
The high positive DRTC mass on these conversion-rule and correction steps is consistent with them forming the core
solution logic in the realized rollout.

\subsection{Phi-4-Mini}
\label{app:phi_cases}

\subsubsection{\texttt{math\_rollout\_2757} (Phi-4-Mini)}
\paragraph{Problem.}
\begin{quote}\small
Let $c$ be a complex number. Suppose there exist distinct complex numbers $r$, $s$, and $t$ such that for every complex
number $z$, we have
\[
  (z - r)(z - s)(z - t) = (z - cr)(z - cs)(z - ct).
\]
Compute the number of distinct possible values of $c$.
\end{quote}

\paragraph{Gold answer.}
\begin{quote}\small
$4$
\end{quote}

\begin{figure}[h]
  \centering
  \begin{subfigure}[t]{0.48\textwidth}
    \centering
    \includegraphics[width=\linewidth]{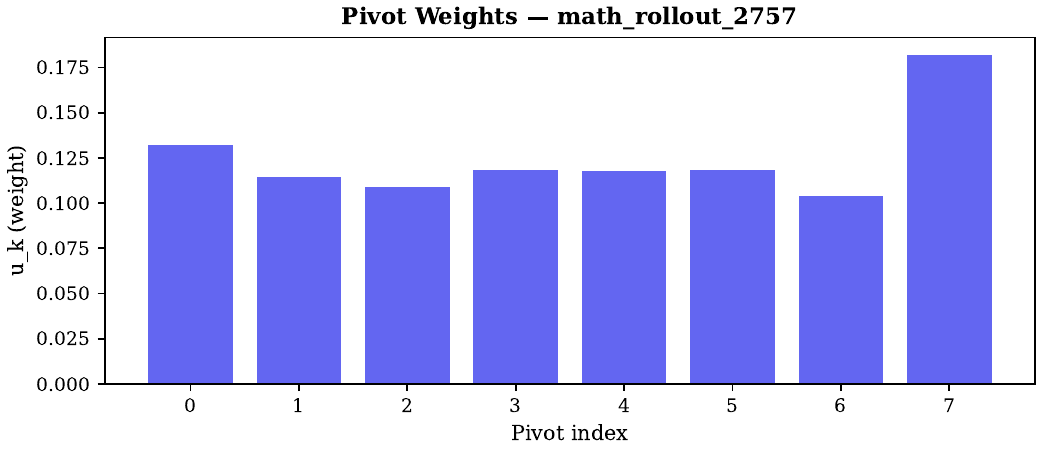}
    \caption{\scriptsize Pivot-weight distribution ($u_k$) for \texttt{math\_rollout\_2757} (Phi-4-Mini)}
  \end{subfigure}
  \hfill
  \begin{subfigure}[t]{0.48\textwidth}
    \centering
    \includegraphics[width=\linewidth]{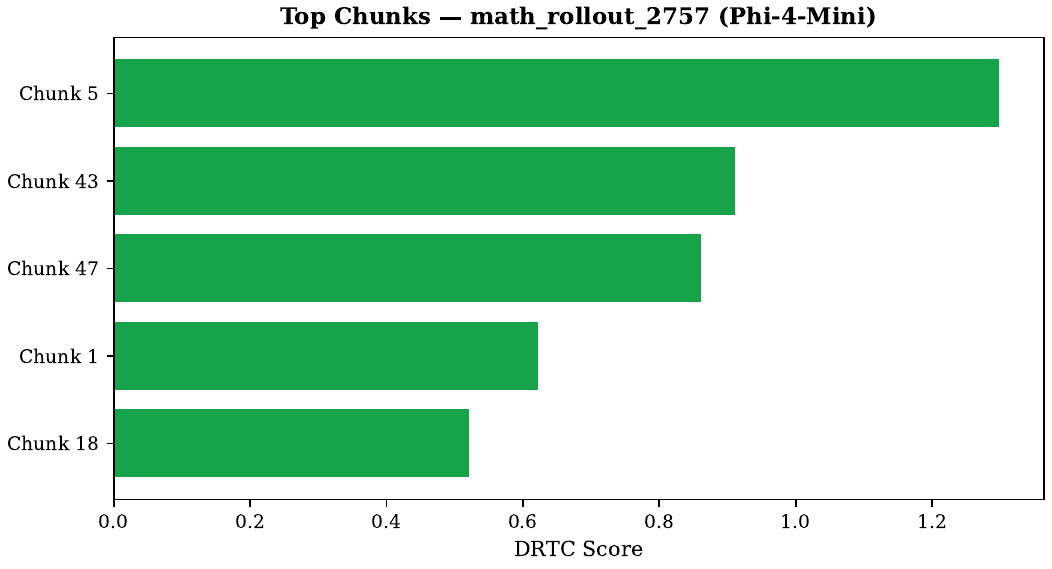}
    \caption{\scriptsize Top chunk attributions (DRTC) for \texttt{math\_rollout\_2757} (Phi-4-Mini)}
  \end{subfigure}

  \caption{
  Case-study visual summary for \texttt{math\_rollout\_2757} (Phi-4-Mini).
  Left: pivot importance weights $u_k$ across the $K$ detected pivots.
  Right: top-ranked chunks by signed DRTC (process-causal directional steering), shown for qualitative auditing.
  }
  \label{fig:phi_2757_visuals}
\end{figure}

\begin{table*}[h]
  \caption{
  Phi-4-Mini, \texttt{math\_rollout\_2757}: top contributing chunks by DRTC.
  Positive values align with the realized rollout direction; negative values oppose it.
  CurvImpact is a geometric diagnostic.
  }
  \label{tab:phi_2757_top}
  \centering
  \scriptsize
  \setlength{\tabcolsep}{4pt}
  \renewcommand{\arraystretch}{1.15}
  \begin{tabularx}{\textwidth}{r c r r r Y}
    \toprule
    Rank & Chunk idx & DRTC & $|\mathrm{DRTC}|$ & CurvImpact & Excerpt \\
    \midrule
    1  & 6  & -3.4363 & 3.4363 & -0.0218 & down step by step. First, both sides of the equation are cubic polynomials \\
    2  & 10 & -3.0111 & 3.0111 & 0.1074  & and set the coefficients equal, I can find conditions on c, r, s \\
    3  & 16 & -2.3690 & 2.3690 & -0.0138 & + (cs r + ctr + crt)z - crt$^3$. Wait, \\
    4  & 19 & -1.9486 & 1.9486 & 0.0502  & st) with c factored in, so (cr * cs + cr * \\
    5  & 3  & -1.6787 & 1.6787 & 0.0036  & )(z - s)(z - t) is equal to (z - cr \\
    6  & 17 & -1.6302 & 1.6302 & 0.0105  & actually, let me check that again. The coefficient for the cubic term is \\
    7  & 4  & -1.6081 & 1.6081 & 0.0063  & )(z - cs)(z - ct). The question is asking for the number \\
    8  & 2  & -1.4205 & 1.4205 & 0.0123  & t such that for every complex number z, the polynomial (z - r \\
    9  & 5  & +1.2991 & 1.2991 & 0.0038  & of distinct possible values of c. Hmm, interesting. Let me try to break \\
    10 & 8  & -1.2564 & 1.2564 & 0.0114  & coefficients must be equal. That makes sense because two polynomials that are equal everywhere \\
    \bottomrule
  \end{tabularx}
\end{table*}

\paragraph{Opposing-sign chunks (directional tension).}
Most top-ranked chunks have \emph{negative} DRTC (e.g., chunks 6, 10, 16, 19, 3, 17, 4, 2, 8), indicating that---after pivot weighting
and relevance gating---their information flow tends to redirect pivot-local distributions \emph{opposite} to the realized rollout direction $g$.
The main positive contributor (chunk 5) suggests a segment whose information flow supports the realized directional progression at the pivots that matter.
As in all case studies, this sign pattern is about \emph{directional steering relative to the realized rollout}, not an outcome-level correctness claim.

\paragraph{Curvature diagnostic (geometry, not attribution).}
CurvImpact reflects turning-angle changes in raw logit space under the same pivot-local masking interventions.
Chunks with larger $|\mathrm{CurvImpact}|$ (notably chunk 10) occur near coefficient-expansion / ``re-check'' transitions, consistent with curvature capturing
local reorientation intensity as the trace shifts between symbolic manipulation modes.
Curvature is diagnostic and does not determine causal sign.

\paragraph{Qualitative interpretation.}
The dominant negative chunks align with algebraic expansion and coefficient-matching subtraces (chunks 6, 10, 16, 19, 17), which in this rollout appear to
counter-steer pivot distributions relative to $g$ after weighting/gating.
The main positive chunk (chunk 5) occurs at a higher-level reframing step, consistent with ``structural'' reasoning moves providing the net directional support in the realized trace.
We interpret this conservatively: DRTC is isolating which segments are directionally aligned with the realized solution path under pivot-local causal probes, not diagnosing correctness of any sub-derivation.

\subsubsection{\texttt{math\_rollout\_3759} (Phi-4-Mini)}
\paragraph{Problem.}
\begin{quote}\small
The smaller square in the figure below has a perimeter of $4$ cm, and the larger square has an area of $16$
$\text{cm}^2$. What is the distance from point $A$ to point $B$? Express your answer as a decimal to the nearest tenth.

\begin{verbatim}
[asy]
draw((0,0)--(12,0));
draw((2,0)--(2,10));
draw((0,0)--(0,2));
draw((0,2)--(2,2));
draw((0,2)--(12,10));
draw((12,0)--(12,10));
draw((2,10)--(12,10));
label("B",(0,2),W);
label("A",(12,10),E);
[/asy]
\end{verbatim}
\end{quote}

\paragraph{Gold answer.}
\begin{quote}\small
$5.8$
\end{quote}

\begin{figure}[h]
  \centering
  \begin{subfigure}[t]{0.48\textwidth}
    \centering
    \includegraphics[width=\linewidth]{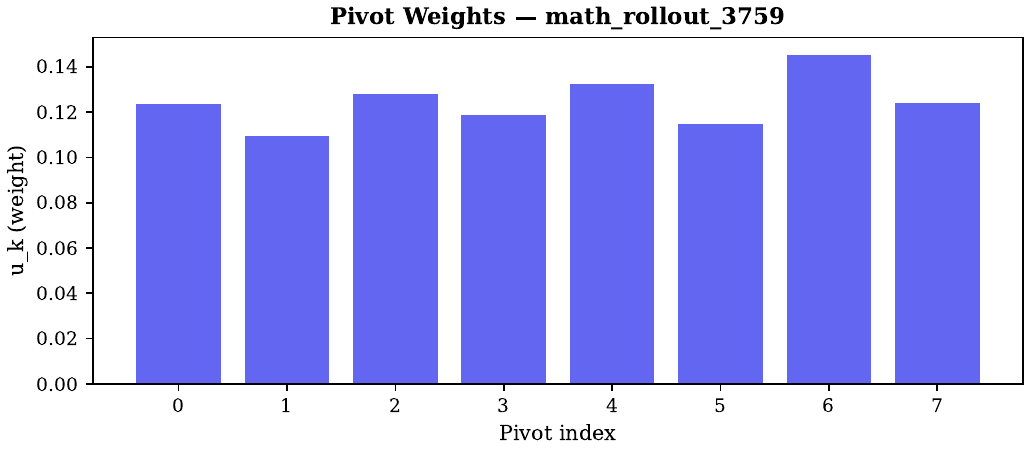}
    \caption{\scriptsize Pivot-weight distribution ($u_k$) for \texttt{math\_rollout\_3759} (Phi-4-Mini)}
  \end{subfigure}
  \hfill
  \begin{subfigure}[t]{0.48\textwidth}
    \centering
    \includegraphics[width=\linewidth]{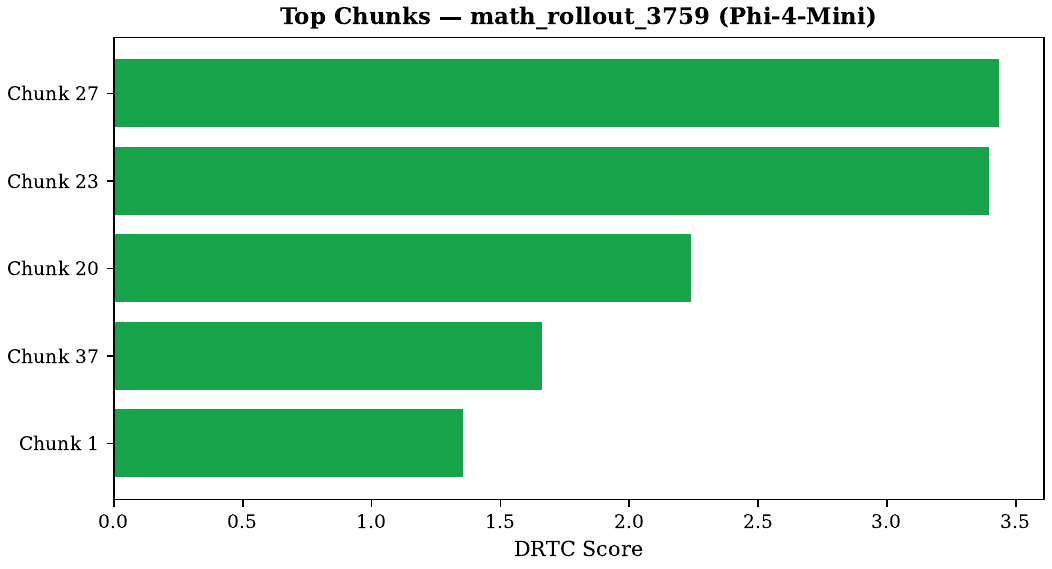}
    \caption{\scriptsize Top chunk attributions (DRTC) for \texttt{math\_rollout\_3759} (Phi-4-Mini)}
  \end{subfigure}

  \caption{
  Case-study visual summary for \texttt{math\_rollout\_3759} (Phi-4-Mini).
  Left: pivot importance weights $u_k$ across the $K$ detected pivots.
  Right: top-ranked chunks by signed DRTC (process-causal directional steering), shown for qualitative auditing.
  }
  \label{fig:phi_3759_visuals}
\end{figure}

\begin{table*}[h]
  \caption{
  Phi-4-Mini, \texttt{math\_rollout\_3759}: top contributing chunks by DRTC.
  Positive values align with the realized rollout direction; negative values oppose it.
  CurvImpact is a geometric diagnostic.
  }
  \label{tab:phi_3759_top}
  \centering
  \scriptsize
  \setlength{\tabcolsep}{4pt}
  \renewcommand{\arraystretch}{1.15}
  \begin{tabularx}{\textwidth}{r c r r r Y}
    \toprule
    Rank & Chunk idx & DRTC & $|\mathrm{DRTC}|$ & CurvImpact & Excerpt \\
    \midrule
    1  & 27 & +3.4374 & 3.4374 & 0.1217 & ) and up to (0,2) and (2,2). So \\
    2  & 23 & +3.3994 & 3.3994 & -0.0892 & Wait, the coordinates in the Asymptote code might be scaled. Because \\
    3  & 20 & +2.2437 & 2.2437 & 0.0044 & 12,10), which is probably connecting point B to point A. There's also \\
    4  & 37 & +1.6648 & 1.6648 & 0.1015 & the Asymptote code is 0.5 cm in reality. Wait \\
    5  & 24 & -1.3774 & 1.3774 & 0.0098 & might be scaled. Because the smaller square has a perimeter of 4 cm, \\
    6  & 1  & +1.3597 & 1.3597 & -0.0041 & find the distance from point A to point B in the given figure. The problem \\
    7  & 16 & +1.3152 & 1.3152 & -0.0089 & to (12,0), which is a horizontal line 12 units long. \\
    8  & 14 & -1.0863 & 1.0863 & 0.0137 & it seems like the figure is drawn with coordinates. Let me try to parse \\
    9  & 39 & +1.0375 & 1.0375 & 0.1260 & of 4 cm. So in reality, each side is 1 cm. \\
    10 & 38 & -0.8626 & 0.8626 & 0.0084 & but maybe not. Let me think again. Wait, the problem says the \\
    \bottomrule
  \end{tabularx}
\end{table*}

\paragraph{Opposing-sign chunks (directional tension).}
Most top-ranked chunks have positive DRTC (e.g., chunks 27, 23, 20, 37, 1, 16, 39), indicating that---after pivot weighting and relevance gating---their information flow
tends to steer pivot-local distributions in the same direction as the realized rollout.
The strongest negative contributors (chunks 24, 14, 38) correspond to moments of uncertainty about scale and diagram parsing, suggesting local segments whose inclusion
nudges the pivot-local trajectory away from $g$ (e.g., premature or inconsistent scaling hypotheses) before later correction.

\paragraph{Curvature diagnostic (geometry, not attribution).}
CurvImpact reflects turning-angle changes in raw logit space under the same pivot-local masking interventions.
Several high-$|\mathrm{CurvImpact}|$ chunks (notably chunks 27 and 39, also chunk 37) occur near the scale-setting and coordinate-to-centimeter conversion steps,
consistent with curvature capturing reorientation intensity as the trace shifts from diagram parsing to quantitative computation.
As elsewhere, curvature is diagnostic and does not determine causal sign.

\paragraph{Qualitative interpretation.}
The dominant positive chunks focus on recognizing that the Asymptote coordinates encode a \emph{scaled} drawing and that the given square measurements determine the scale factor
(chunks 23, 24, 37, 39).
Once the scale is fixed, the distance $AB$ is computed as a Euclidean distance between the plotted points, so the key reasoning burden is correctly mapping ``units in the code''
to centimeters in the problem statement.
The presence of negative DRTC around early scale speculation is consistent with a brief false start that is later corrected, while the large positive mass on the final scale-and-distance computation
aligns with the core solution logic in the realized rollout.

\subsection{R1-Distill-Qwen-1.5B}
\label{app:qwen_cases}

\subsubsection{\texttt{math\_rollout\_3295} (R1-Distill-Qwen-1.5B)}
\paragraph{Problem.}
\begin{quote}\small
If $f(x) = \frac{1 + x}{1 - 3x},\ f_1(x) = f(f(x)),\ f_2(x) = f(f_1(x)),$ and in general
$f_n(x) = f(f_{n-1}(x)),$ then $f_{1993}(3)=$
\end{quote}

\paragraph{Gold answer.}
\begin{quote}\small
$\frac{1}{5}$
\end{quote}

\begin{figure}[h]
  \centering
  \begin{subfigure}[t]{0.48\textwidth}
    \centering
    \includegraphics[width=\linewidth]{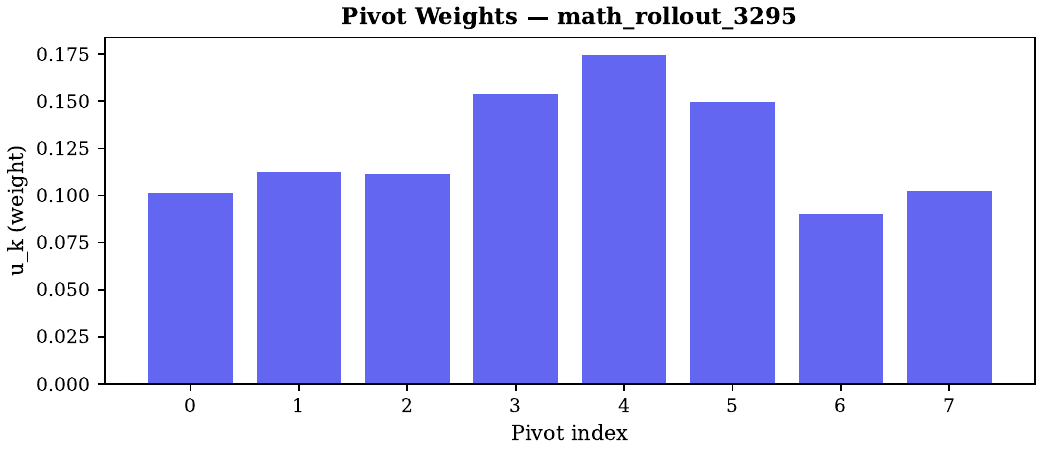}
    \caption{\scriptsize Pivot-weight distribution ($u_k$) for \texttt{math\_rollout\_3295} (R1-Distill-Qwen-1.5B)}
  \end{subfigure}
  \hfill
  \begin{subfigure}[t]{0.48\textwidth}
    \centering
    \includegraphics[width=\linewidth]{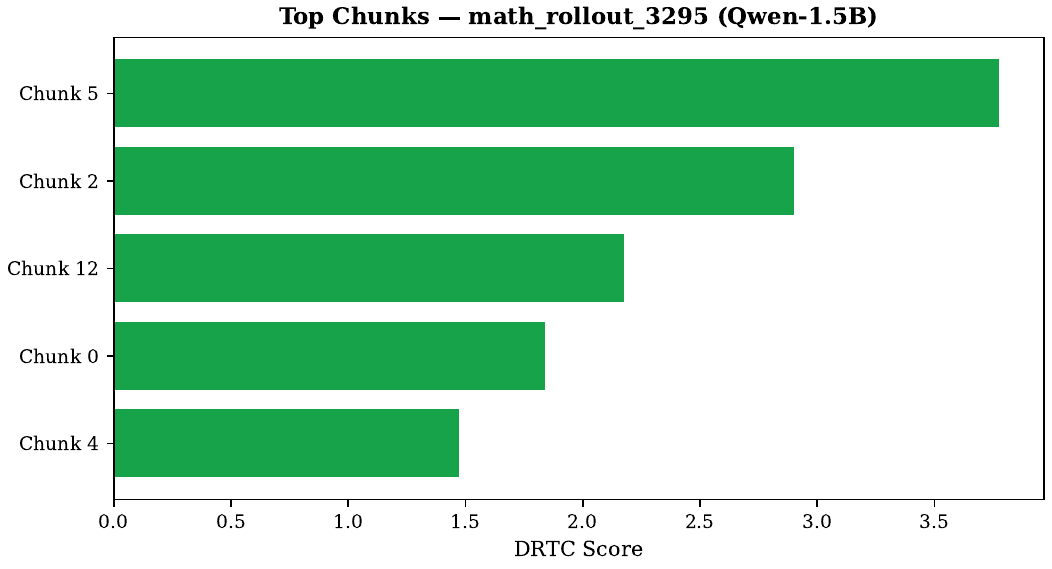}
    \caption{\scriptsize Top chunk attributions (DRTC) for \texttt{math\_rollout\_3295} (R1-Distill-Qwen-1.5B)}
  \end{subfigure}

  \caption{
  Case-study visual summary for \texttt{math\_rollout\_3295} (R1-Distill-Qwen-1.5B).
  Left: pivot importance weights $u_k$ across the $K$ detected pivots.
  Right: top-ranked chunks by signed DRTC (process-causal directional steering), shown for qualitative auditing.
  }
  \label{fig:qwen_3295_visuals}
\end{figure}

\begin{table*}[h]
\caption{
R1-Distill-Qwen-1.5B, \texttt{math\_rollout\_3295}: top contributing chunks by DRTC.
Positive values align with the realized rollout direction; negative values oppose it.
CurvImpact is a geometric diagnostic.
}
\label{tab:qwen_3295_top}
\centering
\scriptsize
\setlength{\tabcolsep}{4pt}
\renewcommand{\arraystretch}{1.15}
\begin{tabularx}{\textwidth}{r c r r r Y}
\toprule
Rank & Chunk idx & DRTC & $|\mathrm{DRTC}|$ & CurvImpact & Excerpt \\
\midrule
1  & 5  & +3.7780 & 3.7780 & -0.0409 & So, f\_n(x) is just f composed n times with itself. \\
2  & 2  & +2.9065 & 2.9065 & 0.0277  & + x)/(1 - 3x). Then, f\_1(x) \\
3  & 12 & +2.1814 & 2.1814 & 0.2078  & ). Wait, no, the denominator is 1 - 3x, so \\
4  & 0  & +1.8452 & 1.8452 & -0.0007 & problem where I need to find f\_1993(3). The \\
5  & 4  & +1.4766 & 1.4766 & -0.0255 & subsequent function being f composed with the previous one. So, f\_n(x) \\
6  & 8  & -1.4092 & 1.4092 & 0.0008  & what f\_1993(3) is without having to compute all \\
7  & 13 & +1.3089 & 1.3089 & -0.0349 & x)/(1 - 3x). Let me compute f\_1(x \\
8  & 1  & +1.0887 & 1.0887 & 0.0050  & function f(x) is given as (1 + x)/(1 - 3 \\
9  & 6  & -0.8988 & 0.8988 & 0.0062  & what this function f(x) looks like and how it behaves when composed multiple times \\
10 & 10 & +0.8265 & 0.8265 & -0.0023 & to see if I can spot a pattern. Starting with f(x) = ( \\
\bottomrule
\end{tabularx}
\end{table*}

\paragraph{Opposing-sign chunks (directional tension).}
Most top-ranked chunks have positive DRTC (e.g., chunks 5, 2, 12, 0, 4, 13, 1, 10), indicating that---after pivot weighting and relevance gating---their information flow
tends to steer pivot-local distributions in the same direction as the realized rollout.
The strongest negative contributors (chunks 8 and 6) correspond to meta-level planning statements about avoiding brute-force computation and reasoning about behavior under repeated composition.
Their negative sign indicates that, in this rollout, masking these segments makes the pivot-local trajectory \emph{more} aligned with $g$, consistent with these planning digressions being partially off-track or redundant relative to the eventual algebraic route taken.

\paragraph{Curvature diagnostic (geometry, not attribution).}
CurvImpact reflects changes in turning angles in raw logit space under the same pivot-local masking interventions.
Chunk 12 shows notably larger CurvImpact magnitude among the top contributors, consistent with a local reorientation moment (e.g., correcting a denominator/sign detail and tightening the algebra).
As elsewhere, curvature is a diagnostic of intervention-response geometry and does not determine causal sign or importance.

\paragraph{Qualitative interpretation.}
The dominant positive chunks emphasize recognizing the iterated-composition structure and concretely expanding early iterates (chunks 5, 2, 4, 10, 13), which is typically the gateway to spotting a functional fixed point, cycle, or closed form for $f^{\circ n}$.
The presence of a correction-oriented chunk with high CurvImpact (chunk 12) is consistent with the trace making a local algebraic adjustment that stabilizes the remainder of the derivation.
Overall, DRTC highlights the segments that support the realized approach (setting up composition mechanics and computing/confirming the algebra), while the negative-sign chunks flag moments where generic planning text is not the main driver of the eventual trajectory.

\subsubsection{\texttt{math\_rollout\_3374} (R1-Distill-Qwen-1.5B)}
\paragraph{Problem.}
\begin{quote}\small
When the base-16 number $66666_{16}$ is written in base 2, how many base-2 digits (bits) does it have?
\end{quote}

\paragraph{Gold answer.}
\begin{quote}\small
$19$
\end{quote}

\begin{figure}[h]
  \centering
  \begin{subfigure}[t]{0.48\textwidth}
    \centering
    \includegraphics[width=\linewidth]{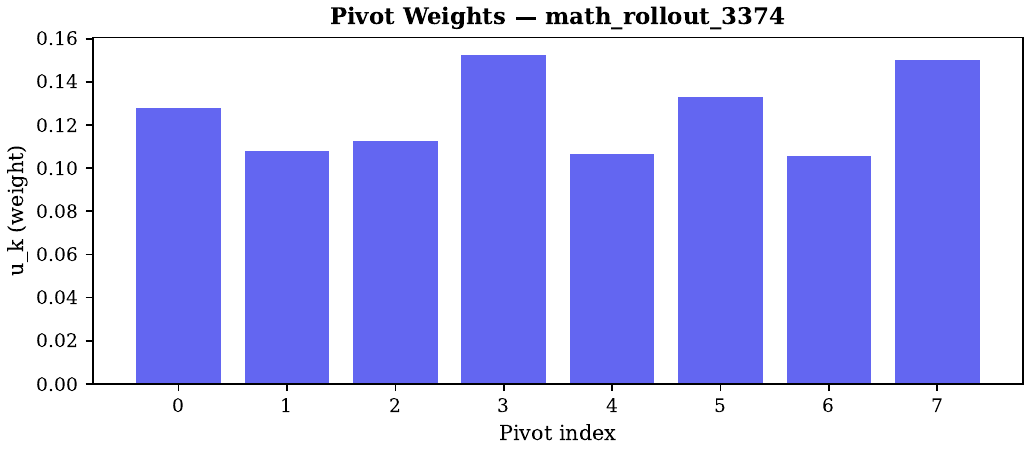}
    \caption{\scriptsize Pivot-weight distribution ($u_k$) for \texttt{math\_rollout\_3374} (R1-Distill-Qwen-1.5B)}
  \end{subfigure}
  \hfill
  \begin{subfigure}[t]{0.48\textwidth}
    \centering
    \includegraphics[width=\linewidth]{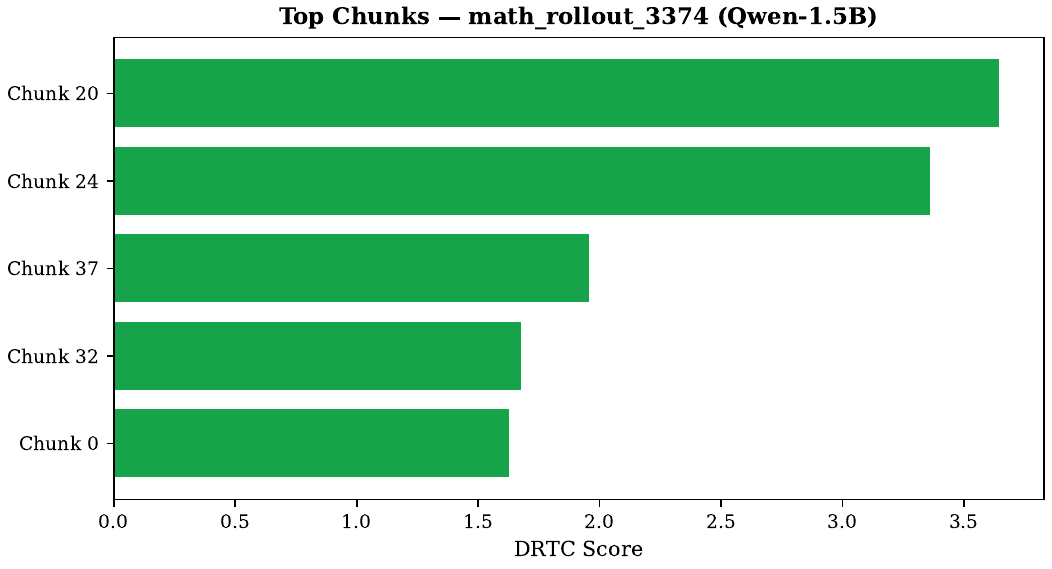}
    \caption{\scriptsize Top chunk attributions (DRTC) for \texttt{math\_rollout\_3374} (R1-Distill-Qwen-1.5B)}
  \end{subfigure}

  \caption{
  Case-study visual summary for \texttt{math\_rollout\_3374} (R1-Distill-Qwen-1.5B).
  Left: pivot importance weights $u_k$ across the $K$ detected pivots.
  Right: top-ranked chunks by signed DRTC (process-causal directional steering), shown for qualitative auditing.
  }
  \label{fig:qwen_3374_visuals}
\end{figure}

\begin{table*}[h]
\caption{
R1-Distill-Qwen-1.5B, \texttt{math\_rollout\_3374}: top contributing chunks by DRTC.
Positive DRTC aligns with the realized rollout direction; negative DRTC opposes it.
CurvImpact is reported as a geometric diagnostic only.
}
\label{tab:qwen_3374_top}
\centering
\scriptsize
\setlength{\tabcolsep}{4pt}
\renewcommand{\arraystretch}{1.15}
\begin{tabularx}{\textwidth}{r c r r r Y}
\toprule
Rank & Chunk idx & DRTC & $|\mathrm{DRTC}|$ & CurvImpact & Excerpt \\
\midrule
1  & 20 & +3.6469 & 3.6469 & 0.1574  & 0-5, but here we have 6. Wait, that can't \\
2  & 24 & +3.3646 & 3.3646 & 0.0441  & Wait, hold on, maybe I misread the problem. Wait, the number \\
3  & 30 & -2.1259 & 2.1259 & 0.0195  & 6, which is hexadecimal, but the problem says base-6.number.6 \\
4  & 37 & +1.9587 & 1.9587 & 0.0612  & but the number is 66666\_16. So \\
5  & 32 & +1.6817 & 1.6817 & 0.0359  & the problem says base-16, but the number is 66 \\
6  & 0  & +1.6305 & 1.6305 & -0.0005 & ) does it have? Okay, so I have this problem here: I need \\
7  & 6  & +1.4222 & 1.4222 & 0.0059  & to convert it to base-2. Alternatively, maybe I can convert it to \\
8  & 41 & +1.3853 & 1.3853 & 0.0044  & 66\_16? Or perhaps it's a typo. Wait, maybe \\
9  & 38 & +1.2265 & 1.2265 & -0.0173 & \_16. So, in base-6, each digit is 6 \\
10 & 9  & +1.1928 & 1.1928 & 0.0042  & \_16. So, in base-10, that would be \\
\bottomrule
\end{tabularx}
\end{table*}

\paragraph{Opposing-sign chunks (directional tension).}
Nearly all top-ranked chunks have positive DRTC (e.g., chunks 20, 24, 37, 32, 0, 6, 41, 38, 9), indicating that---after pivot weighting and relevance gating---their information flow
tends to steer pivot-local distributions in the same direction as the realized rollout.
The single large negative contributor (chunk 30) corresponds to confusion about the base (misparsing as ``base-6'' vs.\ base-16).
Its negative sign indicates that masking this segment makes the pivot-local trajectory \emph{more} aligned with $g$,
consistent with it functioning as a brief derailment or noisy alternative interpretation rather than a supportive step.

\paragraph{Curvature diagnostic (geometry, not attribution).}
CurvImpact reflects turning-angle changes in raw logit space under the same pivot-local masking interventions.
The highest-DRTC chunks (20 and 24) also show relatively larger CurvImpact magnitudes, consistent with a local reorientation event where the trace corrects a misread and locks onto the intended base-16-to-base-2 conversion framing.
As elsewhere, curvature is diagnostic and does not determine the sign or magnitude of DRTC.

\paragraph{Qualitative interpretation.}
DRTC assigns high positive mass to the segments where the rollout resolves an initial parsing error (chunks 20, 24, 37, 32) and re-centers on the intended representation conversion task.
The prominent negative chunk (30) flags the competing ``wrong-base'' branch as oppositional to the realized direction, providing a compact indicator of where the trace momentarily deviates before returning to the correct framing.
Overall, the attribution pattern is consistent with DRTC highlighting (i) the correction/reframing steps that restore the intended interpretation and (ii) the brief competing branch that is subsequently abandoned.

\subsection{Qualitative evidence summary}
\label{app:qual_summary}

Across all four models, the case studies suggest that high-magnitude DRTC chunks are not arbitrary: they frequently
correspond to problem-specific constraints, reframing moves that set the intended strategy, or local correction steps that
stabilize the subsequent derivation.
When negative-sign chunks appear among the top contributors, their excerpts typically reflect competing framings,
meta-level planning, or other detours whose presence (in the realized rollout) steers pivot-local distributions away from
the realized direction.
CurvImpact provides a complementary diagnostic view: large-magnitude CurvImpact often coincides with moments of local
reorientation (e.g., restatement$\rightarrow$strategy, correction, or reinterpretation), but it is not used as an importance score.
Overall, these qualitative patterns complement the quantitative results by supporting semantic plausibility while remaining
agnostic to outcome-level causality and circuit identification.

\section{Expanded related work}
\label{app:related_work_expanded}

Table~\ref{tab:related_work_comparison} situates DRTC relative to recent frameworks for interpreting multi-step reasoning traces. We group prior work by its primary explanatory target and intervention style, and highlight what is unique about DRTC: (i) explicit pivot discovery to localize decision points, (ii) receiver-side interventions applied \emph{only at pivots} that hold the realized prefix fixed and avoid generating a new continuation, (iii) signed directional steering in log-probability space, and (iv) curvature used strictly as an intervention-response diagnostic rather than an intrinsic importance score. We also benchmark DRTC against practical chunk-level attribution baselines (Appendix~\ref{app:baselines}). Circuit-discovery methods such as ACDC/EAP/CD-T target internal mechanisms at a different granularity, and are best viewed as complementary follow-ups seeded by top-DRTC chunks/pivots.

\begin{table*}[h]
\caption{
Comparison of reasoning-interpretability frameworks and their explanatory targets and intervention styles.
}
\label{tab:related_work_comparison}
\centering
\scriptsize
\setlength{\tabcolsep}{4pt}
\renewcommand{\arraystretch}{1.15}
\begin{tabularx}{\textwidth}{p{0.15\textwidth} p{0.17\textwidth} p{0.19\textwidth} p{0.17\textwidth} Y}
\toprule
\textbf{Framework family} & \textbf{Representative works} & \textbf{Primary explanatory target} & \textbf{Intervention / comparison style} & \textbf{Why DRTC is distinct} \\
\midrule

Step faithfulness \& resampling
&
\citet{zhao2026ahamomentsfakeidentifying}; \citet{macar2025thoughtbranchesinterpretingllm}
&
Whether specific CoT steps are causally \emph{used} vs.\ decorative; resilience/necessity of semantic content for the \emph{final answer}
&
Stochastic perturbation of steps; on-policy resampling to compare distributions over alternative traces
&
Targets \emph{necessity or faithfulness} (often outcome- or decision-centric). DRTC instead localizes \emph{decision pivots} and measures \emph{signed directional steering} in log-probability space under a method-faithful counterfactual that \emph{does not resample a new trace}. \\

Recursive attribution graphs
&
\citet{walker2025explainingreasoninglargelanguage}
&
Inter-generational influence pathways across prompt and prior generations
&
Constructs an attribution graph using a base attribution method (e.g., saliency) and marginalizes path contributions
&
Provides a \emph{global dependency structure} over full traces. DRTC uses \emph{sparse pivot-local causal probes} with receiver-side masking and produces \emph{directional (signed) steering} rather than path-marginal dependence scores. \\

Geometry-first trajectory analyses
&
\citet{zhou2025geometryreasoningflowinglogics}; \citet{manson2025curvedinferenceconcernsensitivegeometry}
&
Intrinsic geometric structure of reasoning dynamics (e.g., flow/shape); logic as a carrier-invariant skeleton; concern-sensitive residual-stream geometry
&
Primarily geometry-as-signal: representation/residual trajectories, pullback metrics, curvature/flow descriptors (typically not intervention-grounded causal attribution)
&
Treats geometry as an \emph{intrinsic signature}. DRTC uses geometry as an \emph{intervention-response diagnostic}: curvature is computed in logit space to summarize \emph{how} targeted causal interventions reorient the trajectory, while attribution remains defined by signed directional redirection. \\

Layer-wise dynamics \& decoding-time predictors
&
\citet{he2025deltadecodingstrategybased}; \citet{yan2025additionmovementsmappinglayerwise}
&
Vertical evolution across transformer depth; decoding improvements via layer-wise logit trajectory prediction; stage-wise information progression (probes/logit lens)
&
Cross-layer prediction/probing; layer-wise diagnostics and decoding-time modifications (not token-time causal tests)
&
Operates over \emph{depth} (layers) and often targets decoding quality. DRTC operates over \emph{token-time rollout depth} with pivot-local counterfactuals, explicitly attributing \emph{which earlier chunks steer} the unfolding reasoning trajectory. \\

Chunk-level attribution baselines
&
\citet{li2017understandingneuralnetworksrepresentation}; \citet{sundararajan2017axiomaticattributiondeepnetworks}; \citet{Fong_2017}; \citet{zhang2024bestpracticesactivationpatching}
&
Chunk importance for a chosen scalar (e.g., logit/probability/directional score) at a target position
&
Input perturbations (remove/replace chunks), gradient saliency, optimized masking, or activation patching at selected layers/positions
&
Provides \emph{non-pivoted} chunk attributions that do not explicitly localize decision pivots or enforce pivot-local counterfactuals. DRTC differs by (i) discovering pivots, (ii) applying receiver-side masking only at pivots while holding the realized prefix fixed, and (iii) scoring \emph{directional} redirection relative to the realized trajectory. \\

\midrule
\textbf{Directional causal attribution (DRTC)}
&
\textbf{This work}
&
Causal attribution of long-horizon reasoning: which earlier context redirects the rollout and in what direction
&
Receiver-side pivot masking (block information flow from a selected chunk only at a pivot while holding the realized prefix fixed and evaluating pivot-local counterfactual logits, without generating a new continuation); directional projection in log-probability space
&
\textbf{Unique combination:} (i) explicit pivot discovery to localize decision points; (ii) method-faithful, pivot-local counterfactuals that avoid off-policy resampling; (iii) \emph{signed} directional attribution via alignment with the realized trajectory direction; (iv) curvature used only to organize intervention-response geometry, not as an intrinsic importance score. \\
\bottomrule
\end{tabularx}
\end{table*}

\clearpage
\section{Interactive inspection interface}
\label{app:ui}

Understanding long-horizon reasoning requires not only quantitative attribution scores but also tools for inspecting how those attributions relate to the model’s internal reasoning dynamics. To support qualitative verification of DRTC attributions and trajectory-level diagnostics, we provide an interactive inspection interface that links text spans, causal attribution scores, and geometric structure in the model’s reasoning trajectory.

We include representative static views of this interface here to illustrate how DRTC outputs can be examined by a human analyst. These figures are intended for qualitative inspection only and are not used to derive or validate any quantitative results in the paper. All quantitative results reported in the main text are computed directly from exported artifacts and CSV files, independent of the interface.

\subsection{Textual reasoning with causal attribution}

\begin{figure}[h]
  \centering
  \includegraphics[width=0.95\linewidth]{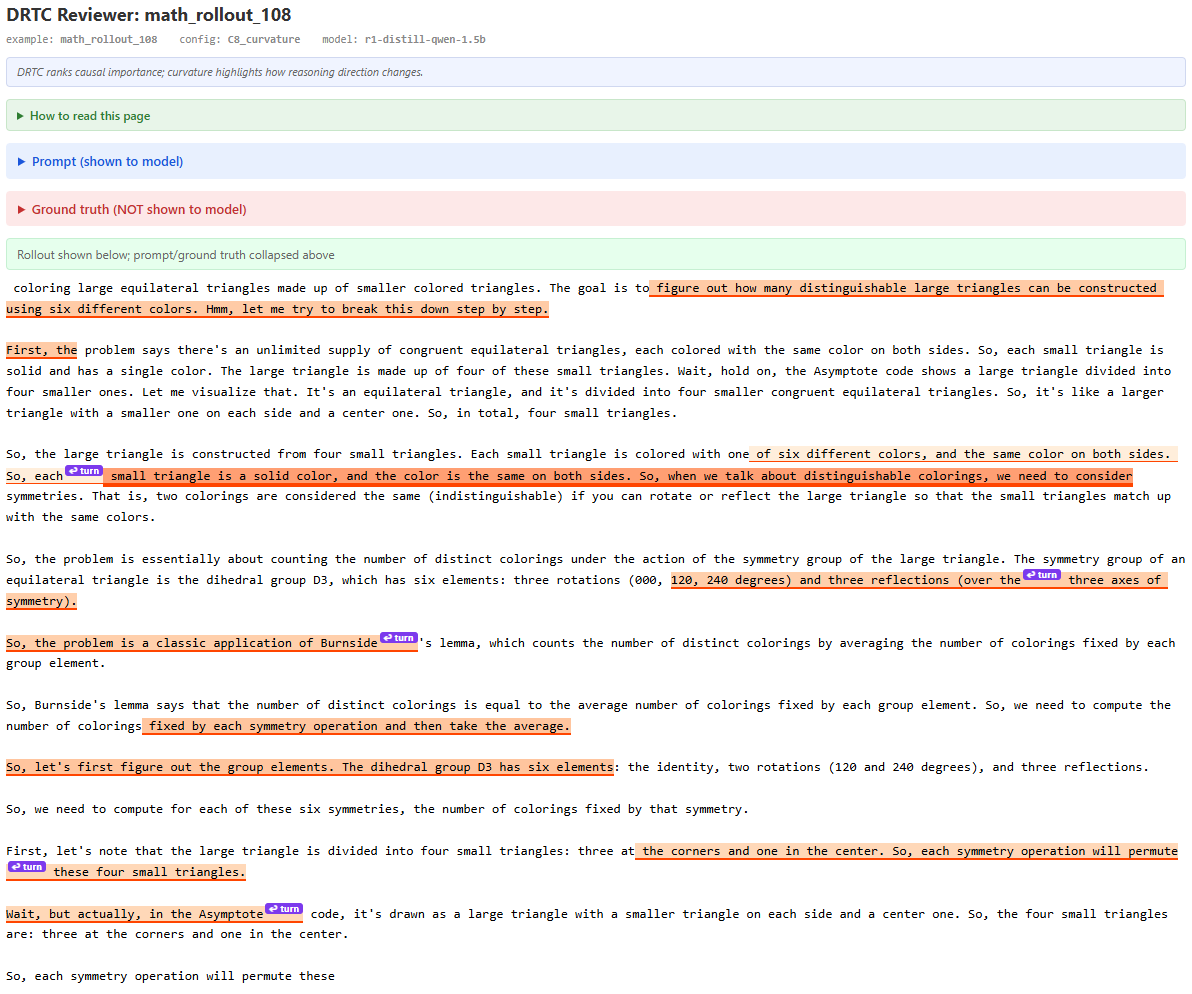}
  \caption{Left panel of the qualitative inspection interface.
  The model’s self-generated reasoning rollout is shown with text spans corresponding to high-magnitude DRTC scores highlighted.
  Highlight intensity reflects attribution strength.
  Prompt and ground-truth information are separated and collapsed to ensure inspection focuses on the model’s own reasoning rather than the problem statement or reference solution.}
  \label{fig:ui_left}
\end{figure}

\clearpage
\subsection{Trajectory geometry and attribution ranking}

\begin{figure}[h]
  \centering
  \begin{minipage}[t]{0.48\linewidth}
    \centering
    \includegraphics[width=\linewidth]{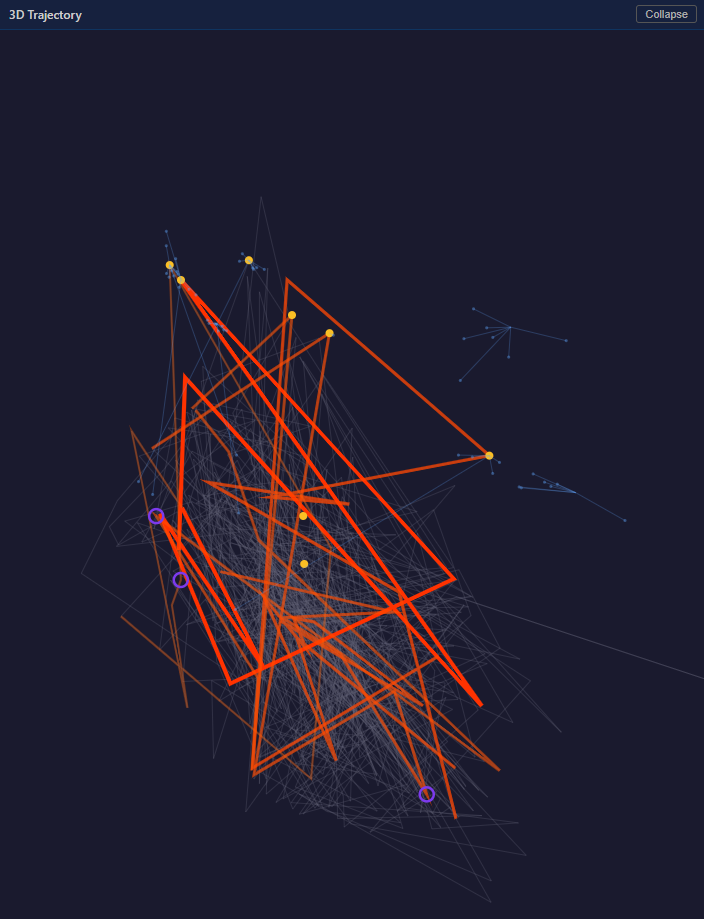}
    \caption*{(a) Trajectory geometry.}
  \end{minipage}
  \hfill
  \begin{minipage}[t]{0.48\linewidth}
    \centering
    \includegraphics[width=\linewidth]{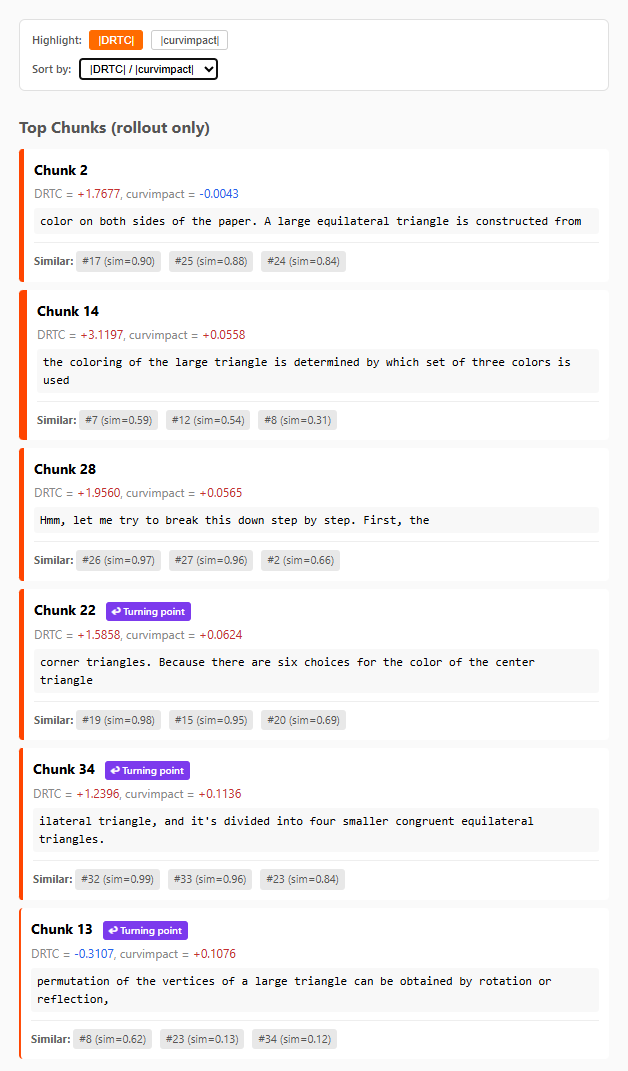}
    \caption*{(b) Chunk-level attribution ranking.}
  \end{minipage}
  \caption{Center and right panels of the qualitative inspection interface.
  (\textbf{a}) The reasoning process is visualized as a trajectory in representation space using a three-dimensional PCA projection of hidden states, with segments corresponding to selected text spans highlighted.
  High-curvature turning points (computed in logit space) indicate sharp local reorientations of the reasoning trajectory under targeted interventions.
  (\textbf{b}) Reasoning chunks are ranked by signed DRTC score.
  Positive scores indicate chunks whose removal disrupts the realized reasoning path, while negative scores indicate anti-aligned influence.
  Curvature diagnostics are geometric descriptors and are not interpreted as causal importance scores.}
  \label{fig:ui_center_right}
\end{figure}

\subsection{Interpretation and usage}

Taken together, these views allow analysts to examine not only which reasoning steps causally influence a model’s trajectory, but also how those steps reshape the direction of reasoning over time. The interface supports bidirectional linking between text spans and trajectory geometry, enabling qualitative inspection of attribution results without relying solely on scalar importance measures.

\end{document}